%% file: main.tex
\documentclass[lettersize,journal]{IEEEtran}
\usepackage[dvipsnames, rgb, table]{xcolor}
\usepackage{lipsum}
\usepackage{color}
\usepackage{soul}
\usepackage[hidelinks]{hyperref}
\usepackage{amsmath,amsfonts,amssymb}
\usepackage{algorithmic}
\usepackage{algorithm}
\usepackage{array}
\usepackage{textcomp}
\usepackage{stfloats}
\usepackage{url}
\usepackage{verbatim}
\usepackage{graphicx}
\usepackage{cite}
\usepackage{subcaption}
\usepackage{tikz}
\usepackage{bm}
\usepackage{graphics}
\usepackage{pgfplots}
\pgfplotsset{compat=1.18}
\usepackage{pgfplotstable}
\usepackage{booktabs}
\usepackage{float}
\usepackage{multirow}
\usepackage{etoolbox}
\usepackage{adjustbox}
\usepackage{diagbox}
\usepackage{epsfig} 
\usepgfplotslibrary{fillbetween}
\usepgfplotslibrary{statistics}
\pgfdeclarelayer{background}
\pgfdeclarelayer{back background}
\pgfsetlayers{back background, background, main}
\usetikzlibrary{calc}
\usetikzlibrary{matrix}
\usetikzlibrary{fit}
\usetikzlibrary{bayesnet}
\usetikzlibrary{arrows}
\usetikzlibrary{arrows.meta}
\usetikzlibrary{positioning}
\usetikzlibrary{shapes}
\usetikzlibrary{shapes.geometric}
\usetikzlibrary{shadows}
\usetikzlibrary{patterns}
\usetikzlibrary{decorations.pathreplacing}
\usetikzlibrary{fadings}
\usepgfplotslibrary{groupplots}

\definecolor{ffblue}{RGB}{097, 108, 140}
\definecolor{ffdarkgreen}{RGB}{086, 140, 135}
\definecolor{fflightgreen}{RGB}{178, 213, 155}
\definecolor{ffyellow}{RGB}{242, 222, 121}
\definecolor{ffred}{RGB}{217, 095, 024}

\definecolor{ffred_pv}{RGB}{202, 074, 046}
\definecolor{fforange_pv}{RGB}{232, 141, 047}
\definecolor{ffgreen_pv}{RGB}{059, 165, 149}
\definecolor{ffgreendark_pv}{RGB}{032, 117, 106}

\definecolor{nature_tab_gray1}{HTML}{D8D6C2}
\definecolor{nature_tab_gray2}{HTML}{ECEADF}
\begin{document}

\title{Knowledge Reutilization in Meta-Reinforcement Learning}

\author{Yuan~Meng$^{1,2,*}$,
        Bo~Wang$^{1,2,*}$,
        Juan~de~los~Rios~Ruiz$^{1}$,
        Xiangtong Yao$^{1}$, \\
        Zhenshan~Bing$^{2,\dagger}$,
        Fuchun~Sun$^{3}$,~\IEEEmembership{Fellow,~IEEE},
        and~Alois~Knoll$^{1}$,~\IEEEmembership{Fellow,~IEEE}

        \thanks{$^{1}$ School of Computation, Information and Technology, Technical University of Munich, Munich, Germany}
        \thanks{$^{2}$ State Key Laboratory for Novel Software Technology, Nanjing University, Suzhou, China}
        \thanks{$^{3}$ Department of Computer Science and Technology, Tsinghua University, Beijing, China}
        \thanks{$^{*}$ Equal contribution. The work was done during the research visit at Nanjing University.}
        \thanks{$^{\dagger}$ Corresponding author: bing@nju.edu.cn}
}

\markboth{Journal of \LaTeX\ Class Files,~Vol.~14, No.~8, August~2021}%
{Shell \MakeLowercase{\textit{et al.}}: A Sample Article Using IEEEtran.cls for IEEE Journals}


\maketitle

\begin{abstract}
Meta-reinforcement learning enables fast adaptation by extracting shared structure from related tasks, but existing end-to-end methods often couple task inference with embodiment-specific control. 
This coupling can obscure non-parametric task semantics, reduce sample efficiency, and limit cross-agent reuse. 
We propose a meta-knowledge reutilization framework that learns task-level knowledge on a dynamics-simplified agent and transfers it to heterogeneous agents. 
The framework uses a Bayesian non-parametric prior to organize latent task modes and a high-level policy to generate task-level magnitude guidance. 
To bridge reusable task knowledge with different embodiments, we introduce a semantic-magnitude interface and a lightweight temporal adaptor, which convert frozen meta-knowledge into temporally aligned subgoals for embodiment-specific low-level controllers. 
Experiments on multiple locomotion agents show that our framework reduces final-step tracking error by \(94.75\%\)--\(99.79\%\) compared with recent state-of-the-art baselines and achieves comparable deployment performance with about \(23.8\%\) of their interaction data.
\end{abstract}


\section{INTRODUCTION}
Rapid adaptation to unseen tasks with limited interaction is a key challenge in robotic deep reinforcement learning (RL). 
Meta-reinforcement learning (Meta-RL)~\cite{duan2016rl,beck2023recurrent,finn2017model,bing2022meta} addresses this challenge by combining outer-loop meta-learning with inner-loop policy adaptation, enabling agents to learn how to adapt from limited experience. 
However, existing Meta-RL approaches still face limitations in how meta-knowledge is represented, acquired, and reused. 
In terms of representation, representative methods such as PEARL~\cite{rakelly2019efficient} adopt an end-to-end framework in which task inference and policy learning are jointly optimized under a fixed Gaussian latent prior. 
Although this design is computationally convenient, it imposes a strong parametric assumption on the latent task structure. 
In robotic tasks, task semantics are often independent of a specific agent's dynamics and may exhibit multi-modal and non-parametric structures~\cite{yu2020meta}. 
Such structures are difficult to describe with a single Gaussian distribution, causing heterogeneous task modes to be compressed into a smooth latent space, which can obscure task boundaries and increase the interaction required for reliable task inference. 
Recent studies have shown that more expressive Bayesian parametric or Bayesian non-parametric priors can improve the representation capability of latent spaces~\cite{bing2022meta,10495171,hughes2013memoized}. 
Nevertheless, a richer prior alone does not solve the acquisition and reuse problem. 
When task inference and policy learning are still optimized jointly under the same embodiment-specific dynamics, the learned latent representation may encode not only task semantics but also agent-specific action patterns. 
As a result, the obtained meta-knowledge can be entangled with control-dependent behaviors, limiting its transferability across agents with different embodiments.

\input{img_tex/introduction_overview}
This coupling further leads to an exploration dilemma in end-to-end Meta-RL. 
The inference module requires informative interaction data to identify the current task, while collecting such data depends on a policy that can already execute meaningful behaviors under the inferred task context. 
During early learning, neither component is sufficiently developed: the policy explores inefficiently, and the inference module receives limited or biased context for task identification. 
This chicken-and-egg problem makes end-to-end Meta-RL highly sample-inefficient, especially for agents with complex morphology and dynamics. 
These observations reveal three key gaps. 
First, commonly used fixed Gaussian latent priors are insufficient for representing non-parametric task structures. 
Second, end-to-end training tightly couples task inference with low-level control, leading to inefficient meta-knowledge acquisition. 
Third, the learned meta-knowledge is often entangled with agent-specific dynamics, making cross-embodiment transfer and reutilization difficult. 
Addressing these gaps requires redesigning both the representation and acquisition processes for meta-knowledge. 
Based on these limitations, we identify three fundamental questions:
\begin{itemize}
    \item \textbf{Q1}: How can we learn expressive meta-knowledge that captures non-parametric task semantics beyond a fixed Gaussian prior?
    \item \textbf{Q2}: How can we decouple task inference from agent-specific dynamics and low-level control?
    \item \textbf{Q3}: How can the learned meta-knowledge be effectively reused across agents with different embodiments?
\end{itemize}

To address these questions, we propose \textbf{ReMAP}: \textbf{Re}usable \textbf{M}eta-knowledge for \textbf{A}daptive \textbf{P}olicy Transfer. An overview is shown in Fig.~\ref{fig:overview}. 
First, to learn transferable meta-knowledge beyond a fixed Gaussian prior (\textbf{Q1}), we train the task inference module and high-level policy on a dynamics-simplified agent. This agent preserves task-relevant variables such as position and velocity, while removing high-dimensional embodiment-specific control complexity. 
We further impose a Bayesian non-parametric prior, specifically a Dirichlet Process Mixture Model (DPMM), on the latent task space, allowing the model to adaptively organize non-parametric task modes without predefining the number of components.
Second, to decouple task inference from low-level control (\textbf{Q2}), we split training into two independent branches with different objectives. 
Unlike standard end-to-end Meta-RL, where one policy must both interpret task context and learn complex embodiment-specific control, ReMAP learns task inference and high-level magnitude generation only on the dynamics-simplified agent. 
Each target agent then trains its low-level policy separately through magnitude-guided curriculum learning, where the objective is not to infer task semantics but to track progressively generated task-level magnitude commands and acquire stable embodiment-specific motion patterns. 
This design avoids the chicken-and-egg problem between task inference and meaningful exploration, and prevents the latent task representation from being optimized through complex robot dynamics.
Third, to reuse meta-knowledge across heterogeneous embodiments (\textbf{Q3}), we introduce a Semantic-Magnitude Alignment Interface (SMAI) as the bridge between task-level reasoning and embodiment-specific execution. 
The frozen task inference module predicts the semantic task context, and the frozen high-level policy generates a scalar magnitude. 
SMAI embeds this magnitude into the corresponding semantic guidance channel, producing a shared semantic-magnitude subgoal that can be interpreted by different low-level controllers. 
Since different embodiments may require different execution durations to realize the same task-level command, we further introduce a lightweight stride predictor for few-shot online adaptation. 
It estimates how long each SMAI subgoal should be executed by the target agent, thereby temporally aligning the frozen task-level guidance with embodiment-specific motion dynamics. 
Thus, the same meta-knowledge module can be reused across agents, while each low-level policy only needs to translate the shared SMAI command into feasible motor actions under its own dynamics. 
We evaluate ReMAP on multiple locomotion agents with different embodiments and compare it with representative Meta-RL baselines, including \(\mathrm{RL}^{2}\)~\cite{duan2016rl}, PEARL~\cite{rakelly2019efficient}, CEMRL~\cite{bing2022meta}, and MELTS~\cite{10495171}. 
Experiments show that ReMAP learns semantically separable task representations, reuses the frozen meta-knowledge module across agents, and achieves accurate downstream execution. 
At the final time step, ReMAP reduces the overall tracking MSE by \(94.75\%\)--\(99.79\%\) compared with the baselines. 
Moreover, it achieves state-of-the-art deployment performance with \(38\)M interactions across four embodiments, which is about \(23.8\%\) of the \(160\)M interactions required by MELTS. 
The main contributions of this work are summarized as follows: 
\begin{itemize} 
    \item \textbf{(Addressing Q1 \& Q2)} 
    We propose a disentangled Meta-RL framework that learns task-level meta-knowledge on a dynamics-simplified agent with a DPMM-regularized latent task space, encouraging the representation to encode task semantics rather than embodiment-specific dynamics. 
    \item \textbf{(Addressing Q3)} 
    We introduce SMAI and a lightweight stride predictor to convert frozen task-level meta-knowledge into temporally aligned semantic-magnitude subgoals, enabling reuse across heterogeneous agents with embodiment-specific low-level controllers.
    \item We provide comprehensive experiments on multiple locomotion embodiments, demonstrating improved tracking accuracy, higher sample efficiency, and lower cross-embodiment transfer cost than state-of-the-art Meta-RL baselines, including \(\mathrm{RL}^{2}\), PEARL, CEMRL, and MELTS.   
\end{itemize}

\section{RELATED WORK}
\subsection{Meta-Reinforcement Learning}
Meta-RL aims to enable agents to rapidly adapt to unseen tasks by leveraging experience collected from a distribution of related tasks. 
Existing methods can be broadly categorized into gradient-based, recurrence-based, and context-based approaches. 
Gradient-based methods learn an initialization or update rule that can be efficiently adapted to a new task through a few optimization steps~\cite{finn2017model,nichol2018first,nguyen2021robust}. 
Although effective for fast within-embodiment adaptation, they mainly update policy parameters under the same agent dynamics, making the learned prior difficult to reuse across different morphologies. 
Recurrence-based methods encode adaptation into a history-dependent policy by aggregating past observations, actions, and rewards with recurrent or sequential models~\cite{duan2016rl,mishra2018simple,beck2023recurrent,shala2025efficient}. However, the resulting task representation is typically implicit and jointly optimized with the control policy, which limits interpretability and cross-embodiment reuse.
Context-based methods explicitly infer a latent task representation from recent experience and condition the policy on this representation~\cite{rakelly2019efficient,zintgraf2020varibad,li2020focal,bing2022meta,wang2023meta,10495171}. 
Recent advances improve task inference through probabilistic modeling, metric learning, self-supervised objectives, mixture-based representations, or Bayesian non-parametric priors, showing that richer latent structures are beneficial for non-stationary and non-parametric task variations. 
Nevertheless, most existing methods still learn task inference and policy control in an end-to-end manner under embodiment-specific dynamics. 
Consequently, the learned meta-knowledge may entangle task semantics with agent-specific motion patterns, leading to inefficient meta-knowledge acquisition and limited transferability when the target agent has different dynamics or morphology. 
In contrast, our work explicitly decouples task inference from low-level control, learns expressive meta-knowledge on a dynamics-simplified agent with a DPMM prior, and reuses the frozen task-level module across different embodiments.

\subsection{Deep Clustering and Non-Parametric Task Representation}
Task inference in Meta-RL is closely related to deep clustering, which combines representation learning with latent structure discovery~\cite{lu2024survey}. Existing generative deep clustering methods, such as VaDE and GMVAE~\cite{jiang2016variational,dilokthanakul2016deep}, commonly employ Gaussian mixture priors~\cite{reynolds2009gaussian} but require a predefined number of clusters, limiting their applicability when task structure is unknown or evolving.
Bayesian non-parametric models address this limitation by adapting model complexity to the data~\cite{hughes2013memoized,meng2025preserving,nalisnick2017stick,goyal2017nonparametric,ronen2022deepdpm}. 
In particular, DPMM infers the number of latent components automatically and explicitly, making them well suited for heterogeneous and unstructured task distributions~\cite{blei2006variational,hughes2013memoized}.
In Meta-RL, Bayesian parametric and non-parametric latent priors have been explored for robotic task inference~\cite{bing2022meta,10495171}. 
However, they are typically embedded in end-to-end pipelines where task representation and policy control are learned under the same embodiment-specific dynamics, which may entangle task semantics with agent-specific motion patterns. 
Consequently, their potential as reusable meta-knowledge across embodiments remains limited. 
Our work addresses this gap by incorporating a Bayesian non-parametric prior into a disentangled framework that separates task-level inference from low-level control.

\subsection{Cross-Embodiment Learning}
Cross-embodiment learning is challenging because agents may differ in state spaces, action spaces, morphology, and dynamics. 
Existing few studies mainly focus on policy transfer or morphology-conditioned controller learning, using hierarchical decomposition or behavior alignment to improve control reuse across agents~\cite{hejna2020hierarchically,gupta2022metamorph}. 
However, these methods primarily transfer executable policies that remain coupled with embodiment-specific dynamics, while the reuse of task-level meta-knowledge in Meta-RL remains less explored. 
End-to-end Meta-RL further entangles task inference with policy control, making latent task representations difficult to reuse when the embodiment changes. 
ReMAP addresses this issue by learning DPMM-regularized task-level meta-knowledge on a dynamics-simplified agent and using it to guide embodiment-specific low-level controllers, thereby separating task reasoning from physical execution.

\section{BACKGROUND}

This section briefly introduces the decision-making foundations used in our framework. We review Markov Decision Processes (MDPs) and Partially Observable MDPs (POMDPs) as standard formulations for fully and partially observable control~\cite{sutton1999between}, and then relate them to Meta-RL, where task identity can be treated as a latent variable inferred from interaction data. This perspective motivates our use of Bayesian non-parametric inference for reusable task-level meta-knowledge acquisition.

\subsection{From MDPs and POMDPs to Meta-RL}\label{subsec:mdp_pomdp_metal}
An MDP provides a standard formulation for sequential decision-making in fully observable environments~\cite{sutton1999between}. 
It is defined by the tuple \(\langle \mathcal{S}, \mathcal{A}, \mathcal{P}, \mathcal{R}, \gamma \rangle\), where \(\mathcal{S}\) and \(\mathcal{A}\) denote the state and action spaces, \(\mathcal{P}(s'|s,a)\) is the transition probability, \(\mathcal{R}(s,a)\) is the reward function, and \(\gamma \in [0,1)\) is the discount factor. At each step, the agent observes \(s_t\), samples an action \(a_t \sim \pi(\cdot|s_t)\), receives reward \(r_t\), and transitions according to \(\mathcal{P}(s_{t+1}|s_t,a_t)\). 
The learning objective is to find a policy that maximizes the expected discounted return:
\begin{equation}
\label{eq}
J(\pi) = \mathbb{E}_{\tau \sim \pi} \left[ \sum_{t=0}^{\infty} \gamma^t \mathcal{R}(s_t,a_t) \right]\text{,}
\end{equation}
where \(\tau\) denotes the trajectory induced by the policy and environment dynamics.

In many robotic tasks, the full state is not directly observable. This setting is modeled as a POMDP, which augments an MDP with an observation space \(\mathcal{O}\) and an observation model \(\mathcal{Z}(o|s)\). 
The agent receives observations \(o_t\) generated from latent states rather than observing \(s_t\) directly. 
Therefore, the policy must depend on the interaction history \(h_t\) or a learned belief representation that summarizes hidden state information. 
From this view, decision-making under partial observability requires both inference and control: the agent must infer latent information from past interactions while selecting actions to maximize long-term return.

RL learns policies through interaction with the environment, typically by optimizing parameters \(\theta\) of a policy \(\pi_\theta\) to maximize \(J(\pi_\theta)\). 
In POMDP settings, the policy is often conditioned on history or latent representations that approximate the belief state. 
While standard RL usually focuses on a single task, Meta-RL aims to generalize across a distribution of related tasks \(p(\mathcal{T})\). During meta-training, the agent learns action patterns from training tasks, whereas during meta-testing, it is evaluated on non-overlapping or even unseen tasks that are disjoint from the training set and require adaptation with limited interaction~\cite{rakelly2019efficient}. 
\subsection{Dirichlet Process and Stick-Breaking}\label{subsec:dp_stickbreaking}

The Dirichlet process (DP) is a Bayesian nonparametric prior over probability measures \cite{hughes2013memoized}.
It is commonly used when the number of latent components is unknown in advance.
Unlike parametric priors with a fixed number of components, a DP can support a countably infinite set of components, while only a finite subset is typically used for any finite dataset.
Formally, let \(H\) be a base distribution over a parameter space \(\Theta\), and let \(\alpha > 0\) be a concentration parameter.
A random probability measure \(G\) drawn from a DP is denoted as \(G \sim \mathrm{DP}(\alpha, H)\).
The base distribution \(H\) defines the expected location of the atoms, while \(\alpha\) controls how strongly \(G\) concentrates around \(H\).

A key property of the DP is that a draw \(G\) is discrete with probability one, even when \(H\) is continuous.
As a result, multiple samples from \(G\) may share the same parameter value, which naturally induces a clustering structure.
A constructive representation of the DP is given by the stick-breaking process:
\begin{equation}
\beta_k \sim \mathcal{B}(1,\alpha), \qquad
\pi_k = \beta_k \prod_{i=1}^{k-1}(1-\beta_i)\text{,}
\label{eq:stick_breaking}
\end{equation}
where \(\beta_k\) determines the fraction assigned to component \(k\) from the remaining stick.
The resulting weights \(\{\pi_k\}_{k=1}^{\infty}\) sum to one and follow the Griffiths-Engen-McCloskey (GEM) distribution.
A draw from the DP can then be written as
\begin{equation}
G = \sum_{k=1}^{\infty} \pi_k \delta_{\theta_k^*}\text{,}
\label{eq:dp_discrete}
\end{equation}
where \(\theta_k^* \sim H\) are atoms and \(\delta_{\theta_k^*}\) denotes a point mass at \(\theta_k^*\).
This representation shows that the DP provides a flexible prior for modeling an unknown number of latent components.

\subsection{Dirichlet Process Mixture Model}\label{subsec:dpmm}

The DPMM~\cite{hughes2013memoized} provides a Bayesian non-parametric formulation for mixture modeling, where the number of active components is inferred from data rather than fixed in advance. 
This flexibility is particularly useful for latent task modeling, since the number of task modes may be unknown or may grow with the observed experience. Using the stick-breaking representation, a DPMM can be written as
\begin{equation}
\begin{aligned}
\theta_k^* \mid H &\sim H, \\
\pi \mid \alpha &\sim \mathrm{GEM}(\alpha), \\
v_i \mid \pi &\sim \mathrm{Cat}(\pi), \\
x_i \mid v_i &\sim F(\cdot \mid \theta_{v_i}^*)\text{,}
\end{aligned}
\label{eq:dpmm_sb}
\end{equation}
where \(H\) is the base distribution, \(\alpha\) is the concentration parameter, \(\pi\) denotes the mixture weights, \(v_i\) is the component assignment of observation \(x_i\), and \(\theta_k^*\) parameterizes component \(k\). Although the model defines countably infinite candidate components, only a finite subset is activated for a finite dataset, allowing the latent structure to adapt to the observed data.

In this work, we interpret each active component as a latent \emph{task mode}.
Different components correspond to different patterns in the underlying task distribution.
Compared with Gaussian mixture models (GMMs), which require the number of components to be specified beforehand, DPMM can adapt the number of active task modes according to the observed trajectories.
This property is well-suited to Meta-RL, where tasks may be heterogeneous, non-parametric, or previously unseen.
Therefore, DPMM provides a principled prior for modeling nonparametric latent task structures, which supports task inference and meta-knowledge reuse.

\subsection{Variational Inference for DPMM}\label{subsec:vi_dpmm}

Exact posterior inference in a DPMM is generally intractable because the model contains a countably infinite number of possible mixture components.
Variational inference provides a tractable approximation by replacing the true posterior with a simpler distribution \cite{hughes2013memoized}.
Following the stick-breaking construction, the joint distribution can be written as
\begin{equation}
\begin{aligned}
p(x,v,\theta,\beta)
=&
\prod_{n=1}^{N}
F(x_n \mid \theta_{v_n})
\mathrm{Cat}(v_n \mid \pi(\beta)) \\
&\times
\prod_{k=1}^{\infty}
\mathcal{B}(\beta_k \mid 1,\alpha)
H(\theta_k \mid \lambda)\text{,}
\end{aligned}
\label{eq:joint_dpmm}
\end{equation}
where \(v_n\) denotes the mixture assignment of \(x_n\), \(\beta_k\) is the stick-breaking variable, \(\pi_k(\beta)=\beta_k\prod_{\ell=1}^{k-1}(1-\beta_\ell)\) is the corresponding mixture weight, and \(\theta_k\) denotes the parameter of component \(k\).

We approximate the posterior \(p(v,\theta,\beta \mid x)\) with a factorized variational distribution \(q(v,\theta,\beta)\).
The variational objective is to minimize the KL divergence
\begin{equation}
q^*(v,\theta,\beta)
=
\arg\min_q
\mathrm{KL}
\left(
q(v,\theta,\beta)
\|p(v,\theta,\beta \mid x)
\right)\text{.}
\end{equation}
%
Since \(\log p(x)\) is constant with respect to \(q\), this is equivalent to maximizing the evidence lower bound (ELBO) objective $\mathcal{L}$:
\begin{equation}
\mathcal{L}(q)
=
\mathbb{E}_q[\log p(x,v,\theta,\beta)]
-
\mathbb{E}_q[\log q(v,\theta,\beta)]\text{.}
\label{eq:elbo}
\end{equation}

To make the inference finite-dimensional, we use a truncated mean-field approximation:
\begin{equation}
q(v,\theta,\beta)
=
\prod_{n=1}^{N} q(v_n)
\prod_{k=1}^{K} q(\beta_k)q(\theta_k)\text{,}
\label{eq:meanfield_dpmm}
\end{equation}
where \(K\) is a sufficiently large truncation level.
This truncation is applied to the variational posterior, while the underlying DP prior remains non-parametric.
The variational factors are defined as
\begin{equation}
\begin{aligned}
q(v_n) &= \mathrm{Cat}(\hat{r}_{n1},\ldots, \hat{r}_{nK}), \\
q(\beta_k) &= \mathcal{B}(\hat{\alpha}_{k1},\hat{\alpha}_{k0}), \\
q(\theta_k) &= H(\theta_k \mid \hat{\lambda}_k)\text{,}
\end{aligned}
\label{eq:var_factors_dpmm}
\end{equation}
where \(\hat{r}_{nk}\) denotes the responsibility of component \(k\) for data point \(x_n\). With \(\hat{\cdot}\) indicates variational factor.

The local assignment factors are updated by
\begin{equation}
\begin{aligned}
\tilde{r}_{nk} &= \exp \left( \mathbb{E}_q[\log \pi_k(\beta)] + \mathbb{E}_q[\log F(x_n \mid \theta_k)] \right), \\
\hat{r}_{nk} &= \frac{\tilde{r}_{nk}} {\sum_{\ell=1}^{K}\tilde{r}_{n\ell}}\text{.}
\end{aligned}
\label{eq:local_update_dpmm}
\end{equation}
The expected log mixture weight is computed from the Beta variational factors as
\begin{equation}
\begin{aligned}
\mathbb{E}_q[\log \pi_k(\beta)]
=&
\psi(\hat{\alpha}_{k1})
-
\psi(\hat{\alpha}_{k1}+\hat{\alpha}_{k0}) \\
&+
\sum_{\ell=1}^{k-1}
\left[
\psi(\hat{\alpha}_{\ell0})
-
\psi(\hat{\alpha}_{\ell1}+\hat{\alpha}_{\ell0})
\right]\text{,}
\end{aligned}
\label{eq:expected_log_weight}
\end{equation}
where \(\psi(\cdot)\) is the digamma function.

For exponential-family likelihoods, the sufficient statistics of component \(k\) are
\begin{equation}
\hat{N}_k = \sum_{n=1}^{N} \hat{r}_{nk}\text{,} \qquad s_k(x) = \sum_{n=1}^{N} \hat{r}_{nk}t(x_n)\text{,}
\label{eq:sufficient_stats_dpmm}
\end{equation}
where \(t(x_n)\) denotes the sufficient statistic of the observation model.
The global variational parameters are then updated as
\begin{equation}
\hat{\alpha}_{k1} = 1+\hat{N}_k, \qquad
\hat{\alpha}_{k0} = \alpha + \sum_{\ell=k+1}^{K}\hat{N}_{\ell}, \qquad
\hat{\lambda}_k = \lambda + s_k(x)\text{.}
\label{eq:global_update_dpmm}
\end{equation}

Under this formulation, the ELBO can be expressed using the sufficient statistics \cite{hughes2013memoized,blei2006variational}:
\begin{equation}
\begin{aligned}
\mathcal{L}(q) = \sum_{k=1}^{K} \Big( &\mathbb{E}_q[\theta_k]^\top s_k(x) - \hat{N}_k\mathbb{E}_q[a(\theta_k)] \\
&+\hat{N}_k\mathbb{E}_q[\log \pi_k(\beta)] - \sum_{n=1}^{N} \hat{r}_{nk}\log \hat{r}_{nk} \\
&+ \mathbb{E}_q \left[ \log \frac{\mathcal{B}(\beta_k \mid 1,\alpha)}{q(\beta_k \mid \hat{\alpha}_{k1},\hat{\alpha}_{k0})}\right] \\
&+\mathbb{E}_q\left[\log\frac{H(\theta_k \mid \lambda)}{q(\theta_k \mid \hat{\lambda}_k)}\right]\Big)\text{,}
\end{aligned}
\label{eq:elbo_expanded}
\end{equation}
where \(a(\theta_k)\) is the log-normalizer of the exponential-family likelihood.
Optimization alternates between updating the local responsibilities in Eq.~\eqref{eq:local_update_dpmm} and updating the global parameters in Eq.~\eqref{eq:global_update_dpmm}.

To improve scalability, Memoized Variational Bayes (MemoVB) \cite{hughes2013memoized} maintains cached sufficient statistics for each data batch and updates the full-dataset statistics incrementally.
For a batch \(b\), let \(S_k^{b}=[\hat{N}_k^{b},s_k^{b}]\) denote the cached sufficient statistics of component \(k\), and let \(S_k^{0}\) denote the full-dataset statistics.
When batch \(b\) is revisited, MemoVB updates
\(S_k^{0} \leftarrow S_k^{0} - S_k^{b,\mathrm{old}} + S_k^{b,\mathrm{new}}\).
%
%
This allows global parameters to be updated using full-dataset sufficient statistics while only processing one mini-batch at a time.

MemoVB also introduces birth and merge moves to adapt the number of active components.
A birth move proposes new components from data that are poorly represented by the current mixture, allowing the model to capture newly observed patterns.
A merge move combines redundant components and accepts the proposal only when it improves the full-data ELBO.
Thus, MemoVB provides a practical mechanism for dynamically adjusting the active mixture structure while retaining a variational objective.

In our framework, the DPMM prior is applied to the latent representations produced by a variational autoencoder (VAE).
The VAE maps trajectory contexts into a structured latent space, while the DPMM clusters these latent variables into task modes.
This design allows the model to infer compact and adaptive latent task variables, providing a principled basis for task inference and meta-knowledge reuse in Meta-RL.

\section{METHOD}\label{sec:method}

\subsection{Preliminary Assumptions and Applicability}\label{subsec:preliminary}
The proposed framework learns task-level meta-knowledge on a dynamics-simplified agent and reuses it across different embodiments. 
This requires task semantics to be separable from embodiment-specific execution. 
We therefore define the following assumptions that characterize the applicability of our method.

\paragraph{Task-level objective invariance}
We focus on locomotion-oriented tasks whose objectives can be defined independently of robot morphology. 
The task goal should depend on environmental variables, such as position or velocity, rather than robot-specific variables such as joint angles, actuator commands, or contact configurations. 
Thus, reaching a spatial goal or tracking a target velocity can be shared across different agents, whereas tasks defined by morphology-specific manipulation interfaces are outside the intended scope.

\paragraph{Reward consistency}
Since the inference module uses rewards as part of the task context, reward signals should preserve consistent task semantics across agents. 
The reward functions need not be numerically identical, but they should measure the same task-related progress, such as distance to a goal or velocity-tracking error. 
Embodiment-specific terms, including joint-energy penalties or morphology-dependent action costs, should be normalized to avoid leaking agent-specific dynamics into the representation.

\paragraph{Dynamics abstraction feasibility}
The task-relevant behavior of each target agent should be representable through a simplified abstraction. Although target agents may have complex contacts, nonlinear actuation, or high-dimensional morphology, these factors should mainly affect task execution rather than task identification. 

Under these assumptions, the simplified agent preserves task-relevant variables while removing unnecessary control complexity. 
Consequently, the learned latent representation can capture task semantics rather than embodiment-specific motion patterns, enabling reuse across heterogeneous agents.

\subsection{Meta-knowledge Acquisition and Disentangled Learning}\label{subsec:training_acquisition}

\input{img_tex/framework}
As illustrated in Fig.~\ref{fig:framework}, the proposed framework comprises two main components: 
disentangled meta-learning (Addressing \textbf{Q1} \& \textbf{Q2}) and meta-knowledge reuse during inference (Addressing \textbf{Q3}).
The training part includes two independent parallel learning branches.
The first branch learns a task inference module and a high-level policy on a dynamics-simplified agent, aiming to acquire reusable task-level meta-knowledge.
The second branch disentangled train low-level policies for target complex agents, enabling them to obtain basic movement and subgoal-following capabilities.
After training, the task inference module, the high-level policy, and corresponding low-level policies are frozen and reused during inference.
This section describes the training part, including simplified agent modeling, meta-knowledge acquisition, and disentangled low-level policy warm-up.

\subsubsection{\textbf{Dynamics Abstraction and Simplified Agent Modeling}}\label{subsubsec:simplified_agent_modeling}
To acquire task-level meta-knowledge without encoding embodiment-specific control patterns, we introduce a dynamics-simplified agent (Mujoco instantiation see Fig.~\ref{fig:toy_agent}a) as a controlled abstraction.
The simplified agent is designed to preserve task-relevant variables, such as position, velocity, and orientation, while removing unnecessary control complexity caused by high-dimensional actuation, contact dynamics, and morphology-specific constraints.
As discussed in Section \ref{subsec:preliminary}, our framework primarily focuses on locomotion tasks; therefore, for tasks involving velocity or simple dynamic constraints, we use a mass-damper model (Fig.~\ref{fig:toy_agent}b):
\begin{equation}\label{eq:toy_dynamic}
    m \frac{d^2 x}{dt^2} = F_{\mathrm{ext}} - c \frac{d x}{dt} - m g \text{,}
\end{equation}
where \(F_{\mathrm{ext}}\) is the external force, \(c\) is the damping coefficient, and \(g\) denotes gravity.
The damping term ensures stable and physically plausible motion, while gravity introduces consistent environmental effects.
For position-based tasks, the agent can be further modeled as a point mass with bounded kinematic dynamics
%
%
\(s' = s + a\), with \(a \in [-a_{\max}, a_{\max}]\),
where \(s\) and \(s'\) denote the current and next states, and \(a\) denotes the abstract action. The bounded action space prevents unrealistic transitions and aligns the exploration statistics with those of more complex agents.
This abstraction keeps the motion physically meaningful while avoiding complex embodiment-specific dynamics.
As shown in Fig.~\ref{fig:toy_agent}, the simplified agent serves as an intermediate task space in which task semantics can be learned more cleanly than in a high-dimensional robot body.
Therefore, for the context-based Meta-RL, the latent representation learned on this agent is encouraged to capture task-level structure rather than low-level motion patterns.
\input{img_tex/simplified_agent_overview}

\subsubsection{\textbf{Meta-Knowledge Acquisition}}\label{subsubsec:meta_knowledge_acquisition}
The upper-left branch of Fig.~\ref{fig:framework} shows the meta-knowledge acquisition process.
The goal is to learn a variational task inference module and a corresponding high-level policy on the simplified agent introduced in Section \ref{subsubsec:simplified_agent_modeling}, so that the resulting task-level knowledge can later be reused by complex agents.
For each training task \(\mathcal{T}_i \sim p(\mathcal{T})\), the simplified agent interacts with the environment and collects trajectories using the current inference model and high-level policy.
The collected data are used to jointly learn a latent task representation, a non-parametric task prior, and a high-level policy for the simplified agent.

At each time step \(t\), the task inference module receives a context window
%
%
\(\mathbf{c}_t = \{(\mathbf{s}_{t-\tau}, r_{t-\tau}, \mathbf{s}_{t-\tau+1})\}_{\tau=0}^{H-1}\), where \(H\) is the context length. Unlike previous work \cite{rakelly2019efficient} that uses the entire trajectory as context, our framework assumes that sufficient context can be inferred from a finite window of recent history. 
Therefore, we use a context length of \(H = 48\) to improve sample efficiency.
This design encourages the latent representation to capture global task-induced transition and reward structures.

The task inference module is implemented as a VAE variant with a Bayesian non-parametric prior.
The encoder \(q_{\phi}(z_t|\mathbf{c}_t)\) maps the context window to a latent task variable \(z_t \in \mathbb{R}^{d_z}\):
\begin{equation}\label{eq:encoder_distribution}
    q_{\phi}(z_t|\mathbf{c}_t) = \mathcal{N} \left( z_t; \boldsymbol{\mu}_{\phi}(\mathbf{c}_t), \mathrm{diag}(\boldsymbol{\sigma}_{\phi}^{2}(\mathbf{c}_t)) \right)\text{,}
\end{equation}
where the encoder is implemented using a sequence model (In our case GRU module).
The latent variable is sampled using the reparameterization trick \cite{kingma2013auto}.
%
The decoder \(p_{\xi}(\mathbf{s}_{t+1}, r_t, y_t \mid z_t, a_t, \mathbf{s}_t)\) reconstructs the next state, reward, and task confidence from the latent variable and the current state. 
By jointly predicting the transition, the reward, and the task category, the decoder provides complementary supervision signals that encourage \(z_t\) to encode task-level information.
\begin{equation}\label{eq:decoder_prediction}
    (\hat{\mathbf{s}}_{t+1}, \hat{r}_t, \boldsymbol{\ell}_t) = p_{\xi}(z_t, a^{\mathrm{high}}_t, \mathbf{s}_t)\text{,} \qquad \hat{\mathbf{y}}_t = \mathrm{softmax}(\boldsymbol{\ell}_t)\text{,}
\end{equation}
where \(\boldsymbol{\ell}_t \in \mathbb{R}^{K_y}\) denotes the task-prediction logits, \(\hat{\mathbf{y}}_t\) is the predicted task-confidence distribution, \(K_y\) is the number of task categories, and \(a^{\mathrm{high}}_t\) is the high-level action from simplified agent with minimum dynamic complexity.
We define the decoder reconstruction losses as
\begin{equation}\label{eq:decoder_losses}
\begin{aligned}
    \mathcal{L}_{\mathrm{state}} &= \mathbb{E}_{z_t \sim q_{\phi}(\cdot|\mathbf{c}_t)} \left[ \left\| \mathbf{s}_{t+1}-\hat{\mathbf{s}}_{t+1} \right\|_2^2 \right]\text{,} \\
    \mathcal{L}_{\mathrm{reward}} &= \mathbb{E}_{z_t \sim q_{\phi}(\cdot|\mathbf{c}_t)} \left[ \left\| r_t-\hat{r}_t \right\|_2^2 \right]\text{,} \\
    \mathcal{L}_{\mathrm{task}} &= -\mathbb{E}_{z_t \sim q_{\phi}(\cdot|\mathbf{c}_t)} \left[ \sum_{k=1}^{K_y} y_{t,k} \log \hat{y}_{t,k} \right]\text{,}
\end{aligned}
\end{equation}
where \(y_t\) denotes the ground-truth task label, represented as a one-hot vector. The overall decoder reconstruction objective is then
\begin{equation}\label{eq:rec_loss}
    \mathcal{L}_{\mathrm{rec}} = \lambda_s\mathcal{L}_{\mathrm{state}} + \lambda_r\mathcal{L}_{\mathrm{reward}} + \lambda_y\mathcal{L}_{\mathrm{task}},
\end{equation}
where \(\lambda_s, \lambda_r, \lambda_y\) are factors that balance the reconstruction weighting.

To model non-parametric task structures, we impose a DPMM prior \(p_{\theta}(z)\) on the latent space.
This prior encourages similar tasks to form compact clusters and allows new task modes to emerge when observed trajectories cannot be explained by existing components.
In our framework, we asynchronously update the neural inference model and the DPMM prior iteratively.
When updating the neural inference model, the DPMM prior is kept fixed and serves as a structured latent regularizer.
When updating the DPMM, the encoder is fixed and the latent samples generated from recent context windows are treated as observations for the mixture model.
Specifically, given a latent buffer \(\mathcal{Z}=\{z_n\}_{n=1}^{N_z}\), we update the local component responsibilities and global variational parameters according to Eq.~\eqref{eq:local_update_dpmm} and Eq.~\eqref{eq:global_update_dpmm}.
For mini-batch training, we further follow the MemoVB update in 
Sec.~\ref{subsec:vi_dpmm},
which updates the full-dataset sufficient statistics by replacing the old cached statistics of the current batch with the newly computed ones.
This allows the DPMM prior to track the evolving latent space without refitting the mixture model from scratch.
In addition, we use birth and merge moves to adapt the number of active components: birth moves introduce new components for latent samples that are poorly explained by the current mixture, while merge moves remove redundant components when the full-data ELBO (Eq.~\ref{eq:elbo_expanded}) improves.
After the DPMM update, its parameters are fixed again and used to regularize the next neural update.
To avoid unstable hard assignments, we compute the latent regularization using a soft component-wise KL term.
Specifically, the KL regularization is approximated as a responsibility-weighted average between the encoder posterior and the DPMM components.
To summarize, the task inference objective is
\begin{equation}\label{eq:inference_loss}
    \mathcal{L}_{\mathcal{I}} = \mathcal{L}_{\mathrm{rec}} + \beta_{\mathrm{KL}} 
    \sum_{k} \rho_{tk} D_{\mathrm{KL}} \left( q_{\phi}(z_t|\mathbf{c}_t) \| p_{\theta_k}(z_t) \right)
    \text{,}
\end{equation}
where \(\beta_{KL}\) balances reconstruction fidelity and latent prior regularization, \(\rho_{tk}\) denotes the responsibility of component \(k\) for \(z_t\), and \(p_{\theta_k}(z_t)\) is the \(k\)-th DPMM component.
This soft weighting reduces the training noise caused by assigning a latent sample to an incorrect component.

As shown in Fig.~\ref{fig:framework}, the inferred latent variable \(z_t\) is then used to condition the high-level policy.
The policy operates on the augmented task state \((\mathbf{s}_t,z_t)\), where \(\mathbf{s}_t\) denotes the task-relevant state of the simplified agent.
It outputs a continuous high-level action
\begin{equation}
    a_t^{\mathrm{high}}
    \sim
    \pi_{\mathrm{high}}
    \left(
    \cdot \mid \mathbf{s}_t, z_t
    \right)\text{,}
    \label{eq:high_level_policy}
\end{equation}
where \(a_t^{\mathrm{high}}\) is an abstract action in the simplified state space.
Since the simplified agent is defined over task-level variables, this high-level action can be used to generate task-aligned subgoals, such as a desired intermediate position or velocity.

The high-level policy is trained with an entropy-regularized actor-critic objective \cite{haarnoja2018soft}.
To prevent the policy update from modifying the task-inference module, the latent task variable \(z_t\) is detached before being used by the high-level policy: \(\bar{z}_t = \mathrm{sg}(z_t)\),
%
where \(\mathrm{sg}(\cdot)\) denotes the stop-gradient operation.
We use two high-level critics \(Q_{\chi_1}\) and \(Q_{\chi_2}\), which are trained with the standard clipped double-Q soft Bellman objective. 
The high-level actor is updated by
\begin{equation}
\begin{aligned}
    \mathcal{L}_{\pi}^{\mathrm{high}} = 
    \mathbb{E}_{a_t^{\mathrm{high}}\sim\pi_{\mathrm{high}}} \Big[
    &\eta \log \pi_{\mathrm{high}}
    \left(a_t^{\mathrm{high}} \mid \mathbf{s}_t,\bar{z}_t \right) \\
    &-\min_{j=1,2} Q_{\chi_j}
    \left(\mathbf{s}_t,\bar{z}_t,a_t^{\mathrm{high}}\right)
    \Big]\text{,}
\end{aligned}
\label{eq:high_actor_loss}
\end{equation}
where \(a_t^{\mathrm{high}}\) is sampled from the high-level policy, \(\eta\) is the entropy coefficient, and the minimum over the two critics mitigates overestimation bias.
The actor and critics are optimized with their respective losses, and the target critics are updated by Polyak averaging. 
Since \(\bar{z}_t\) is detached during policy optimization, gradients from the high-level actor-critic objective are not propagated to the encoder, decoder, or DPMM prior. 
These modules are therefore updated only by the reconstruction and latent-structure objectives in Eq.~\eqref{eq:inference_loss}. 
This design reduces the influence of control-specific optimization on the latent space and encourages \(z_t\) to encode task-level information from transitions and rewards.

After convergence, the learned task inference module and high-level policy form the reusable meta-knowledge module:
\begin{equation}
    \mathcal{M}_{\mathrm{meta}} = (\mathcal{I}, \pi_{\mathrm{high}})\text{,} \qquad \mathcal{I}=(q_{\phi},p_{\xi})\text{,}
    \label{eq:meta_module}
\end{equation}
where \(q_{\phi}\) infers latent task variables, \(p_{\xi}\) reconstructs task-relevant transition and reward information, and \(p_{\theta}\) provides the DPMM task prior.
This module is frozen after training and reused during cross-embodiment deployment.
\subsubsection{\textbf{Magnitude-triggered Disentangled Learning}}\label{subsubsec:lowlevelpretrain}
In parallel with meta-knowledge acquisition, the upper-right branch of Fig.~\ref{fig:framework} trains an embodiment-specific low-level policy for each complex agent.
This learning branch is independent of the task inference module and the high-level policy.
Its goal is not to learn task semantics, but to provide each target agent with basic behaviors, stable exploration, and the ability to adapt its motion according to task-level subgoal magnitudes.
This is necessary because complex agents may fail to perform meaningful exploration if they cannot first maintain stable movement.
We leverage Soft Actor-Critic (SAC)~\cite{haarnoja2018soft} as the optimization backbone for training the low-level policy.
However, instead of applying vanilla SAC directly, we introduce two modifications to make the low-level controller suitable for cross-agent meta-knowledge reuse.
First, we introduce a \emph{Semantic-Magnitude Alignment Interface} (SMAI), which can later bridge the task-level guidance from the trained meta-knowledge module \(\mathcal{M}_{\mathrm{meta}}\) to the low-level controller through magnitude signals.
This interface does not require the low-level policy to imitate a fixed high-level trajectory.
Instead, it enables the agent to adapt its action pattern according to the magnitude and direction of the issued task-level command.
Second, we use a curriculum-based adaptation schedule in a non-stationary Meta-RL setting, where the Low-level policies must rapidly and dynamically adapt their behavior within a single episode in response to the magnitude signal, while preserving sufficient exploration under non-stationary conditions.
The curriculum gradually increases the target magnitude, switching frequency, trajectory horizon, and initial-state diversity within a rollout episode, so that the low-level policy first learns stable locomotion and then acquires the ability to adjust its behavior in response to changing subgoal magnitudes.

For each target agent \(i\), we train a low-level controller as a magnitude-guided action policy:
\begin{equation}\label{eq:low_level_policy}
    a_t^{(i)} \sim \pi_{\mathrm{low}}^{(i)} \left( \cdot \mid s_t^{(i)}, g_t \right)\text{,}
\end{equation}
where \(s_t^{(i)}\) is the state of complex agent \(i\), \(a_t^{(i)}\) is the motor action, and \(g_t\) is the task-level guidance signal provided through the SMAI.
The role of SMAI is to represent a task-level command by coupling its semantic guidance channel with its scalar magnitude. 
Let \(u_t \in \{1,\ldots,K_g\}\) denote the active guidance channel, and let \(\mathbf{b}_t=\mathbf{e}(u_t)\in\{0,1\}^{K_g}\) be its channel selector. 
Given a scalar magnitude \(m_t\in\mathbb{R}\), the SMAI signal is defined as
\begin{equation}\label{eq:smai_signal}
    g_t = m_t \mathbf{b}_t \in \mathbb{R}^{K_g}\text{,}
\end{equation}
where \(K_g\) is the number of task-level guidance channels. 
The active channel specifies the semantic type of the command, while the value assigned to that channel specifies the corresponding magnitude.
Importantly, \(g_t\) does not contain morphology-dependent quantities such as joint torques, contact forces, or gait patterns.
Therefore, the same SMAI signal can be shared across different agents, while each low-level controller learns how to adapt its motion scale and action pattern according to its own dynamics.

During low-level policy warm-up, the guidance signal is not generated by the meta-knowledge module. 
Instead, its semantic channel and magnitude are sampled from a curriculum-dependent task-level distribution:
\begin{equation}
    (u_t,m_t) \sim p_e(\mathcal{U},\mathcal{M}), 
    \qquad
    g_t = m_t \mathbf{e}(u_t)\text{,}
    \label{eq:subgoal_sampling}
\end{equation}
where \(p_e\) denotes the curriculum-dependent distribution over task-level guidance channels and their associated magnitudes at training epoch \(e\). \(\mathcal{U}\) denotes the set of semantic guidance channels in the SMAI representation, and \(\mathcal{M}\) denotes the scalar magnitude space.

The low-level reward encourages the agent to match the task-relevant quantity associated with the active SMAI channel, while regularizing excessive control effort:
\begin{equation}\label{eq:low_level_reward}
    r_{\mathrm{low}}^{(i)} = - \left\| \Phi_i^{u_t}(s_t^{(i)}) - m_t \right\|_2^2 - \lambda_a \left\| a_t^{(i)} \right\|_2^2 \text{,}
\end{equation}
where \(\Phi_i^{u_t}(\cdot)\) extracts the task-relevant quantity of agent \(i\) associated with the active guidance channel \(u_t\), such as center-of-mass position for goal-reaching commands or center-of-mass velocity for velocity-tracking commands, and \(\lambda_a\) controls the action regularization term.

To stabilize SMAI-conditioned low-level policy learning under the non-stationary setting, we introduce a curriculum-based adaptation schedule that progressively increases the difficulty of the training distribution. 
The curriculum jointly controls the commanded target magnitude, within-trajectory task-switching frequency, rollout horizon, and initial-state diversity. 
Given a sampled guidance channel \(u_t\) and its target magnitude \(m_t^{*}\), the curriculum first adjusts the commanded magnitude according to the task type:
\begin{equation}
    \tilde{m}_t =
    \left\{
    \begin{aligned}
        \alpha_e m_t^{*}, & \quad u_t \in \mathcal{U}_{\mathrm{vel}}\text{,} \\
        m_t^{*},          & \quad u_t \in \mathcal{U}_{\mathrm{goal}}\text{,}
    \end{aligned}
    \right.
    \qquad
    \alpha_e =
    \left\{
    \begin{aligned}
        \frac{1}{2}, & \quad e < E_{\mathrm{vel}}\text{,} \\
        1,           & \quad e \ge E_{\mathrm{vel}}\text{,}
    \end{aligned}
    \right.
    \label{eq:curriculum_magnitude_scaling}
\end{equation}
where \(\mathcal{U}_{\mathrm{vel}}\) and \(\mathcal{U}_{\mathrm{goal}}\) denote the velocity-tracking and goal-reaching guidance channels, respectively. \(e\) is the current training epoch, and \(E_{\mathrm{vel}}\) is the epoch after which the full velocity range is enabled. Thus, velocity-tracking tasks are initially trained with reduced target magnitudes to avoid unstable cold-start exploration, whereas goal-tracking tasks preserve their original target specification and are made progressively harder through the other curriculum factors.

The temporal complexity of rollouts is increased by gradually introducing more frequent task changes. 
Let \(C(e)\) denote the prescribed number of command updates within one rollout at epoch \(e\), and let \(H(e)\) denote the current rollout horizon. 
When \(C(e)>0\), the command update interval is defined as
\begin{equation}
    \Delta(e) = \left\lfloor \frac{H(e)}{C(e)} \right\rfloor\text{.}
    \label{eq:curriculum_switch_interval}
\end{equation}
At the beginning of a rollout and every \(\Delta(e)\) steps thereafter, the active semantic guidance channel and its target magnitude are resampled from the curriculum-dependent distribution:
\begin{equation}
    (u_{\tau},m_{\tau}^{*}) \sim p_e(\mathcal{U},\mathcal{M}), \quad \text{if } \tau \bmod \Delta(e)=0 \text{,}
    \label{eq:curriculum_task_switch}
\end{equation}
where \(\tau \in \{0,\ldots, H(e)-1\}\) is the time index within the current rollout.
Between two consecutive command-update steps, the sampled pair \((u_{\tau},m_{\tau}^{*})\) is kept unchanged, so that the low-level controller has sufficient time to respond to the current task-level command. 
The sampled magnitude \(m_{\tau}^{*}\) is then adjusted according to the curriculum schedule in Eq.~\eqref{eq:curriculum_magnitude_scaling}, and the resulting commanded magnitude is embedded into the active semantic channel through Eq.~\eqref{eq:subgoal_sampling}. 
This produces the executed SMAI signal \(\tilde{g}_{\tau}\) that is provided to the low-level policy.
Early stages use no or few within-trajectory command updates, allowing the controller to first acquire stable tracking of a single guidance signal. 
As training progresses, increasing \(C(e)\) exposes the controller to more frequent transitions between semantic channels and target magnitudes, forcing it to recover from changing task-level commands and adapt online without losing stability.
Meanwhile, the rollout horizon is progressively enlarged as
\begin{equation}
    H(e) = H_k, \quad \text{if } E_{k-1} \le e < E_k, \quad k = 1,\ldots,K\text{,}
    \label{eq:curriculum_horizon}
\end{equation}
where \(0=E_0 < E_1 < \cdots < E_{K-1} < E_K\) denote the curriculum transition epochs, and \(H_1 < H_2 < \cdots < H_K\).
Thus, the rollout horizon is explicitly increased as training progresses: early stages use short horizons to facilitate stabilization and error recovery, whereas later stages require the controller to maintain stable behavior over longer trajectories.
Finally, the reset distribution is widened after a predefined curriculum stage:
\begin{equation}
    s_0 \sim
    \left\{
    \begin{aligned}
        p_{\mathrm{default}}(s_0), & \quad e < E_{\mathrm{rand}} \text{,} \\
        p_{\mathrm{rand}}(s_0),    & \quad e \ge E_{\mathrm{rand}} \text{,}
    \end{aligned}
    \right.
    \label{eq:curriculum_reset}
\end{equation}
where \(p_{\mathrm{rand}}\) randomizes task-relevant initial quantities such as position and velocity. 
In addition to this staged schedule, we employ a performance-triggered task switch: when the task-relevant quantity associated with the active guidance channel,
\(\Phi_i^{u_t}(s_t^{(i)})\), is sufficiently close to the current commanded magnitude \(\tilde{m}_t\), a new subgoal is sampled with a fixed probability. 
This prevents the agent from remaining in already solved easy conditions and encourages continuous adaptation to new subgoals. 
Overall, the proposed curriculum first enables the controller to learn stable locomotion under simple targets and then progressively exposes it to larger velocity commands, more frequent task transitions, longer horizons, and more diverse initial states. 
Consequently, the warmed-up low-level controller learns not only reliable embodiment-specific motion generation, but also stable translation of task-level magnitude guidance signal from the frozen meta-knowledge module into executable actions.

\subsection{Meta-Knowledge Reutilization}\label{subsec:metareuse}

After meta-knowledge acquisition, the task inference module and the high-level policy are frozen and reused for agents with heterogeneous embodiments. 
As illustrated in Fig.~\ref{fig:framework} bottom, deployment for meta-knowledge reuse follows a closed-loop procedure in which the target agent first maps its raw embodiment-specific state into the shared abstract state space, then infers the task context using the frozen inference module, generates a task-level magnitude signal through the frozen high-level policy, and finally executes the resulting structured subgoal through its own frozen low-level controller. 
In this process, the reusable components only determine \emph{what} task-level behavior should be achieved, whereas embodiment-specific execution is handled by the low-level policy and a lightweight temporal adaptor.

\subsubsection{\textbf{Meta-Knowledge Reuse and Magnitude guidance Generation}}

For a target agent \(i\), let \(s_t^{(i)}\) denote its full state at time step \(t\). 
Since the meta-knowledge module is trained in the simplified agent space, the raw state is first projected into a shared task-relevant representation:
\begin{equation}
    \bar{s}_t^{(i)} = \mathcal{R}^{(i)}(s_t^{(i)}),
    \label{eq:abstract_state_remap}
\end{equation}
where \(\mathcal{R}^{(i)}\) is an agent-specific abstract state remapping function. 
It removes morphology-dependent quantities, such as joint torques, actuator states, and contact-specific variables, while retaining task-aligned variables such as position, velocity, and orientation. 
Based on this representation, we construct a recent interaction context
\begin{equation}
    \bar{c}_t^{(i)} = \left\{ \left( \bar{s}_{t-\tau}^{(i)}, r_{t-\tau}^{(i)},  \bar{s}_{t-\tau+1}^{(i)} \right) \right\}_{\tau=0}^{H-1}\text{,}
    \label{eq:deployment_context}
\end{equation}
which has the same semantic structure as the context used during meta-knowledge acquisition. 
The frozen inference module then produces a latent task representation
\begin{equation}
    z_t^{(i)} \sim q_{\phi} \left( z \mid \bar{c}_t^{(i)} \right)\text{,}
    \label{eq:deployment_latent}
\end{equation}
where \(z_t^{(i)}\), as described in Sec.~\ref{subsubsec:meta_knowledge_acquisition}, captures the current task semantics in an embodiment-independent manner. 
The DPMM prior associated with the inference module provides the latent task-mode structure, while the task prediction head provides a semantic task-confidence vector \(\hat{\mathbf{y}}_t^{(i)}\).

Conditioned on the remapped abstract state and the inferred latent task variable, the frozen high-level policy generates an abstract high-level action:
\begin{equation}
    a_t^{\mathrm{high},(i)} \sim \pi_{\mathrm{high}} \left( \cdot \mid \bar{s}_t^{(i)}, z_t^{(i)} \right)\text{.}
    \label{eq:deployment_high_action}
\end{equation}
This action is not directly used as the final subgoal magnitude. 
Directly interpreting the high-level action as a subgoal may produce overly aggressive or overly conservative commands, especially when transferred to agents with different dynamics. 
Instead, we use it as a supervisory signal that controls the progression rate of the generated subgoal magnitude. 
Specifically, the high-level action with values in \([-1,1]\) is first transformed into a normalized progression coefficient:
\begin{equation}
    \alpha_t^{(i)} = \frac{1}{2} \left( a_t^{\mathrm{high},(i)} + 1 \right)\text{,} \qquad \alpha_t^{(i)} \in [0,1]\text{.}
    \label{eq:progression_coefficient}
\end{equation}
Given the target magnitude \(m_{\star,t}^{(i)}\) in the shared task space and the previously issued magnitude signal \(m_{t-1}^{(i)}\), the new scalar magnitude is generated through a residual update:
\begin{equation}
    m_t^{(i)} = m_{t-1}^{(i)} + \alpha_t^{(i)} \left( m_{\star,t}^{(i)} - m_{t-1}^{(i)} \right)\text{,}
    \label{eq:action_guided_magnitude}
\end{equation}
where \(m_t^{(i)}\) may correspond to a desired position magnitude, velocity magnitude, or other task-dependent scalar command. 
For velocity-tracking tasks, the generated magnitude is further constrained within the valid velocity range. 
This residual formulation has an adaptive contraction effect: when the current magnitude is far from the target, the high-level action can induce a large progression step; as the magnitude approaches the target, the residual term naturally decreases, preventing abrupt or oscillatory guidance.

The final task-level subgoal provided to the low-level controller is constructed by embedding the generated magnitude into the semantic channel indicated by the predicted task confidence. 
Specifically, the task prediction \(\hat{\mathbf{y}}_t^{(i)}\) is first converted into a semantic selector
\begin{equation}
    \hat{u}_t^{(i)} = \arg\max_k \hat{y}_{t,k}^{(i)}\text{,}
    \qquad
    \mathbf{b}_t^{(i)} = \mathrm{onehot} \left( \hat{u}_t^{(i)} \right) \in \{0,1\}^{K_g}\text{,}
    \label{eq:semantic_selector}
\end{equation}
%
The generated scalar magnitude \(m_t^{(i)}\) is then injected into the selected semantic channel, so that the deployment-time subgoal follows the same SMAI representation as defined in Eq.~\ref{eq:smai_signal}.
%
%
In this representation, the active channel specifies the semantic type of the subgoal, while its value specifies the corresponding task-level magnitude. 
This design provides the low-level controller with a compact and unified guidance signal, where task identity and quantitative command strength are encoded in the same task-aligned vector space.

\subsubsection{\textbf{Stride Prediction for Temporal Alignment}}

Directly updating \(g_t^{(i)}\) at every low-level control step can be unstable for complex agents because of inertia, contact dynamics, actuation delay, and morphology-dependent response time. 
To compensate for this mismatch, we introduce an agent-specific stride predictor \(\mathcal{P}_{\zeta}^{(i)}\), where \(\zeta\) denotes its parameters. 
Given the current subgoal, the stride predictor estimates an execution stride \(k_t^{(i)} \in \mathbb{Z}^{+}\), which specifies how many low-level control steps the current subgoal should be held fixed.

After executing the low-level policy for \(k_t^{(i)}\) steps, the achieved state is compared with the issued magnitude signal in the task-relevant dimension associated with the active semantic channel. 
The stride predictor is updated online by minimizing
\begin{equation}
    \mathcal{L}_{\mathcal{P}}^{(i)}(\zeta) = \left\| \Phi_i^{\hat{u}_t^{(i)}} \left( s_{t+k_t^{(i)}}^{(i)} \right) - m_t^{(i)} \right\|_2^2\text{,}
    \label{eq:stride_prediction_loss}
\end{equation}
where \(\Phi_i^{\hat{u}_t^{(i)}}(\cdot)\) extracts the controlled task-relevant quantity of agent \(i\) corresponding to the active semantic channel \(\hat{u}_t^{(i)}\), such as its center-of-mass position or velocity. 
The distinction between \(\mathcal{R}^{(i)}\) and \(\Phi_i^{\hat{u}_t^{(i)}}\) is important: \(\mathcal{R}^{(i)}\) maps raw states into the abstract state space used for task inference, whereas \(\Phi_i^{\hat{u}_t^{(i)}}\) evaluates how well the physical agent has achieved the issued magnitude signal under the selected task channel. 
This temporal adaptation operates only at the lightweight stride predictor and does not update the frozen inference module, high-level policy, or low-level policies.

\subsubsection{\textbf{Policy Adaptive Deployment and Abstract State Remapping}}

Given \(g_t^{(i)}\) and the predicted \(k_t^{(i)}\), the agent-specific low-level policy executes the task-level guidance signal over multiple control steps:
\begin{equation}
    a_{\tau}^{(i)} \sim \pi_{\mathrm{low}}^{(i)} \left( \cdot \mid s_{\tau}^{(i)}, g_t^{(i)} \right), \qquad \tau = t,\ldots,t+k_t^{(i)}-1 \text{.}
    \label{eq:deployment_low_policy}
\end{equation}
The low-level policy translates the abstract task-level command into executable motor actions under the dynamics of agent \(i\). 
After execution, the generated transition is mapped back to the abstract state space through \(\mathcal{R}^{(i)}\). 
Specifically, \(\mathcal{R}^{(i)}\) extracts global task-relevant quantities from the complex agent, such as the position and velocity of the body center of mass, while filtering out embodiment-specific joint states, actuator variables, and contact-related details. 
This observation mapping aligns the deployed agent's state representation with the state space used by the high-level encoder, allowing the frozen inference module to process interaction contexts from heterogeneous agents in a consistent abstract space. 
The resulting interaction tuple is then appended to the context window for the next inference step. 
This closes the loop between abstract task inference, high-level magnitude generation, temporal alignment, and embodiment-specific control.
During deployment, the frozen meta-knowledge module remains unchanged. Online adaptation is restricted to the lightweight stride predictor. 
This separation preserves the embodiment-independent nature of the learned task representation while allowing each agent to adapt its physical execution to the generated subgoals. 
Consequently, the same meta-knowledge module can be reused across heterogeneous agents: the inference module and high-level policy operate in the shared abstract task space, whereas \(\pi_{\mathrm{low}}^{(i)}\), \(\mathcal{P}_{\zeta}^{(i)}\), \(\mathcal{R}^{(i)}\), and \(\Phi_i\) absorb the morphology- and dynamics-specific differences.
In summary, the overall learning and inference procedures of our framework are provided in Algs. 1 and 2 of Supplementary Sec. A.
\section{EXPERIMENTS}

\subsection{Experimental Setup}\label{subsec:experiment_setup}

\textbf{Embodiment}. We evaluate the proposed framework in a cross-embodiment Meta-RL setting.
The goal of this experiment is to examine whether the meta-knowledge learned from the simplified agent can be reused across agents with different embodiments.
Specifically, we consider the Mujoco-based task suite \cite{todorov2012mujoco} with four embodiments: \textit{Hopper}, \textit{Walker}, \textit{Half-Cheetah}, and \textit{Ant}.
As shown in Fig.~\ref{fig:cross_agent_tasks}, these agents exhibit distinct morphologies and dynamics, including single-leg hopping, bipedal walking, planar quadrupedal-like running, and full quadrupedal locomotion.
This diversity provides a challenging benchmark for evaluating whether the learned meta-knowledge is independent of embodiment-specific control dynamics.
%
\input{img_tex/env_setup}

\textbf{Task configuration}. For each agent, we evaluate four non-parametric locomotion tasks:
\textit{Velocity Forward}, \textit{Velocity Backward}, \textit{Goal Forward}, and \textit{Goal Backward}.
The velocity-tracking tasks require the agent to match a target horizontal velocity, while the goal-reaching tasks require the agent to move toward a target horizontal position.
Table~\ref{tab:low_level_task_config} summarizes the task parameter configuration.
For the velocity-tracking tasks, the target velocity is sampled from $[1,3]$ for forward running and from $[-3,-1]$ for backward running.
For the goal-reaching tasks, the target position is sampled from $[0,10]$ for goals in front and from $[-10,0]$ for goals behind the agent.
To preserve the Meta-RL setting, task instances for the simplified-agent environment and those for each complex agent are independently sampled from the corresponding task distributions. 
This avoids a complete overlap between the reusable meta-knowledge acquired on the simplified agent and the task experience observed by complex agents during disentangled learning.
Additionally, the reward is defined as the negative \(\ell_1\) distance between the target state and the current task-relevant state.
Following the reward consistency assumption in Sec.~\ref{subsec:preliminary}, this unified reward formulation is shared across environments, as it measures the shared task-level progress while leaving each agent to learn its own policy under embodiment-specific dynamics.
\begin{table}[ht!]
\centering
\caption{Parameter configurations of the four non-parametric locomotion tasks}
\vskip -.05in
\label{tab:low_level_task_config}
\resizebox{\linewidth}{!}{
\begin{tabular}{lcccc}
\toprule
\rowcolor{nature_tab_gray1}
\textbf{Base Task} & \textbf{Target} & \textbf{Unit} & \textbf{Range} & \textbf{Reward} \\
\midrule
Velocity Forward  & $v_x$ & $m/s$& $1 \leq v_x^* \leq 3$ & $-|v_x^* - v_x|$ \\

\midrule
\rowcolor{nature_tab_gray2}
Velocity Backward & $v_{-x}$ & $m/s$& $-3 \leq v_{-x}^* \leq -1$ & $-|v_{-x}^* - v_{-x}|$ \\

\midrule
Goal Forward      & $p_x$& $m$& $0 < p_x^* \leq 10$ & $-|p_x^* - p_x|$ \\

\midrule
\rowcolor{nature_tab_gray2}
Goal Backward     & $p_{-x}$ & $m$& $-10 \leq p_{-x}^* < 0$ & $-|p_{-x}^* - p_{-x}|$ \\
\bottomrule
\end{tabular}
}
\end{table}

\textbf{Baselines}.
We compare ReMAP with four representative Meta-RL baselines that cover recurrent, probabilistic context-based, mixture-based, and Bayesian non-parametric task inference paradigms:
\begin{itemize}
    \item \(\bm{\mathrm{RL}^{2}}\)~\cite{duan2016rl}: a recurrence-based Meta-RL method that encodes the adaptation process in the hidden state of a recurrent policy, allowing the agent to condition its behavior on previous observations, actions, rewards, and terminations.
    
    \item \textbf{PEARL}~\cite{rakelly2019efficient}: an off-policy context-based Meta-RL method that performs probabilistic task inference from recent experience and conditions the actor-critic policy on the inferred latent task variable.
    
    \item \textbf{CEMRL}~\cite{bing2022meta}: a context-based Meta-RL method that introduces a Gaussian Mixture Model (GMM) latent representation to model clustered task distributions and improve adaptation in non-stationary and dynamic environments.
    
    \item \textbf{MELTS}~\cite{10495171}: a Bayesian non-parametric context-based Meta-RL method that combines DPMM-based task self-discovery with a VAE-style inference model to identify non-parametric task structures in a self-adaptive manner.
\end{itemize}
For all Meta-RL baselines, the training and testing task sets are sampled independently from the corresponding task distributions to avoid complete overlap between meta-training and meta-testing tasks.
These baselines represent strong state-of-the-art Meta-RL methods for task inference and adaptation. 
However, recent methods are trained in an end-to-end manner under a specific agent embodiment, where task inference and control are optimized jointly. 
Once the embodiment changes, the task representation and policy must be trained from scratch or substantially adapted for the new dynamics. 
This embodiment-specific coupling greatly increases the training cost and limits the transferability of the learned meta-knowledge. 
In contrast, ReMAP learns the task inference module and high-level policy once on a dynamics-simplified agent and reuses them across target agents. 
Therefore, only the embodiment-specific execution layer needs to be prepared for each agent, which improves cross-embodiment reutilization and transfer efficiency.

In addition, we conduct ablation studies to evaluate the contribution of the Bayesian non-parametric prior used in ReMAP. 
We compare the ReMAP full model with three prior variants: 
(i) a single Gaussian prior, which assumes that all task embeddings are generated from one shared latent distribution; 
(ii) a Gaussian Mixture Model prior with \(K=2\), which introduces multiple latent components but relies on a predefined fixed and potentially improper assumption about the number of task modes; and 
(iii) a stick-breaking prior, which provides a Bayesian non-parametric formulation with a theoretically unbounded number of components. 
These variants are used to examine whether the DPMM-based latent organization is necessary for learning compact, semantically separable, and reusable task representations.

\input{img_tex/toy_results}
\subsection{Meta-Knowledge Acquisition}

Before transferring the learned meta-knowledge to complex agents, we first evaluate whether the task inference module and the high-level policy trained on the simplified agent can acquire executable, semantically structured task knowledge. 
The purpose of this stage is therefore to verify the acquired meta-knowledge from three aspects. 
First, we evaluate whether the optimization process of the task inference module and the high-level policy converges during training. 
Second, after convergence, we deploy the trained modules and validate whether they support continuous, reasonable task inference sufficient to guide downstream policies. 
Third, we analyze the latent space to determine whether the learned representation contains distinguishable task semantics rather than merely encoding trajectory-level variations.


Fig.~\ref{fig:toy_multi_results}a shows the training reward curve of the simplified agent in the given tasks. 
The reward increases steadily during training and gradually converges to a stable level at -50. 
This result indicates that the task inference module and the high-level policy reach a stable optimization state under the simplified environment. 
Since the agent provides an abstraction with simplified embodiment dynamics, the convergence behavior suggests that the framework can consistently learn task-related behavioral patterns without being dominated by complex low-level control effects. 
However, reward alone does not reflect whether the learned latent representation contains informative task semantics. 


To further validate, we deploy the trained task inference module and high-level policy on the simplified agent to evaluate whether the learned meta knowledge can actually guide task execution. 
Fig.~\ref{fig:toy_multi_results}b shows example tracking trajectories under different task settings. 
The dashed lines denote the desired task specifications, while the solid lines represent the actual executed trajectories. 
The simplified agent continuously adjusts its position and velocity according to the inferred task representation, and the resulting trajectories closely follow the corresponding task targets. 
This result demonstrates that the learned latent representation contains actionable task-level semantic information rather than coupled latent variations, as in a single Gaussian latent space. 
Moreover, the high-level policy can successfully translate the inferred task semantics into executable subgoals that guide downstream locomotion behaviors. The close alignment between the executed trajectories and the desired task specifications indicates that the acquired meta knowledge is not only semantically meaningful but also practically usable for hierarchical task execution.

\input{img_tex/all_agents_reward_multi.tex}

To further analyze whether the learned meta-knowledge contains clear task semantics, we visualize the inferred latent variables using t-SNE projection. 
As shown in Fig.~\ref{fig:toy_multi_results}c, the latent space forms four clearly separated clusters corresponding to different task types. 
The strong separation between clusters indicates that our DPMM-regularized variational encoding process can organize the latent representation according to underlying task semantics. 
Task inference results with similar objectives are clustered together, while tasks with different semantic meanings remain distinguishable in the latent space. 
This structured latent organization demonstrates that the learned meta-knowledge captures task-level information that generalizes beyond the simplified environment. 
Overall, these results answer Q1 by demonstrating that the proposed method learns task-discriminative representations for non-parametric task semantics. This property is essential for downstream deployment, since it enables the reused inference module to provide semantic guidance for generating appropriate high-level subgoals across different embodiments.

\subsection{Disentangled Learning}
In parallel with meta-knowledge acquisition, we train low-level policies for complex agents through magnitude-triggered SMAI conditioning. 
This experiment addresses Q2 by testing whether task inference can be separated from embodiment-specific control. 
End-to-end Meta-RL suffers from a chicken-and-egg problem: task inference requires informative trajectories, but such trajectories can only be collected when the policy already performs meaningful exploration. 
For complex locomotion agents, early exploration often fails due to unstable motion, making this coupling especially inefficient. 
Our disentangled learning strategy therefore trains low-level controllers independently from the inference module and high-level policy. 
With SMAI and the curriculum schedule, each controller first learns stable embodiment-specific motion and then acquires the ability to follow changing magnitude commands, without entangling task semantics with low-level control.

Fig.~\ref{fig:four_agents_reward} shows the training reward curves of the low-level controllers for four embodiments: Hopper, Walker, Cheetah, and Ant. 
Under the proposed curriculum, convergence indicates that the corresponding low-level policy has acquired both stable locomotion and subgoal-magnitude-following capability under its own dynamics and action space. 
Different embodiments exhibit substantially different learning difficulties due to their morphology and stability characteristics. 
Cheetah and Ant have relatively stable support structures during locomotion, which allows them to maintain balance more easily during exploration and converge within about 300 epochs. 
In contrast, Walker is a bipedal agent and Hopper relies on a single support leg, making balance control considerably more difficult. 
For these agents, small deviations of the center of mass may destabilize the body and interrupt locomotion. 
As a result, Walker and Hopper require longer training, typically around 3000 epochs, before reaching stable locomotion behaviors.
The importance of stable low-level control during cold-start exploration, together with representative collapse failures of unstable agents, is further discussed in Supplementary Sec.~B.


In our framework, the objective of low-level policy learning is therefore not to acquire full task semantics, which are learned by the meta-knowledge module, but to establish the embodiment-specific execution capability required for downstream deployment. 
This includes maintaining balance, generating stable locomotion patterns, and tracking the issued SMAI magnitudes. 
Once the low-level policy reaches a stable locomotion regime, it becomes a valid execution layer for the shared task inference module and high-level policy. 
The reusable meta-knowledge can then generate semantic-magnitude guidance, while the embodiment-specific low-level controller translates this guidance into executable motor actions. 
This separation between task reasoning and motor control provides a positive answer to Q2 and forms the basis for stable and efficient cross-embodiment deployment in ReMAP.

\subsection{Cross-Embodiment Meta Knowledge Reutilization}

\input{img_tex/single_task_inference_results}
\textbf{Case Study: Disentangling Task Semantics From Dynamics.}
Before evaluating cross-embodiment reutilization under the full non-parametric task setting, we first examine whether the learned meta-knowledge mainly encodes task-level semantics rather than embodiment-specific dynamics. 
This analysis is important for answering Q2, since reusable meta-knowledge should remain consistent across agents even when their physical structures and motion patterns differ. 
To avoid the additional ambiguity introduced by multi-task inference, we conduct a controlled single-task case study on the goal-forward task. 
The low-level policies are trained only for this task, while the task inference module and high-level policy learned from the dynamics-simplified agent with full non-parametric tasks (Fig.~\ref{fig:toy_multi_results}) are kept frozen and reused following the deployment pipeline in Fig.~\ref{fig:framework}c.

Fig.~\ref{fig:ant_single_task_inference_visualization}a shows the goal-tracking trajectories of four target agents, including Half-Cheetah, Ant, Walker, and Hopper. 
Although these agents have substantially different embodiments, ranging from quadrupedal and multi-legged structures to bipedal and single-legged locomotion, all agents can stably follow the goal-forward commands generated from the reused meta-knowledge module. 
This indicates that the frozen task-level module provides guidance that can be physically realized by different embodiment-specific low-level controllers.

To further inspect whether the inferred representation is affected by agent dynamics, we sample 50 trajectories for each agent and compare the inferred meta-knowledge encodings with the corresponding ground-truth target goals after the trajectories enter the stable execution stage, i.e., from the 50th time step onward. 
As shown in Fig.~\ref{fig:ant_single_task_inference_visualization}b, the inferred encodings exhibit a clear linear relationship with the target goals. 
Moreover, the regression trends across different agents are highly consistent, with no evident agent-specific deviation. 
This suggests that the inference module captures the parametric variation of the task itself, rather than the specific dynamics or action patterns used by each agent to execute the task.

Together with the non-parametric task clustering results in Fig.~\ref{fig:toy_multi_results}, this case study shows that the proposed meta-knowledge inference module can represent both categorical task semantics and within-task parametric variations. 
The embodiment-specific dynamics are instead absorbed by the downstream low-level policies. 
These results provide evidence for Q2: the proposed disentangled design separates task-level inference from low-level control, enabling the same meta-knowledge to be reused across heterogeneous agents.

\input{img_tex/comparison_4agents}
\textbf{Cross-Embodiment Meta-Knowledge Reutilization.}
We further evaluate whether the meta-knowledge acquired on the dynamics-simplified agent can be smoothly and stably reused across heterogeneous embodiments under the non-parametric task setting. 
The task space contains four task types: goal forward, goal backward, velocity forward, and velocity backward. 
During deployment, the task inference module, high-level policy, and target agent low-level policies are kept frozen. 
After online adaptation for the stride predictor, each target agent executes the generated semantic-magnitude subgoals through its own low-level controller. 
Fig.~\ref{fig:dpmm_multi_agent_inference_all} summarizes the deployment results on Half-Cheetah, Ant, Walker, and Hopper.

Figs.~\ref{fig:dpmm_multi_agent_inference_all}a--d show the representative inference trajectories for the four agents. 
For each agent, the upper plot reports goal-position tracking and the lower plot reports velocity tracking. 
Solid curves denote the executed trajectories, while dashed horizontal lines denote the corresponding task targets. 
Across all embodiments, the agents robustly follow the generated goal- or velocity-level commands through SMAI conditioning despite their substantially different dynamics and morphologies. 
This indicates that reused meta-knowledge provides executable task-level guidance, while embodiment-specific low-level policies translate it into feasible actions.

To analyze whether the transferred inference module preserves semantic task structure, we visualize the inferred latent representations after deployment using t-SNE. 
As shown in Figs.~\ref{fig:dpmm_multi_agent_inference_all}e--h, the latent embeddings form clearly separated clusters for the four task categories across all target agents. 
The cluster structure remains consistent even though the trajectories are collected from different embodiments. 
This suggests that the frozen inference module mainly captures task semantics, rather than agent-specific motion patterns, after being reused on complex agents.

We further quantify the semantic decodability of the inferred latent variables using decoder prediction confusion matrices. 
Figs.~\ref{fig:dpmm_multi_agent_inference_all}i--l show that the predictions concentrate strongly along the diagonal for all four agents, indicating that the inferred latent representations remain informative enough to recover the corresponding task identities after cross-agent deployment. 
Together with the t-SNE visualizations, these results show that the deployed latent space preserves both categorical task information and clear task separability across heterogeneous embodiments.

An additional observation from the trajectory plots is that some agents may exhibit temporary overshoot or transient tracking errors during the early stage of deployment, especially in the goal-tracking tasks. 
This is expected because the inference module relies on a finite context window and may require sufficient trajectory history before accurately identifying the current task mode and magnitude. 
As more interaction data becomes available, the inferred task representation becomes more reliable, allowing the high-level policy to generate corrected semantic-magnitude subgoals. 
The agents, therefore, recover from early deviations and continue tracking the desired task objectives instead of permanently diverging.

Overall, Fig.~\ref{fig:dpmm_multi_agent_inference_all} demonstrates that ReMAP can reuse the same frozen task-level meta-knowledge across agents with different embodiments. 
The inference module preserves task semantics, the decoder maintains high semantic prediction consistency, and the generated subgoals remain executable through embodiment-specific low-level controllers. 
These results provide direct evidence for cross-embodiment meta-knowledge reutilization and support the claim that task-level reasoning is effectively decoupled from low-level dynamics in the proposed framework.

\textbf{Baseline Comparison.}
We further compare ReMAP with four Meta-RL baselines, including \(\mathrm{RL}^{2}\), PEARL, CEMRL, and MELTS. 
Different from ReMAP, these methods jointly learn task inference and control under the target embodiment. 
Therefore, when the embodiment changes, both the task representation and the policy are coupled with the new dynamics and must be relearned or substantially adapted. 
This setting makes task inference less efficient, especially during early exploration, because the agent must simultaneously identify the task and discover feasible motor behaviors.
\input{img_tex/all_baselines_task_inferencen_results.tex}

Fig.~\ref{fig:baseline_multi_inference_all} shows the inference trajectories of the baselines on four non-parametric tasks for the Half-Cheetah embodiment. 
\(\mathrm{RL}^{2}\) and PEARL exhibit clear tracking limitations. 
\(\mathrm{RL}^{2}\) lacks an explicit task representation and mainly relies on recurrent hidden states to encode adaptation history, which leads to weak separation between different task modes. 
PEARL improves task inference by using probabilistic context variables, but its single-Gaussian latent prior tends to compress different task modes into a smooth representation. 
As a result, PEARL often produces averaged behaviors and shows limited adaptation to task parameters that are away from the dominant training distribution. 
This can be observed from the goal-tracking trajectories, where the predicted behaviors either drift away from the desired targets or converge to intermediate responses rather than accurately matching each target.

CEMRL and MELTS achieve stronger tracking performance than \(\mathrm{RL}^{2}\) and PEARL, especially for velocity-tracking tasks. 
CEMRL benefits from its GMM-based task representation and can distinguish multiple task modes when the predefined number of mixture components matches the underlying task structure. 
However, this assumption is restrictive in task-unknown settings: an improper component number can easily lead to incorrect task partitioning and unstable inference, as further examined in the ablation study. 
MELTS introduces a Bayesian non-parametric prior and can obtain more flexible task discovery than fixed-component mixture models. 
Nevertheless, because it is still trained end-to-end on the complex embodiment, its early-stage inference can be inaccurate when the agent has not yet collected informative task-semantic trajectories. 
Only after sufficient rollout context is accumulated can the inferred task representation become reliable enough to guide accurate tracking.

In contrast, as shown in Fig.~\ref{fig:dpmm_multi_agent_inference_all}, ReMAP decouples task-level meta-knowledge acquisition from embodiment-specific motor learning. 
The task inference module and high-level policy are learned on the dynamics-simplified agent and then reused as a frozen task-level module, while the target agent only needs to execute the generated semantic-magnitude subgoals through its low-level controller. 
This design enables more efficient task exploration from the beginning of deployment and avoids entangling task semantics with the complex dynamics of the target embodiment.

\input{img_tex/tracking_mse_comparison.tex}
To quantitatively evaluate tracking accuracy, Fig.~\ref{fig:tracking_mse_comparison} reports the per-step MSE between the task targets and the executed trajectories. 
For each method, each task is evaluated using 50 sampled trajectories with its best-achieved model, and the MSE is computed at each time step. 
For goal tracking, ReMAP achieves an average MSE of \(2.16\) over the full rollout, which reduces the error by \(95.9\%\), \(94.8\%\), \(89.0\%\), and \(89.3\%\) compared with PEARL (\(52.21\)), RL\(^{2}\) (\(41.35\)), CEMRL (\(19.68\)), and MELTS (\(20.24\)), respectively. 
At the final time step, ReMAP further reduces the goal-tracking MSE to \(0.005\), compared with \(74.91\), \(40.28\), \(8.15\), and \(2.79\) for PEARL, \(\mathrm{RL}^{2}\), CEMRL, and MELTS, respectively. 
This corresponds to absolute MSE reductions of \(74.90\), \(40.27\), \(8.14\), and \(2.79\), or relative reductions of \(99.99\%\), \(99.99\%\), \(99.93\%\), and \(99.81\%\). 
These results indicate that ReMAP not only converges to more accurate goal tracking, but also maintains stable long-horizon execution after the task has been inferred.

For velocity tracking, ReMAP obtains an average MSE of \(0.34\) over the full rollout, which is comparable to CEMRL (\(0.32\)) and MELTS (\(0.32\)), and substantially lower than PEARL (\(4.55\)) and RL\(^{2}\) (\(4.26\)). 
In the early stage of inference, ReMAP also shows faster error reduction: within the first 10 time steps, its velocity-tracking MSE is \(1.58\), compared with \(1.81\) for CEMRL, \(1.89\) for MELTS, \(4.07\) for PEARL, and \(4.46\) for RL\(^{2}\). 
At the final time step, ReMAP achieves a velocity-tracking MSE of \(0.157\), compared with \(3.73\), \(3.59\), \(0.190\), and \(0.301\) for PEARL, RL\(^{2}\), CEMRL, and MELTS, respectively. 
This yields absolute MSE reductions of \(3.57\), \(3.43\), \(0.033\), and \(0.144\), corresponding to relative reductions of \(95.78\%\), \(95.62\%\), \(17.42\%\), and \(47.80\%\). 
These results show that ReMAP achieves final tracking accuracy comparable to or better than the strongest baselines while reducing inference delay during deployment.

Overall, ReMAP reduces the final-step tracking MSE by \(94.75\%\)--\(99.79\%\) over the baselines under joint goal and velocity tracking. These results show that separating task-level meta-knowledge from embodiment-specific control enables more accurate goal tracking, comparable velocity tracking, and faster early-stage deployment than end-to-end Meta-RL baselines.


\input{img_tex/cost_comparison}
\textbf{Learning Cost for Cross-Embodiment.}
Finally, we evaluate the training cost required to achieve cross-embodiment deployment. 
Although the previous tracking results show that ReMAP achieves comparable or better execution accuracy than the strongest baselines, its main advantage lies in the amount of interaction data required to reach this performance. 
As shown in Fig.~\ref{fig:cost_comparison_transfer}, ReMAP requires only approx. \(38\)M environment interactions across the four embodiments, which is about \(23.8\%\) of the \(160\)M interactions required by MELTS. 
This indicates that ReMAP can achieve state-of-the-art deployment performance while incurring substantially lower training costs.

The reduction in training cost comes from two sources. 
First, task-level meta-knowledge is acquired on the dynamics-simplified agent, where task semantics can be learned more efficiently because the model does not need to solve high-dimensional embodiment-specific control at the same time. 
Once trained, the task inference module and the high-level policy are frozen and reused across target agents, so this cost is amortized over multiple embodiments. 
In contrast, end-to-end Meta-RL baselines need to relearn or substantially adapt both task inference and control whenever the embodiment changes.

Second, the objective of low-level learning in ReMAP is fundamentally different from that of end-to-end Meta-RL. 
In previous methods, the policy must simultaneously interpret the inferred task context and discover complex dynamics-dependent action patterns, which makes exploration and training highly sample-inefficient. 
In ReMAP, the low-level controller is trained independently through SMAI-conditioned curriculum learning. 
It only needs to track semantic-magnitude commands that gradually increase in difficulty, while acquiring embodiment-specific locomotion and exploration capabilities. 
Because task semantics are handled by the reusable meta-knowledge module, the low-level policy can focus on executable motor behavior rather than jointly learning task reasoning and control. 
This decoupled design substantially reduces the required interaction data while preserving strong downstream execution performance.
Overall, these results provide a positive answer to Q3: the learned task-level meta-knowledge can be effectively reused across heterogeneous embodiments.

Additional ablation studies on alternative latent priors are provided in Supplementary Sec.~C.

\section{CONCLUSION}
This paper presented ReMAP, a meta-knowledge reutilization framework for cross-embodiment Meta-RL. ReMAP learns DPMM-regularized task-level meta-knowledge on a dynamics-simplified agent and transfers it to heterogeneous embodiments through SMAI-guided low-level controllers. By decoupling task inference from embodiment-specific execution, the frozen task inference module and high-level policy can be reused without retraining for each target agent. Experiments on multiple locomotion embodiments show that ReMAP learns semantically separable task representations, achieves accurate goal and velocity tracking, and substantially reduces transfer cost compared with end-to-end Meta-RL baselines. These results demonstrate the effectiveness of reusable task-level meta-knowledge for efficient cross-embodiment policy transfer. Future work will extend the semantic interface and investigate sim-to-real deployment.



\vspace{-2cm}
\begin{IEEEbiography}[{\includegraphics[width=1in, height=1.25in, clip, keepaspectratio]{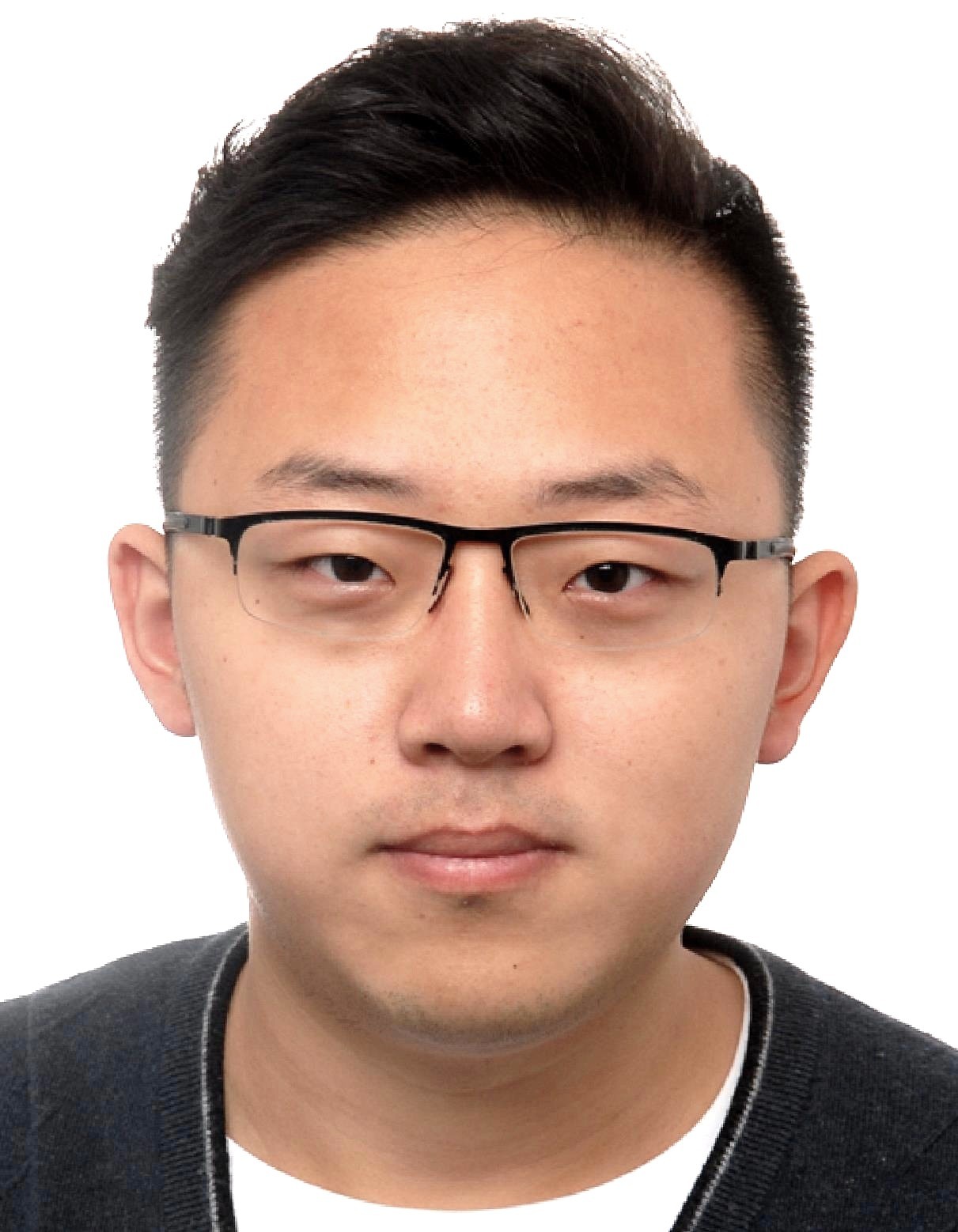}}]{Yuan Meng}
(Student Member, IEEE) received the B.Sc. degree in mechanical engineering in 2020 from RWTH Aachen University, Aachen, Germany, and the M.Sc. degree in mechatronics and robotics in 2023 from the Technical University of Munich, Munich, Germany, where he is currently working toward the Ph.D. degree in computer science with the Informatics 6.
His current research interests include robotic lifelong reinforcement learning and Vision-Language-Action model systems.
\end{IEEEbiography}
\vspace{-1.5cm}

\begin{IEEEbiography}[{\includegraphics[width=1in, height=1.25in, clip, keepaspectratio]{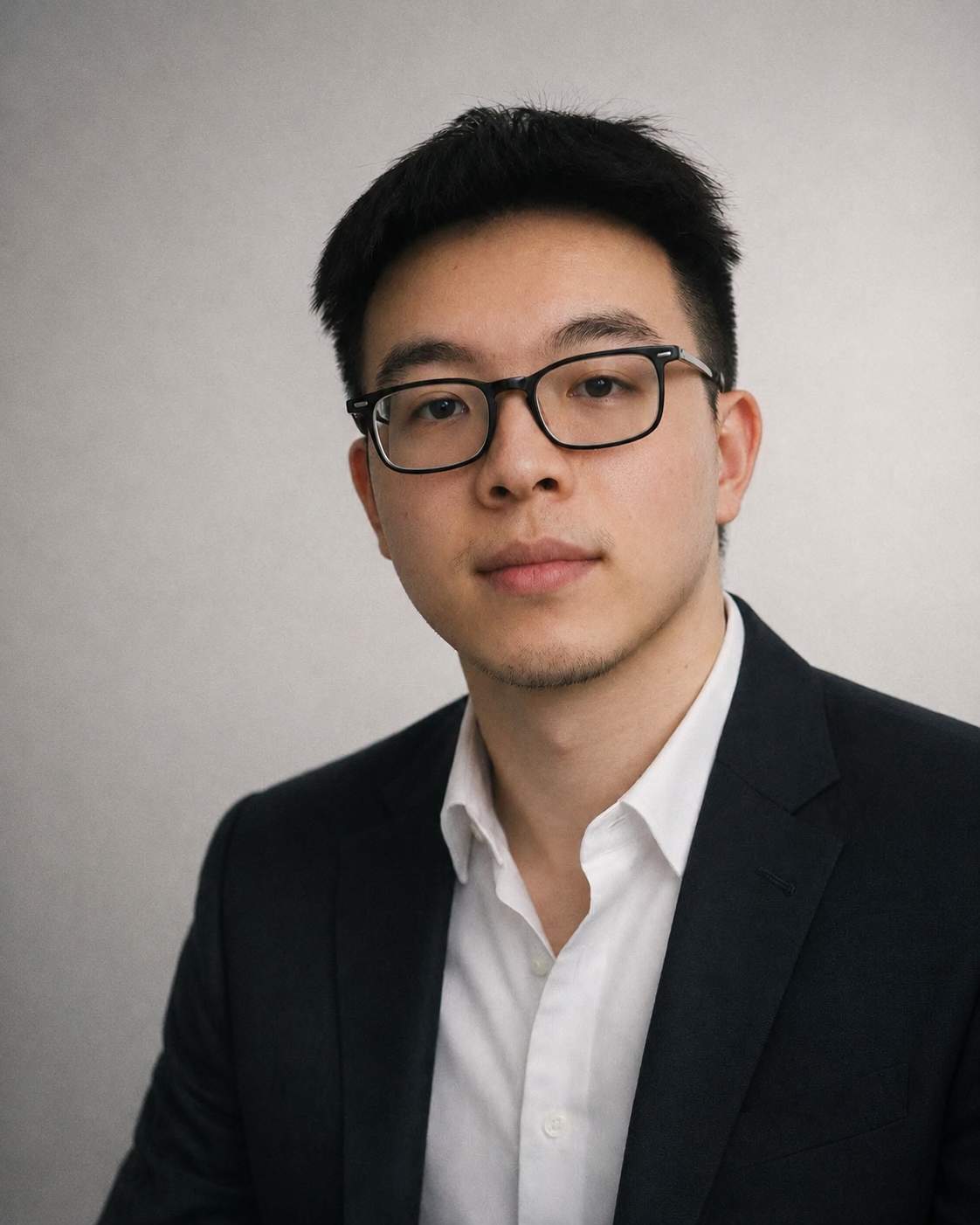}}]{Bo Wang} 
received the B.Sc. degree in computer science from Ostfalia University of Applied Sciences, Wolfenbuettel, Germany, in 2024. He is currently working toward the M.Sc. degree in Robotics, Cognition, and Intelligence at the Technical University of Munich, Munich, Germany. His current research interests include robotic meta-reinforcement learning, humanoid manipulation, and Vision-Language-Action model systems. 
\end{IEEEbiography}
\vspace{-1.5cm}

\begin{IEEEbiography}[{\includegraphics[width=1in, height=1.25in, clip, keepaspectratio]{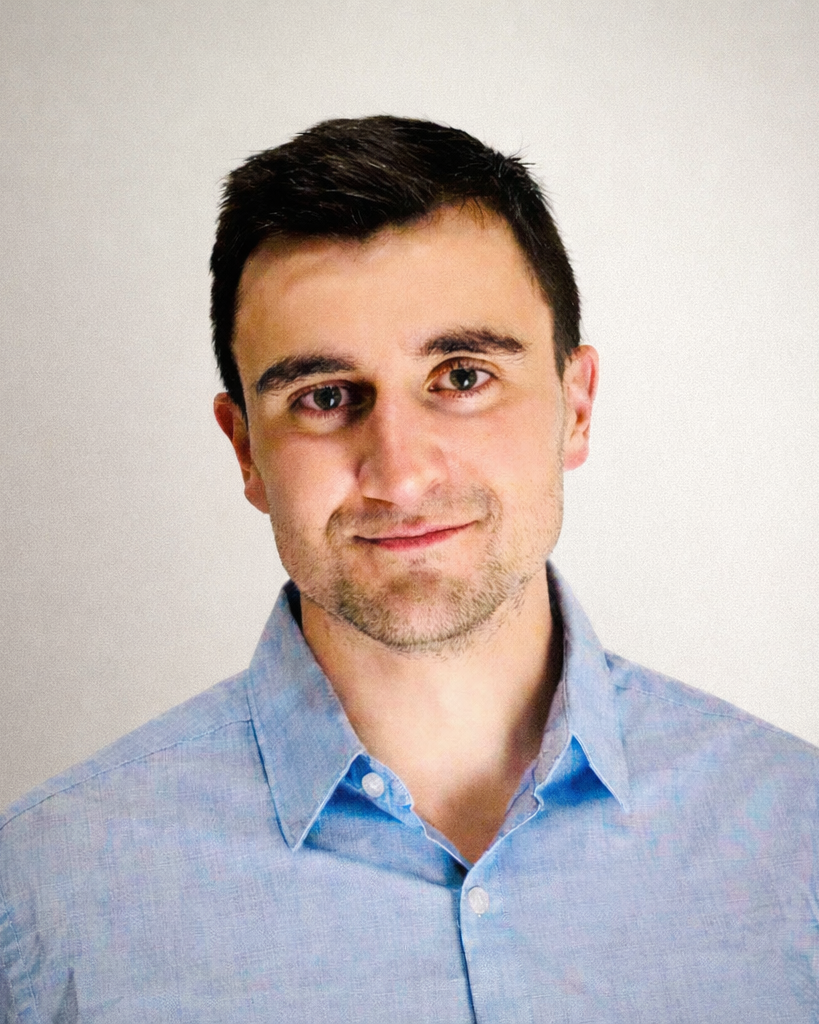}}]{Juan de los Rios}
received the B.Sc. degree in Engineering Science and the M.Sc. degree in Robotics, Cognition and Intelligence from the Technical University of Munich. His master’s thesis focused on meta reinforcement learning. His research interests include autonomous robotic systems, multi-agent reinforcement learning, and the coordination and orchestration of robotic agents.
\end{IEEEbiography}
\vspace{-1.5cm}

\begin{IEEEbiography}[{\includegraphics[width=1in, height=1.25in, clip, keepaspectratio]{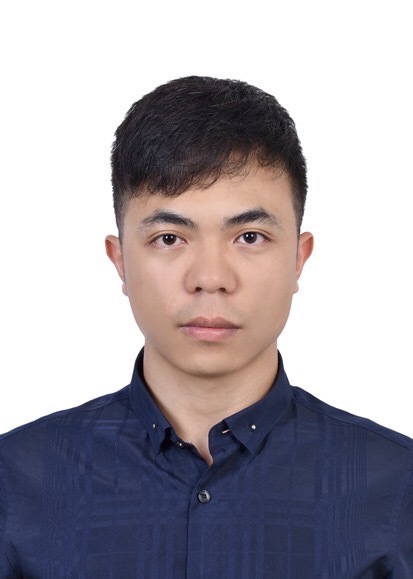}}]{Xiangtong Yao}
(Student Member, IEEE) received the B.Sc. and M.Sc. degrees in computer science from Sun Yat-sen University, Guangzhou, China, in 2017 and 2020, respectively. He is currently working toward the Ph.D. degree in computer science with Informatics 6, Technical University of Munich, Munich, Germany.
His current research interests include language control robotics and sensor-based control for manipulation.
\end{IEEEbiography}
\vspace{-1.75cm}

\begin{IEEEbiography}[{\includegraphics[width=1in, height=1.25in, clip, keepaspectratio]{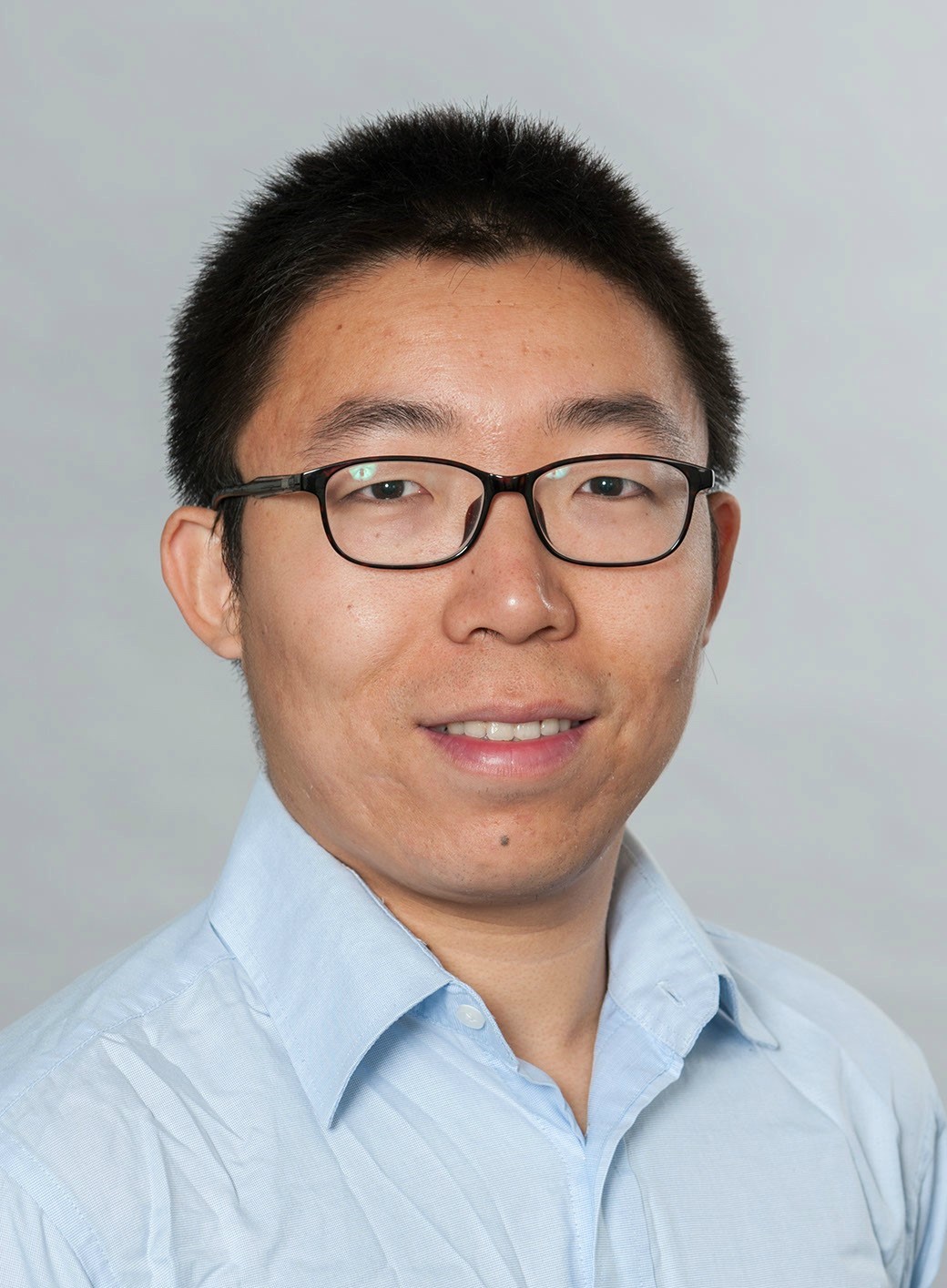}}]{Zhenshan Bing}
(Member, IEEE) received the B.S. degree in mechanical design manufacturing and automation and the M.Eng degree in mechanical engineering from the Harbin Institute of Technology, Harbin, China, in 2013 and 2015, respectively, and the doctorate degree in computer science from the Technical University of Munich, Munich, Germany, in 2019.
He worked as a Postdoctoral Researcher with Informatics 6, Technical University of Munich, from 2019 to 2025. 
He is currently an Associate Professor with the State Key Laboratory for Novel Software Technology, Nanjing University (Suzhou Campus), Suzhou, China
\end{IEEEbiography}
\vspace{-1.5cm}

\begin{IEEEbiography}[{\includegraphics[width=1in, height=1.25in, clip, keepaspectratio]{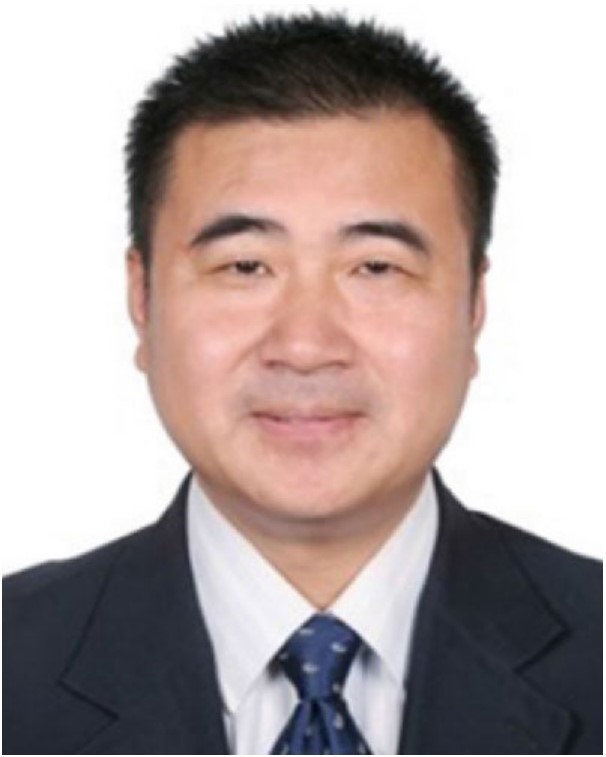}}]{Fuchun Sun}
(Fellow, IEEE) received the Ph.D. degree in computer science from Tsinghua University, Beijing, China, in 1997.
He is currently a Full Professor with the Department of Computer Science and Technology,
Tsinghua University. His current research interests include robotic perception and cognition.
Dr. Sun was the recipient of the National Science Fund for Distinguished Young Scholars.
\end{IEEEbiography}
\vspace{-1.5cm}

\begin{IEEEbiography}[{\includegraphics[width=1in, height=1.25in, clip, keepaspectratio]{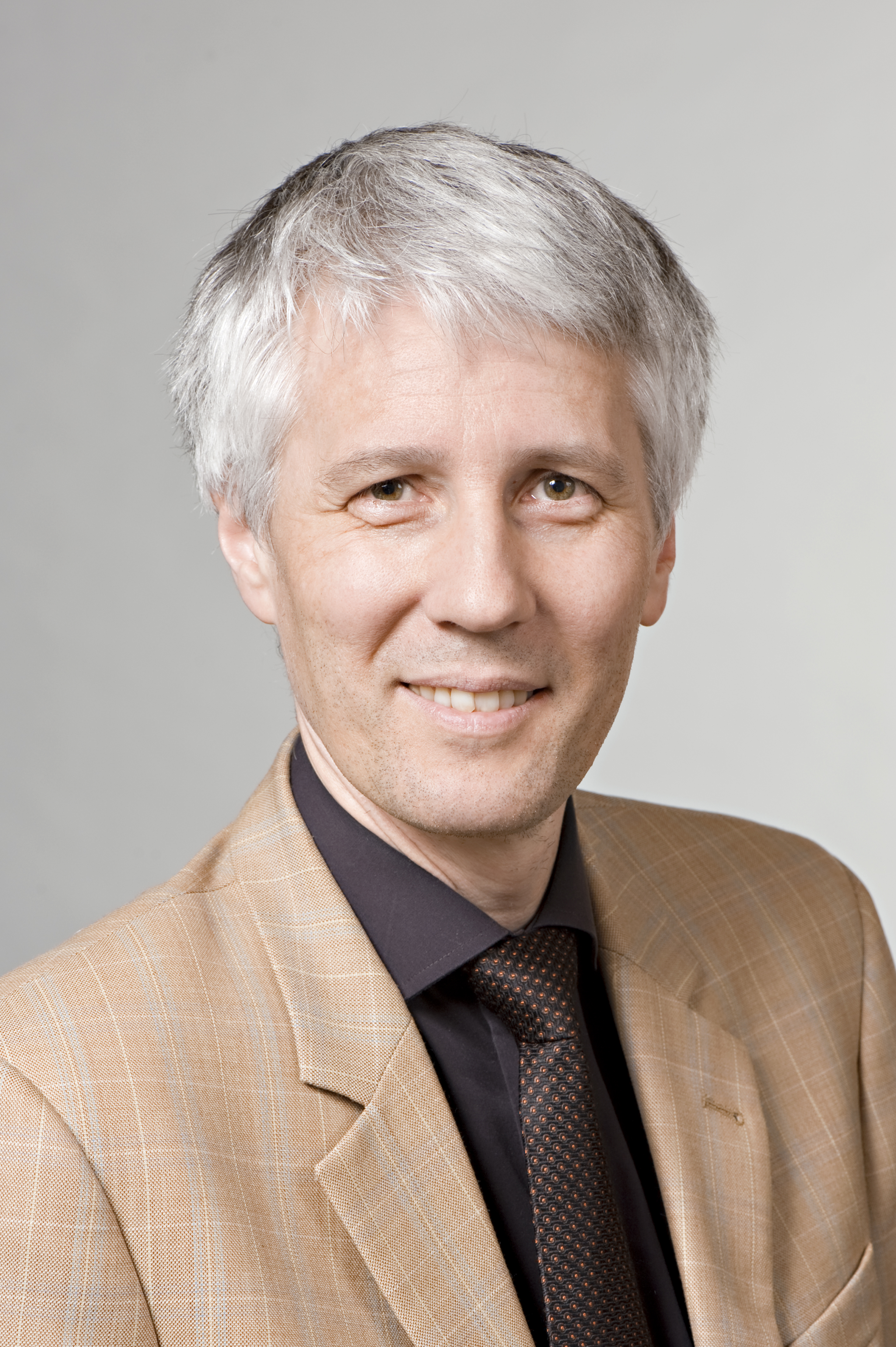}}]{Alois Knoll}
(Fellow, IEEE) received the diploma (M.Sc.) degree in electrical/communications
engineering from the University of Stuttgart, Stuttgart, Germany, in 1985, and the Ph.D. (summa cum laude) degree in computer science from the Technical University of Berlin (TU Berlin), Berlin, Germany, in 1988.
He served on the Faculty of the Computer Science Department, TU Berlin, until 1993. 
He joined the University of Bielefeld, Germany, as a Full Professor and served as the Director of the Technical Informatics research group until 2001. 
Since 2001, he has been a Professor with the Department of Informatics, Technical University of Munich, Munich, Germany
\end{IEEEbiography}


\end{document}

%% file: img_tex/introduction_overview.tex
\begin{figure}[t!]
    \centering
    \resizebox{\linewidth}{!}{
    \begin{tikzpicture}[font=\small\sffamily]
    \coordinate (zero) at (0,0);
        \node[shading=axis, right color=fforange_pv!20, left color=ffgreen_pv!20, rounded corners=2pt, minimum width=12cm, minimum height=6cm](bg)at(0, 0){};
        \node[inner sep=0pt, fill=white, fill opacity=.5, text opacity=1, right](dpmm)at([xshift=1.75cm, yshift=1.25cm]bg.west){\includegraphics[width=2cm]{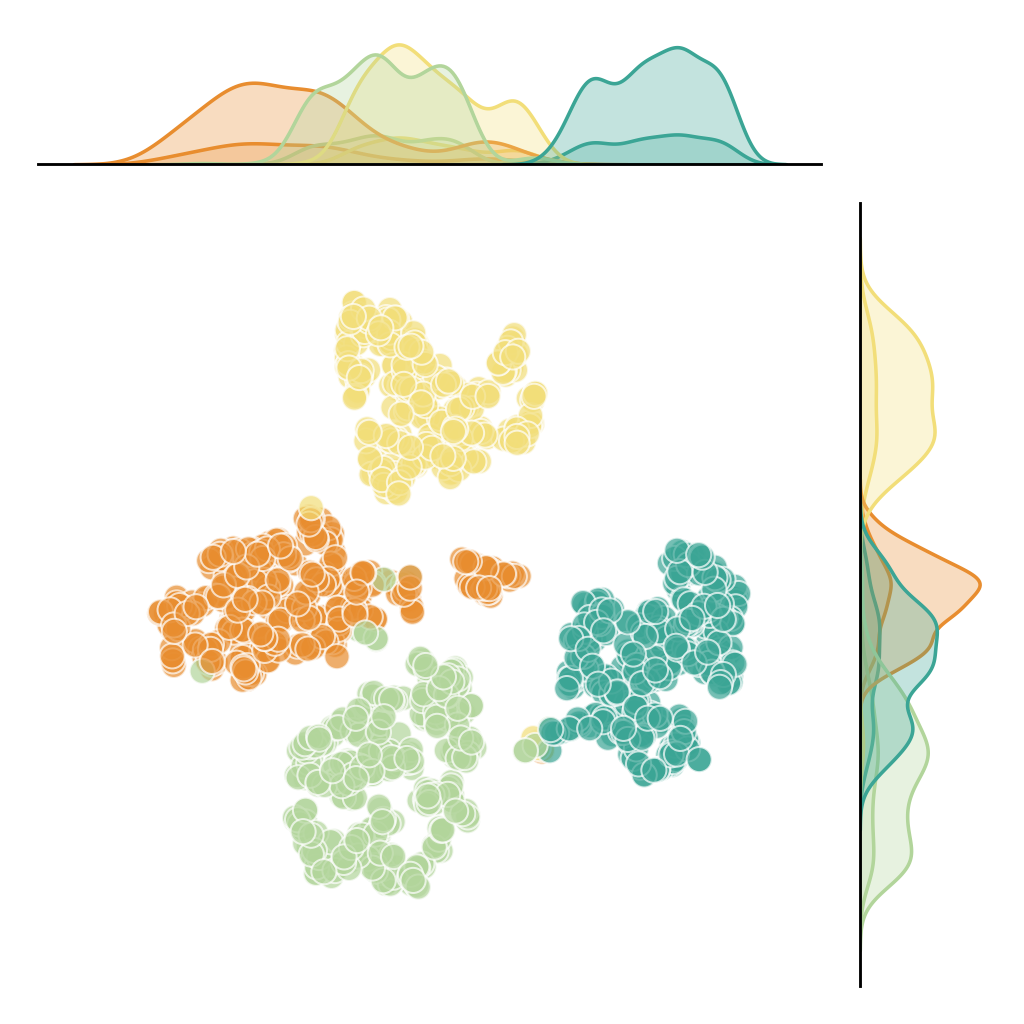}};
        \node[trapezium, fill=ffgreen_pv!50, draw=ffgreen_pv, line width=1pt, rounded corners=2pt, minimum width=1.5cm, minimum height=1cm, rotate=-90, below](encoder)at([xshift=-.2cm]dpmm.west){};
        \node[trapezium, fill=ffgreen_pv!50, draw=ffgreen_pv, line width=1pt, rounded corners=2pt, minimum width=1.5cm, minimum height=1cm, rotate=90, below](decoder)at([xshift=.2cm]dpmm.east){};
        \node[rectangle, fill=ffgreen_pv!50, draw=ffgreen_pv, line width=1pt, rounded corners=2pt, minimum width=2cm, minimum height=.25cm, below](hlpolicy)at([yshift=-.2cm]dpmm.south){};
        \node[inner sep=0pt, below](toy)at([yshift=-.2cm]hlpolicy.south){\includegraphics[width=2cm]{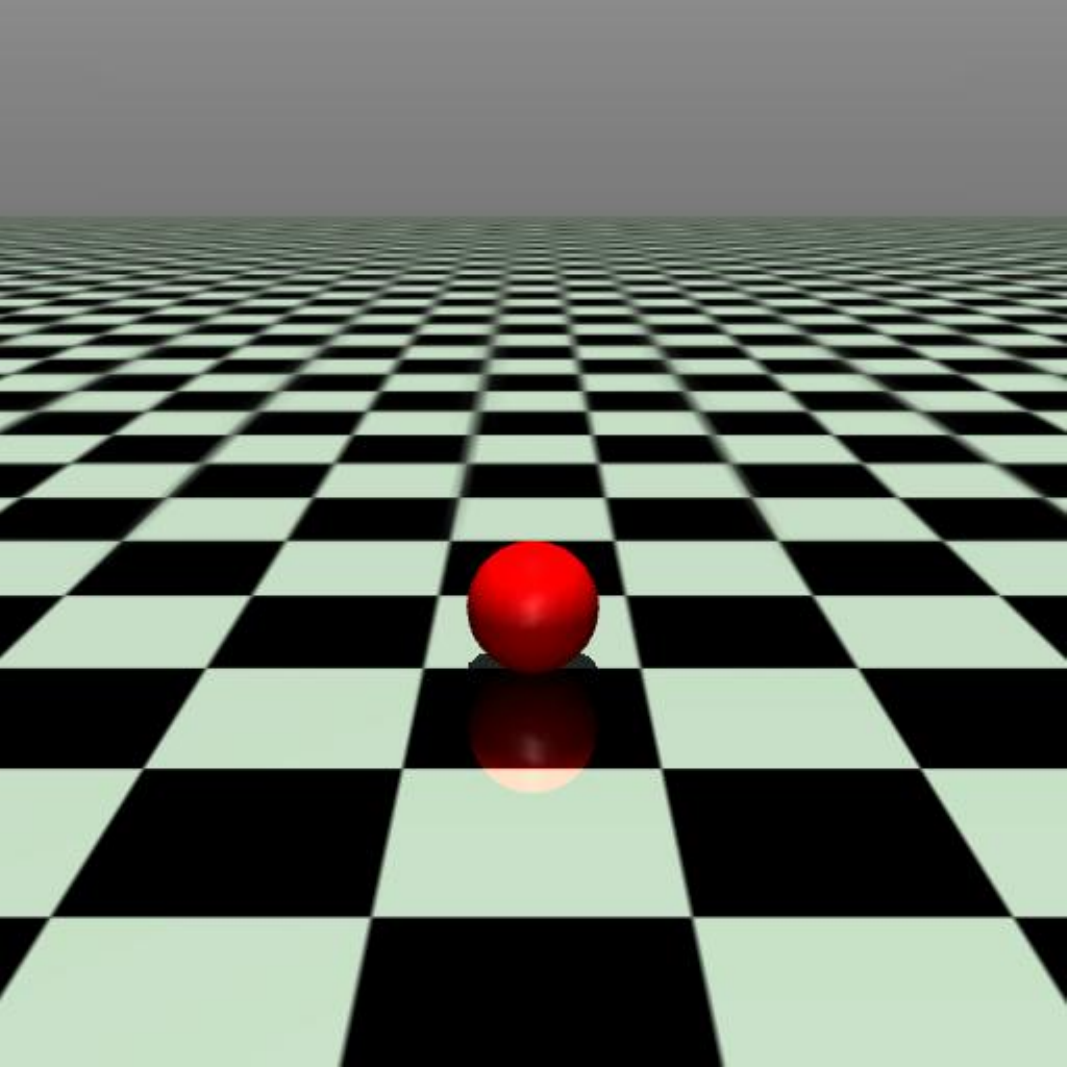}};

        \node[inner sep=0pt, right](smai)at(zero){\includegraphics[width=1.5cm]{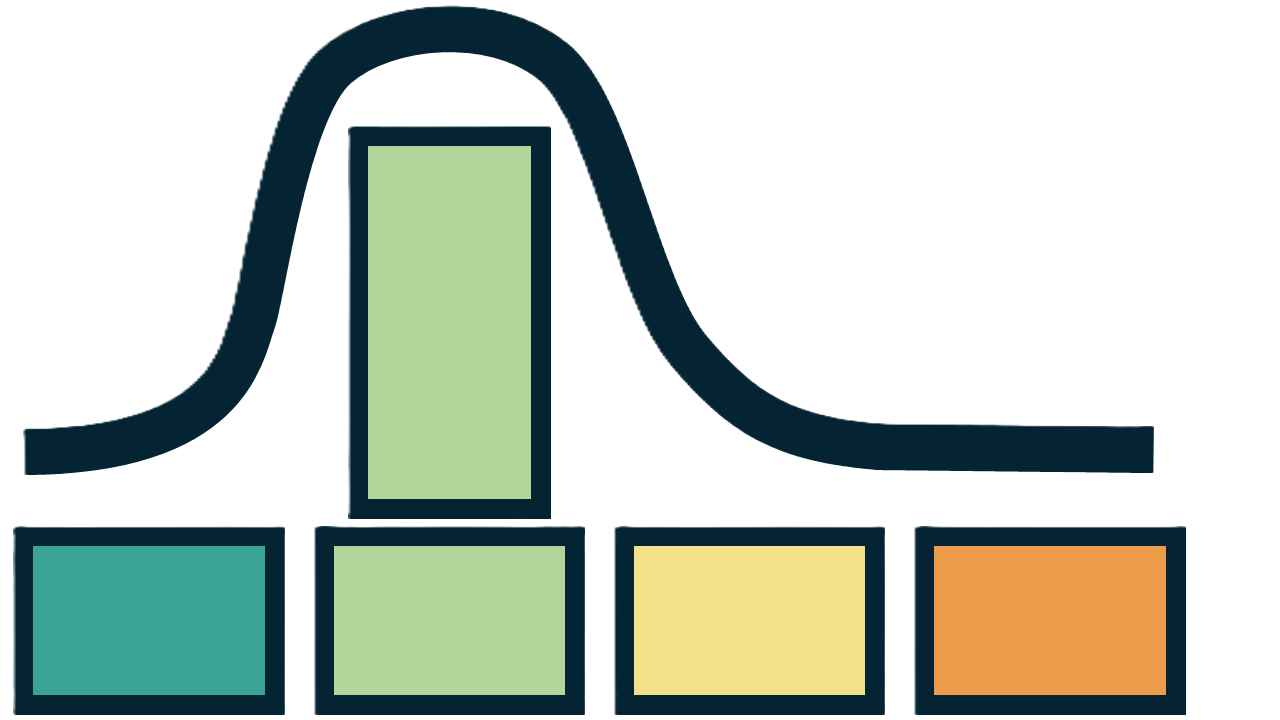}};

        \node[inner sep=0pt, draw=fforange_pv, fill=white, opacity=.5, text opacity=1, line width=1pt, rounded corners=2pt, above left](actor)at([xshift=-.75cm, yshift=0cm]bg.east){\includegraphics[width=2.5cm]{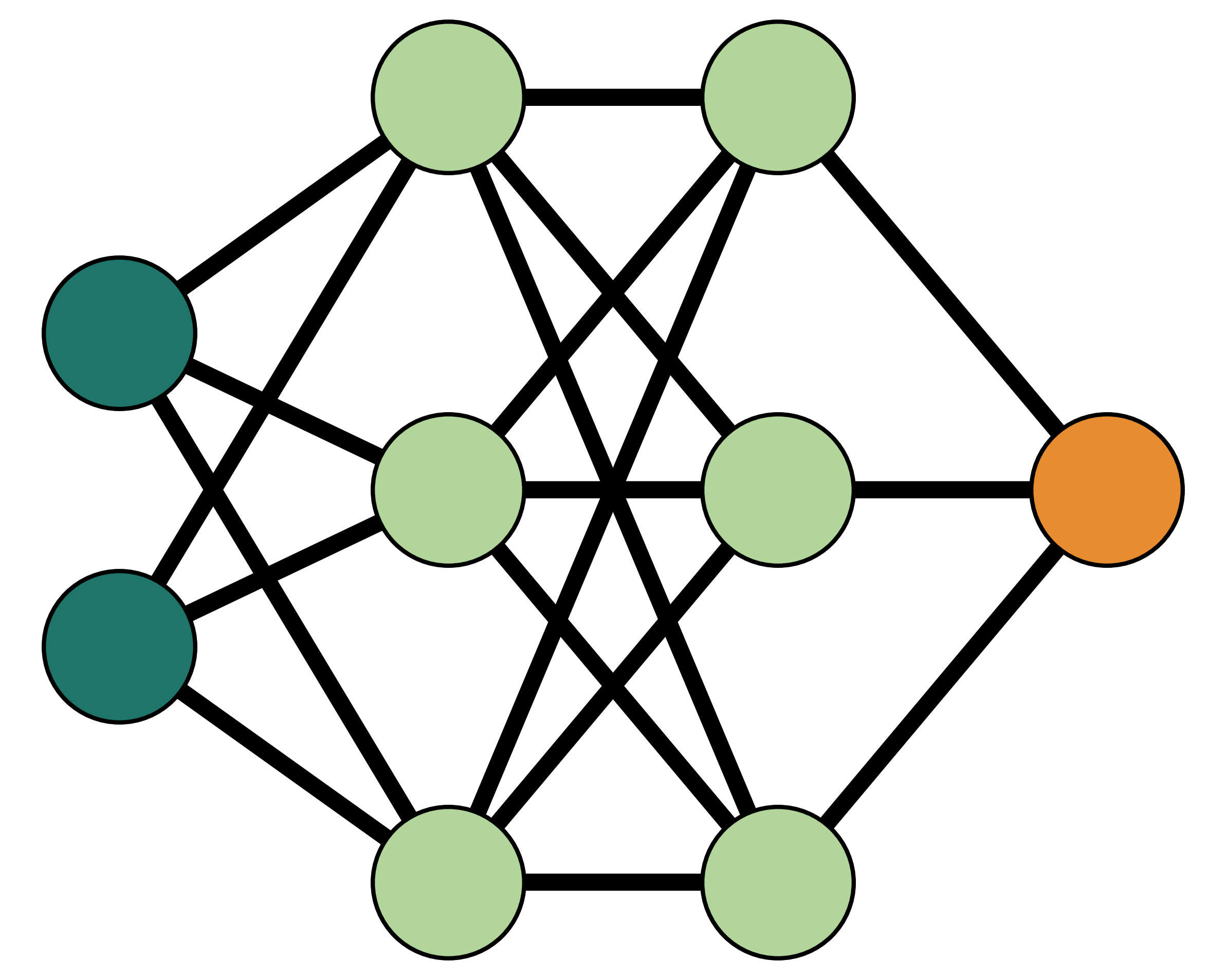}};
        \node[inner sep=0pt, below](ant)at([yshift=-.4cm]actor.south){\includegraphics[width=2cm]{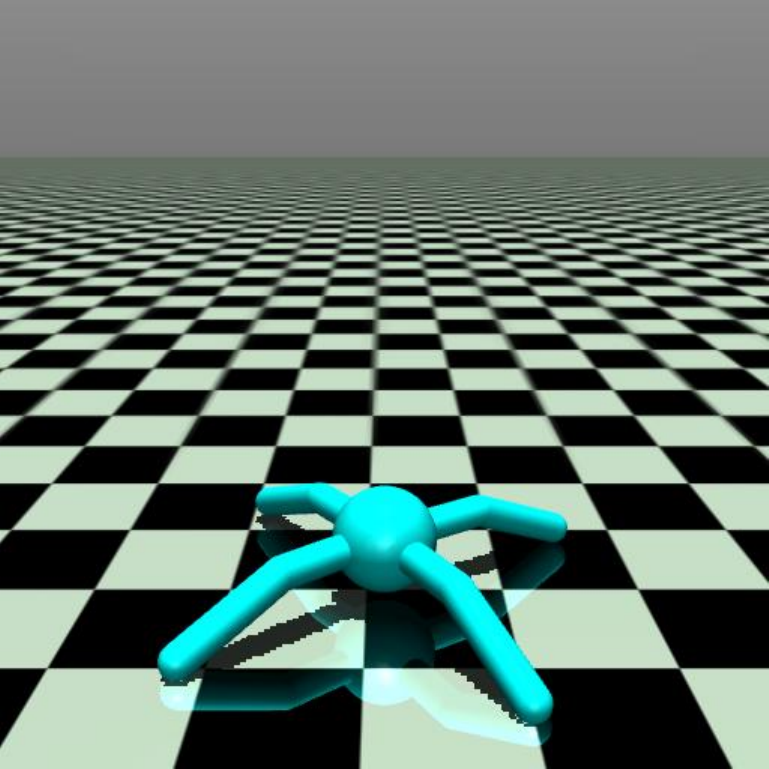}};

        \draw[->, color=ffgreen_pv, line width=2pt, rounded corners=2pt](decoder.south) -- ++ (.75cm, 0) -- ([xshift=-.35cm,yshift=.2cm]smai.north);
        \draw[->, color=ffgreen_pv, line width=2pt, rounded corners=2pt](toy.east) -- ++ (1.5cm, 0) -- ([xshift=-.35cm,yshift=-.2cm]smai.south);
        \draw[->, color=ffgreen_pv, line width=2pt, rounded corners=2pt](smai.east) -- ++ (1cm, 0);

        \node[above,]at(dpmm.north){\scriptsize Meta-Knowledge Acquisition};
        \node[above, text width=4cm, align=center]at(actor.north){\scriptsize Magnitude-triggered Disentangled Learning};
        \node[above]at(bg.south){\textbf{Meta-Knowledge Reuitilization}};
        \node[]at(encoder.center){\scriptsize Encoder};
        \node[]at(decoder.center){\scriptsize Decoder};
        \node[above]at(dpmm.south){\scriptsize DPMM Prior};
        \node[below right]at(toy.east){\scriptsize Magnitude $m_t$};
        \node[above right]at(decoder.south){\scriptsize Task Prediction $y_t$};
        \node[below, text width=3.3cm, align=center]at([xshift=1cm, yshift=0cm]smai.south){\scriptsize Semantic-Magnitude \\ Interface};
        \node[below]at(ant.north){\scriptsize\textcolor{white}{Complex Agent}};
        \node[below]at(toy.north){\scriptsize\textcolor{white}{Simplified Agent}};
    \end{tikzpicture}
    }
    \vskip -.05in
    \caption{
    Overview of our proposed framework. 
    The framework learns task-level meta-knowledge on a dynamics-simplified agent, trains target-agent low-level policies via magnitude-guided adaptation, and reuses the frozen meta-knowledge as semantic-magnitude subgoals for efficient deployment across heterogeneous agents.
    }
    \label{fig:overview}
\vskip -.15in
\end{figure}


%% file: img_tex/framework.tex
\begin{figure*}[t]
    \centering
    \resizebox{.98\textwidth}{!}{
    \begin{tikzpicture}[font=\sffamily\footnotesize]
        \node[shading=axis, top color=fforange_pv!20, bottom color=ffgreen_pv!20, rounded corners=2pt, minimum width=20cm, minimum height=16cm](bg0)at(0, 0){};
        \node[below right](label1)at(bg0.north west){\Large \textcolor{fforange_pv}{\textbf{I. Meta-knowledge Acquisition \& Disentangled Learning}}};
        \node[above right](label2)at(bg0.south west){\Large \textcolor{ffgreen_pv}{\textbf{II. Meta-Knowledge Reutilization}}};

        \node[draw=fforange_pv, fill=fforange_pv!30, rounded corners=2pt, minimum width=12cm, minimum height=7cm, below right](bg1)at([xshift=.5cm, yshift=-.75cm]bg0.north west){};
        \node[draw=fforange_pv, fill=fforange_pv!30, rounded corners=2pt, minimum width=6.5cm, minimum height=6.5cm, below left](bg2)at([xshift=-.5cm, yshift=-1.25cm]bg0.north east){};
        \node[draw=fforange_pv, fill=fforange_pv!30, rounded corners=2pt, minimum width=6.5cm, minimum height=6.5cm, above left](bg2-2)at([xshift=-.15cm, yshift=0.25cm]bg2.south east){};
        \node[draw=fforange_pv, fill=fforange_pv!30, rounded corners=2pt, minimum width=6.5cm, minimum height=6.5cm, above left](bg2-3)at([xshift=-.3cm, yshift=0.5cm]bg2.south east){};

        \node[inner sep=0pt, fill=white, fill opacity=.5, text opacity=1, below left](dpmm)at([xshift=-.5cm,yshift=-.25cm]bg1.north){\includegraphics[width=2.5cm]{imgs/toy-tsne.png}};
        \node[trapezium, fill=fforange_pv!50, draw=fforange_pv, line width=1pt, rounded corners=2pt, minimum width=1.5cm, minimum height=1cm, rotate=-90,](encoder)at([xshift=-1cm]dpmm.west){};
        \node[rectangle, fill=fforange_pv!50, draw=fforange_pv, line width=1pt, rounded corners=2pt, minimum width=3cm, minimum height=.5cm, right](deco-rew)at([xshift=.5cm]dpmm.east){Reward Generation};
        \node[rectangle, fill=fforange_pv!50, draw=fforange_pv, line width=1pt, rounded corners=2pt, minimum width=3cm, minimum height=.5cm, above](deco-state)at([yshift=.1cm]deco-rew.north){State Prediction};
        \node[rectangle, fill=fforange_pv!50, draw=fforange_pv, line width=1pt, rounded corners=2pt, minimum width=3cm, minimum height=.5cm, below](deco-task)at([yshift=-.1cm]deco-rew.south){Task Confidence};
        \node[inner sep=0pt, above left](toy)at([xshift=-.5cm, yshift=.25cm]bg1.south){\includegraphics[width=2.5cm]{imgs/toy.pdf}};
        \draw[line width=1pt, dashed, color=fforange_pv, rounded corners=2pt] ([xshift=-.1cm, yshift=.1cm]deco-state.north west) rectangle ([xshift=.1cm, yshift=-.1cm]deco-task.south east);
        \node[rectangle, fill=fforange_pv!50, draw=fforange_pv, line width=1pt, rounded corners=2pt, minimum width=1.5cm, minimum height=1cm, right](hlpolicy)at([xshift=1cm]toy.east){High-level Policy};
        \node[rectangle, fill=fforange_pv!10, draw=ffblue, line width=1pt, rounded corners=2pt, minimum width=1.25cm, minimum height=.5cm, right](L-task)at([xshift=1cm]deco-task.east){$\mathcal{L}_{\mathrm{task}}$};
        \node[rectangle, fill=fforange_pv!10, draw=ffblue, line width=1pt, rounded corners=2pt, minimum width=1.25cm, minimum height=.5cm, right](L-rew)at([xshift=1cm]deco-rew.east){$\mathcal{L}_{\mathrm{reward}}$};
        \node[rectangle, fill=fforange_pv!10, draw=ffblue, line width=1pt, rounded corners=2pt, minimum width=1.25cm, minimum height=.5cm, right](L-state)at([xshift=1cm]deco-state.east){$\mathcal{L}_{\mathrm{state}}$};
        \node[rectangle, fill=fforange_pv!10, draw=ffblue, line width=1pt, rounded corners=2pt, minimum width=1cm, minimum height=.5cm, below right](L-kld)at([yshift=-.5cm]dpmm.south west){$D_{\mathrm{KL}}$};
        \node[rectangle, fill=fforange_pv!10, draw=ffdarkgreen, line width=1pt, rounded corners=2pt, minimum width=1.25cm, minimum height=.5cm, right](L-hlpolicy)at([xshift=1cm]hlpolicy.east){$\mathcal{L}_{\pi}^{\mathrm{high}}$};

        \draw[->, color=fforange_pv, line width=1pt, rounded corners=2pt](toy.west) -| ([xshift=-1cm, yshift=.5cm]encoder.south) -- ([yshift=.5cm]encoder.south);
        \draw[->, color=fforange_pv, line width=1pt, rounded corners=2pt](toy.west) -| ([xshift=-1cm]encoder.south) -- (encoder.south);
        \draw[->, color=fforange_pv, line width=1pt, rounded corners=2pt](toy.west) -| ([xshift=-1cm, yshift=-.5cm]encoder.south) -- ([yshift=-.5cm]encoder.south);
        \draw[->, color=fforange_pv, line width=1pt, rounded corners=2pt]([yshift=.2cm]encoder.north) -- ([yshift=.2cm]dpmm.west);
        \draw[->, color=fforange_pv, line width=1pt, rounded corners=2pt]([yshift=.2cm]dpmm.east) -- ([xshift=-.1cm, yshift=.2cm]deco-rew.west);
        \draw[->, color=fforange_pv, line width=1pt, rounded corners=2pt](hlpolicy.west) -- (toy.east);
        \draw[->, color=fforange_pv, line width=1pt, rounded corners=2pt](dpmm.south) |- ([xshift=.5cm, yshift=1.5cm]hlpolicy.north) -- ([xshift=.5cm]hlpolicy.north);
        \draw[->, color=fforange_pv, line width=1pt, rounded corners=2pt]([xshift=.5cm]toy.north) |- ([xshift=-.5cm, yshift=1cm]hlpolicy.north) -- ([xshift=-.5cm]hlpolicy.north);
        
        \draw[->, color=ffblue, dashed, line width=1pt, rounded corners=2pt](L-task.west) -- ([xshift=.1cm]deco-task.east);
        \draw[->, color=ffblue, dashed, line width=1pt, rounded corners=2pt](L-rew.west) -- ([xshift=.1cm]deco-rew.east);
        \draw[->, color=ffblue, dashed, line width=1pt, rounded corners=2pt](L-state.west) -- ([xshift=.1cm]deco-state.east);
        \draw[->, color=ffblue, dashed, line width=1pt, rounded corners=2pt](L-kld.north) --++ (0, .5cm);
        \draw[->, color=ffblue, dashed, line width=1pt, rounded corners=2pt]([xshift=-.1cm,yshift=-.2cm]deco-rew.west) -- ([yshift=-.2cm]dpmm.east);
        \draw[->, color=ffblue, dashed, line width=1pt, rounded corners=2pt]([yshift=-.2cm]dpmm.west) -- ([yshift=-.2cm]encoder.north);
        \draw[->, color=ffdarkgreen, dashed, line width=1pt, rounded corners=2pt](L-hlpolicy.west) -- (hlpolicy.east);

        \node[below]at([yshift=-.2cm]encoder.east){Encoder};
        \node[below]at([yshift=-.1cm]deco-task.south){Decoder};
        \node[above]at([yshift=.1cm]dpmm.south){DPMM Prior};
        \node[above right]at([xshift=.5cm, yshift=.5cm]hlpolicy.north){$\bar{z}_t = \mathrm{sg}(z_t)$};
        \node[above right]at([xshift=-.5cm, yshift=.5cm]hlpolicy.north){$s_t$};
        \node[below left]at(hlpolicy.west){$a^{\mathrm{high}}_{t}$};
        \node[below left]at(toy.west){$\mathbf{c}_t=\{(s, s', r)\}^{H}_{t}$};
        \node[above left](s)at([xshift=-.5cm]encoder.south){$s$};
        \node[above left]at([xshift=-.4cm,yshift=.5cm]encoder.south){$s'$};
        \node[above left]at([xshift=-.5cm,yshift=-.5cm]encoder.south){$r$};
        \node[above left]at(bg1.south east){\textbf{Branch I: Meta-Knowledge Acquisition}};
        \node[below]at(toy.north){\textcolor{white}{Simplified Agent}};

        \node[inner sep=0pt, draw=fforange_pv, fill=white, opacity=.5, text opacity=1, line width=1pt, rounded corners=2pt, above left](actor)at([xshift=-1cm]bg2-3.center){\includegraphics[width=1.5cm]{imgs/neural_network.png}};
        \node[inner sep=0pt, draw=fforange_pv, fill=white, opacity=.5, text opacity=1, line width=1pt, rounded corners=2pt, above right](critic)at([xshift=1cm]bg2-3.center){\includegraphics[width=1.5cm]{imgs/neural_network.png}};
        \node[inner sep=0pt, below right](ant)at([yshift=-.5cm]actor.south west){\includegraphics[width=2cm]{imgs/ant.pdf}};
        \node[rectangle, fill=fforange_pv!10, draw=ffdarkgreen, line width=1pt, rounded corners=2pt, minimum width=1.25cm, minimum height=.5cm,](L-llpolicy)at([yshift=.1cm]bg2-3.center){$\mathcal{L}_{\pi}^{\mathrm{low}}$};
        \node[inner sep=0pt, below right, label={[label distance=0pt]-90:\scriptsize\textsf{Magnitude}$\uparrow$}](mag)at([xshift=.4cm, yshift=-.2cm]ant.north east){\includegraphics[width=1cm]{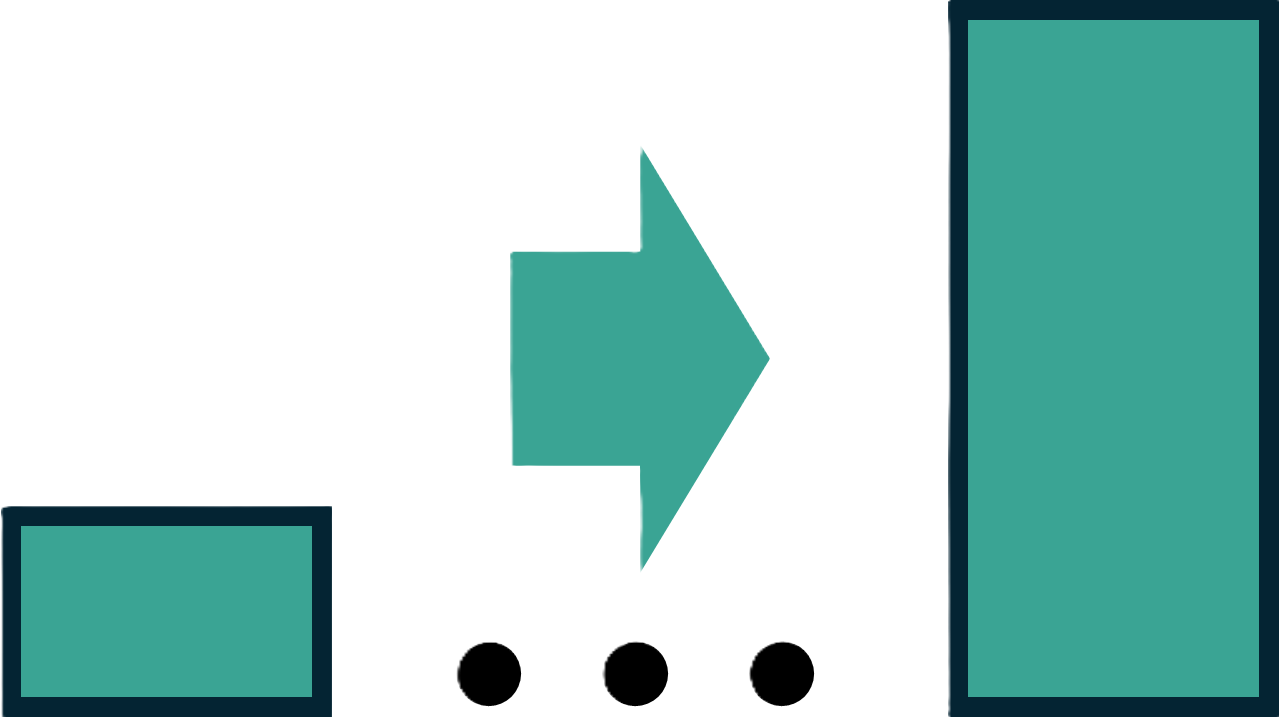}};
        \node[inner sep=0pt, below, label={[label distance=0pt]-90:\scriptsize\textsf{Switch}$\uparrow$}](temp)at([yshift=-.4cm]mag.south){\includegraphics[width=1cm]{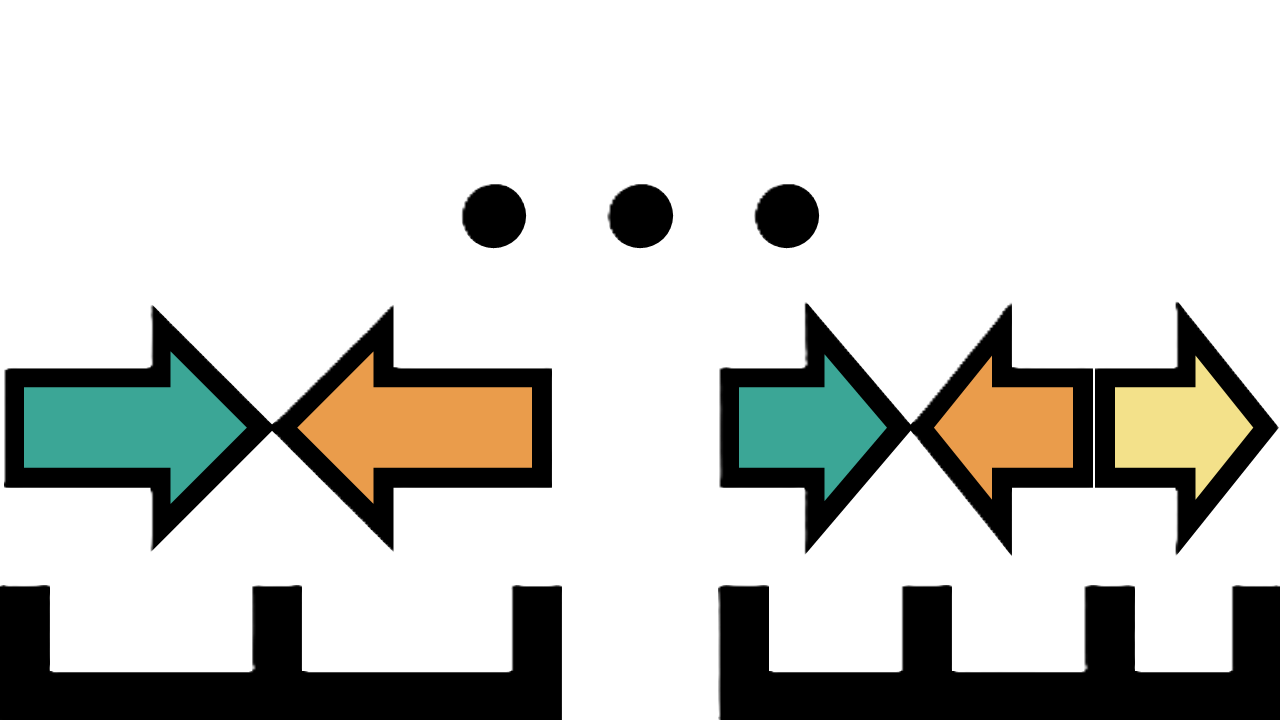}};
        \node[inner sep=0pt, right, label={[label distance=0pt]-90:\scriptsize\textsf{Horizon}$\uparrow$}](traj)at([xshift=.75cm]mag.east){\includegraphics[width=1cm]{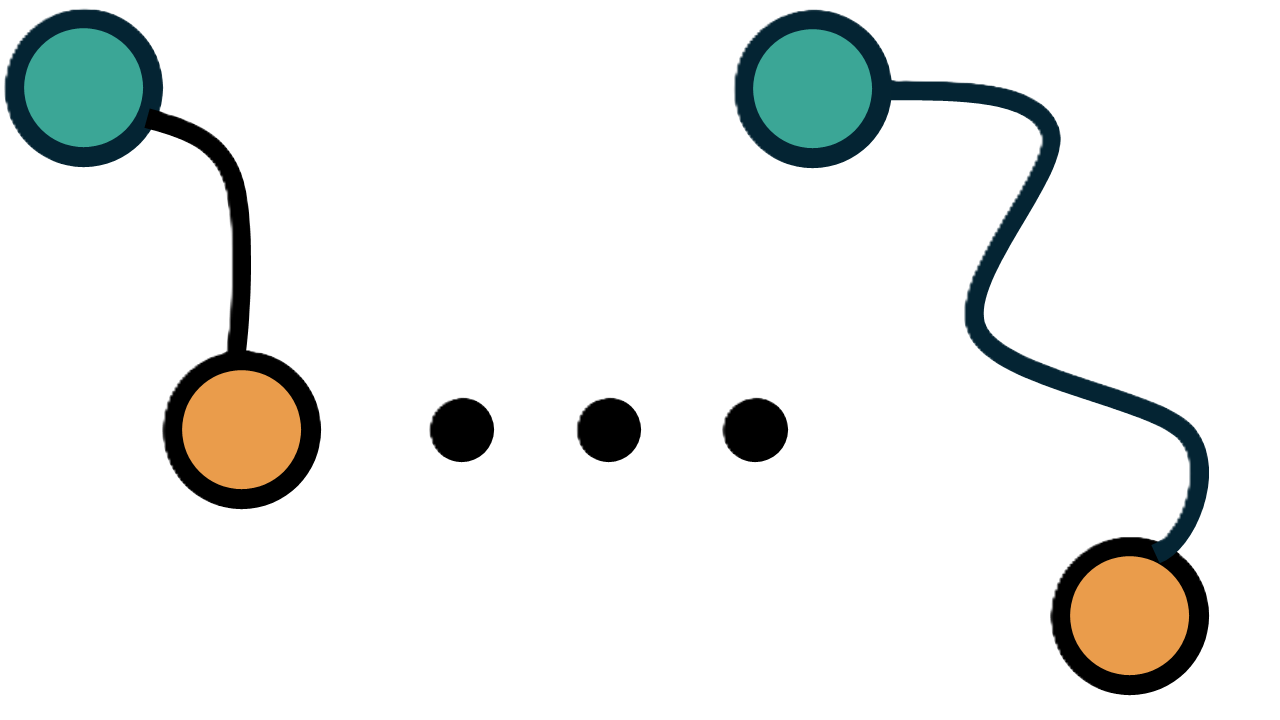}};
        \node[inner sep=0pt, right, label={[label distance=0pt]-90:\scriptsize\textsf{Distribution}$\uparrow$}](dist)at([xshift=.75cm]temp.east){\includegraphics[width=1cm]{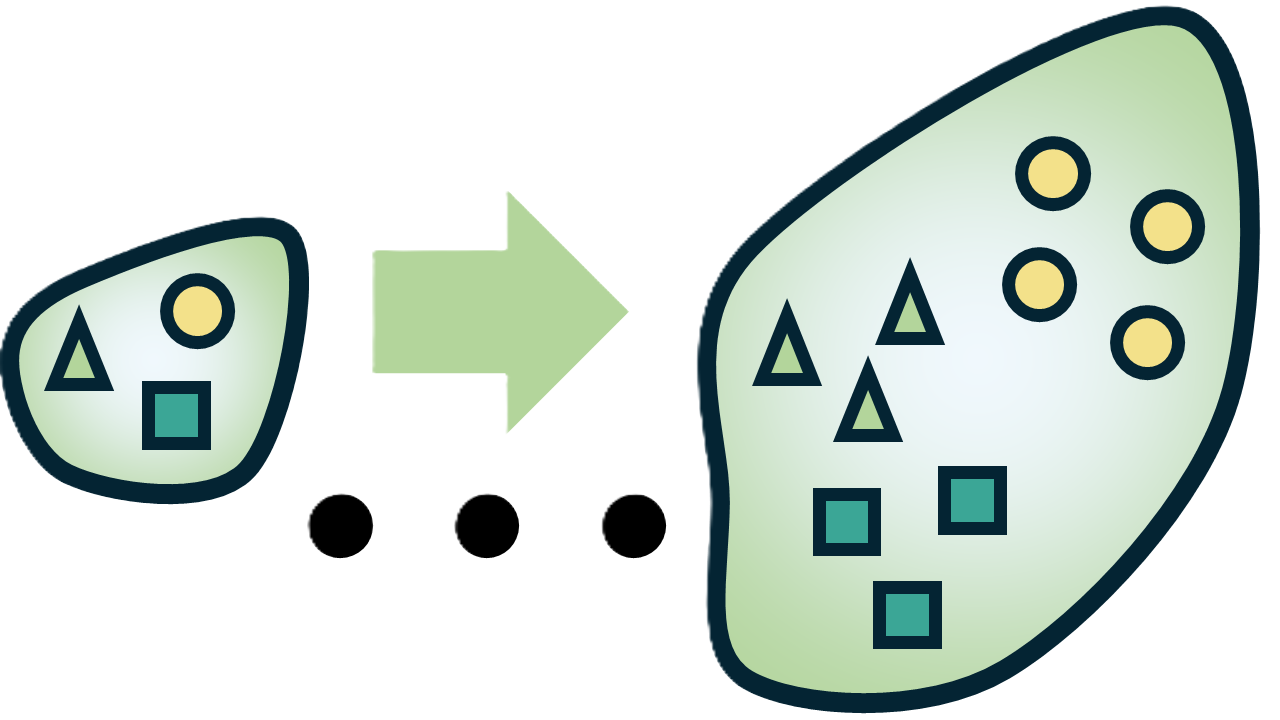}};

        \node[inner sep=0pt, above](smai)at([xshift=.5cm, yshift=1cm]actor.north east){\includegraphics[width=1.5cm]{imgs/smai.png}};

        \node[above left]at(bg2-3.south east){\textbf{Branch II: Disentangled Adaptive Learning}};
        \node[below right]at(bg2.south west){\large\(\bm{\cdots}\)};
        \node[fill=white, opacity=.8, text opacity=1]at(actor.center){Actor};
        \node[fill=white, opacity=.8, text opacity=1]at(critic.center){Critics};
        \node[below right]at(actor.south){$a^{\mathrm{low}}_t$};
        \node[right]at(smai.south east){SMAI};
        \node[below right](smai-n1)at(smai.north east){Task Confidence};
        \node[below left](smai-n2)at(smai.north west){Magnitude};
        \node[above]at([yshift=.5cm]actor.north west){$s^{\mathrm{low}}_t$};
        \node[above]at([yshift=.5cm]critic.north east){$s^{\mathrm{low}}_t, r^{\mathrm{low}}_t$};
        \node[below left]at([yshift=-.2cm]critic.south east){Curriculum Schedule};

        \draw[->, color=fforange_pv, line width=1pt, rounded corners=2pt](actor.south) --++ (0, -.5cm);
        \draw[<->, color=fforange_pv, line width=1pt, rounded corners=2pt]([yshift=.4cm]actor.east) -- ([yshift=.4cm]critic.west);
        \draw[->, color=ffdarkgreen, dashed, line width=1pt, rounded corners=2pt]([xshift=.2cm]L-llpolicy.north) |- (critic.west);
        \draw[->, color=ffdarkgreen, dashed, line width=1pt, rounded corners=2pt]([xshift=-.2cm]L-llpolicy.north) |- (actor.east);
        \draw[->, color=fforange_pv, line width=1pt, rounded corners=2pt]([xshift=-.2cm]ant.south) |- ([xshift=-.4cm, yshift=-.2cm]ant.south west) |- ([xshift=-.5cm,yshift=.5cm]actor.north) -- ([xshift=-.5cm]actor.north);
        \draw[->, color=fforange_pv, line width=1pt, rounded corners=2pt]([xshift=.2cm]ant.south) |- ([xshift=3.5cm, yshift=-.2cm]ant.south east) |- ([xshift=.5cm, yshift=.5cm]critic.north) -- ([xshift=.5cm]critic.north);
        \draw[->, color=fforange_pv, line width=1pt, rounded corners=2pt]([xshift=-.1cm, yshift=-.2cm]smai.south east) |- ([xshift=.5cm, yshift=.5cm]actor.north) -- ([xshift=.5cm]actor.north);
        \draw[->, color=fforange_pv, line width=1pt, rounded corners=2pt]([xshift=.1cm, yshift=-.2cm]smai.south east) |- ([xshift=-.5cm, yshift=.5cm]critic.north) -- ([xshift=-.5cm]critic.north);
        \draw[-, line width=1pt](smai-n1.west) --++ (-.5cm, -.2cm);
        \draw[-, line width=1pt](smai-n2.east) --++ (.5cm, -.1cm);

        \node[draw=ffgreen_pv, fill=ffgreen_pv!30, rounded corners=2pt, minimum width=19cm, minimum height=7cm, above right](bg3)at([xshift=.5cm, yshift=.75cm]bg0.south west){};
        
        \node[inner sep=0pt, fill=white, fill opacity=0.5, text opacity=1, left](dpmm2)at([xshift=-4cm, yshift=2cm]bg3.center){\includegraphics[width=2.5cm]{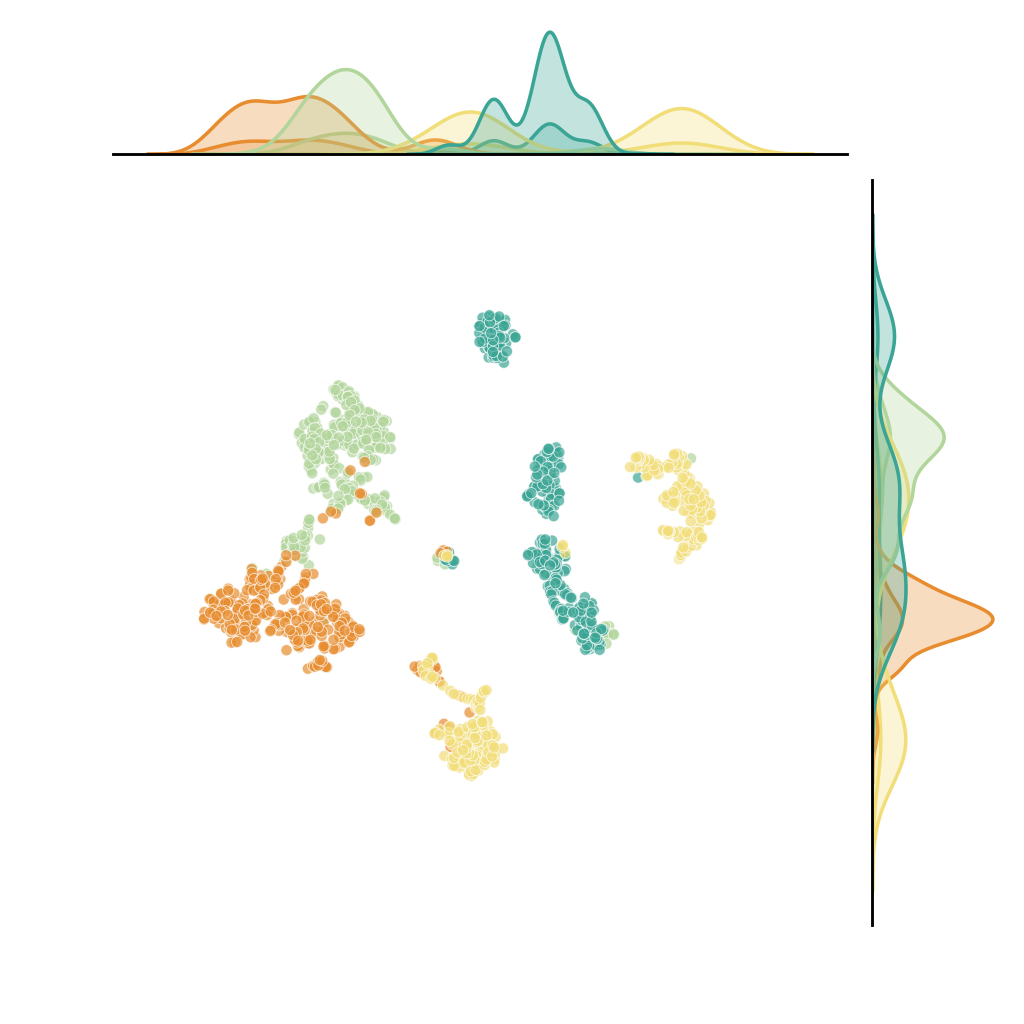}};
        \node[trapezium, fill=ffgreen_pv!50, draw=ffgreen_pv, line width=1pt, rounded corners=2pt, minimum width=1.5cm, minimum height=1cm, rotate=-90,](encoder2)at([xshift=-1cm]dpmm2.west){};
        \node[rectangle, fill=ffgreen_pv!50, dashed, draw=ffgreen_pv, line width=1pt, rounded corners=2pt, minimum width=3cm, minimum height=.5cm, right](deco-rew2)at([xshift=.5cm]dpmm2.east){\textcolor{ffgreen_pv!50}{Reward Generation}};
        \node[rectangle, fill=ffgreen_pv!50, dashed, draw=ffgreen_pv, line width=1pt, rounded corners=2pt, minimum width=3cm, minimum height=.5cm, above](deco-state2)at([yshift=.1cm]deco-rew2.north){\textcolor{ffgreen_pv!50}{State Prediction}};
        \node[rectangle, fill=ffgreen_pv!50, draw=ffgreen_pv, line width=1pt, rounded corners=2pt, minimum width=3cm, minimum height=.5cm, below](deco-task2)at([yshift=-.1cm]deco-rew2.south){Task Confidence};
        \draw[line width=1pt, dashed, color=ffgreen_pv, rounded corners=2pt] ([xshift=-.1cm, yshift=.1cm]deco-state2.north west) rectangle ([xshift=.1cm, yshift=-.1cm]deco-task2.south east);
        \node[rectangle, fill=ffgreen_pv!50, draw=ffgreen_pv, line width=1pt, rounded corners=2pt, minimum width=1.5cm, minimum height=1cm, below right](hlpolicy2)at([yshift=-2cm]deco-task2.south west){High-level Policy};

        \node[rectangle, fill=fflightgreen!50, draw=fflightgreen, line width=1pt, rounded corners=2pt, minimum width=2cm, minimum height=.5cm, above](map)at([yshift=.25cm]bg3.south){Observation Mapping};
        \node[rectangle, shading=axis, top color=fforange_pv!50, bottom color=ffgreen_pv!50, draw=ffgreen_pv, line width=1pt, rounded corners=2pt, minimum width=2cm, minimum height=1cm, text width=1.5cm, align=center, right](stride)at([xshift=1.5cm]bg3.center){Stride \\ Predictor};
        \node[rectangle, fill=ffgreen_pv!10, draw=ffblue, line width=1pt, rounded corners=2pt, minimum width=1cm, minimum height=.5cm, above](L-stride)at([yshift=.5cm]stride.north){$\mathcal{L}_{\mathcal{P}}$};
        \node[inner sep=0pt, left](smai2)at([xshift=-.5cm]stride.west){\includegraphics[width=1.5cm]{imgs/smai.png}};
        \node[inner sep=0pt, opacity=.5]at(encoder2.center){\includegraphics[width=.5cm]{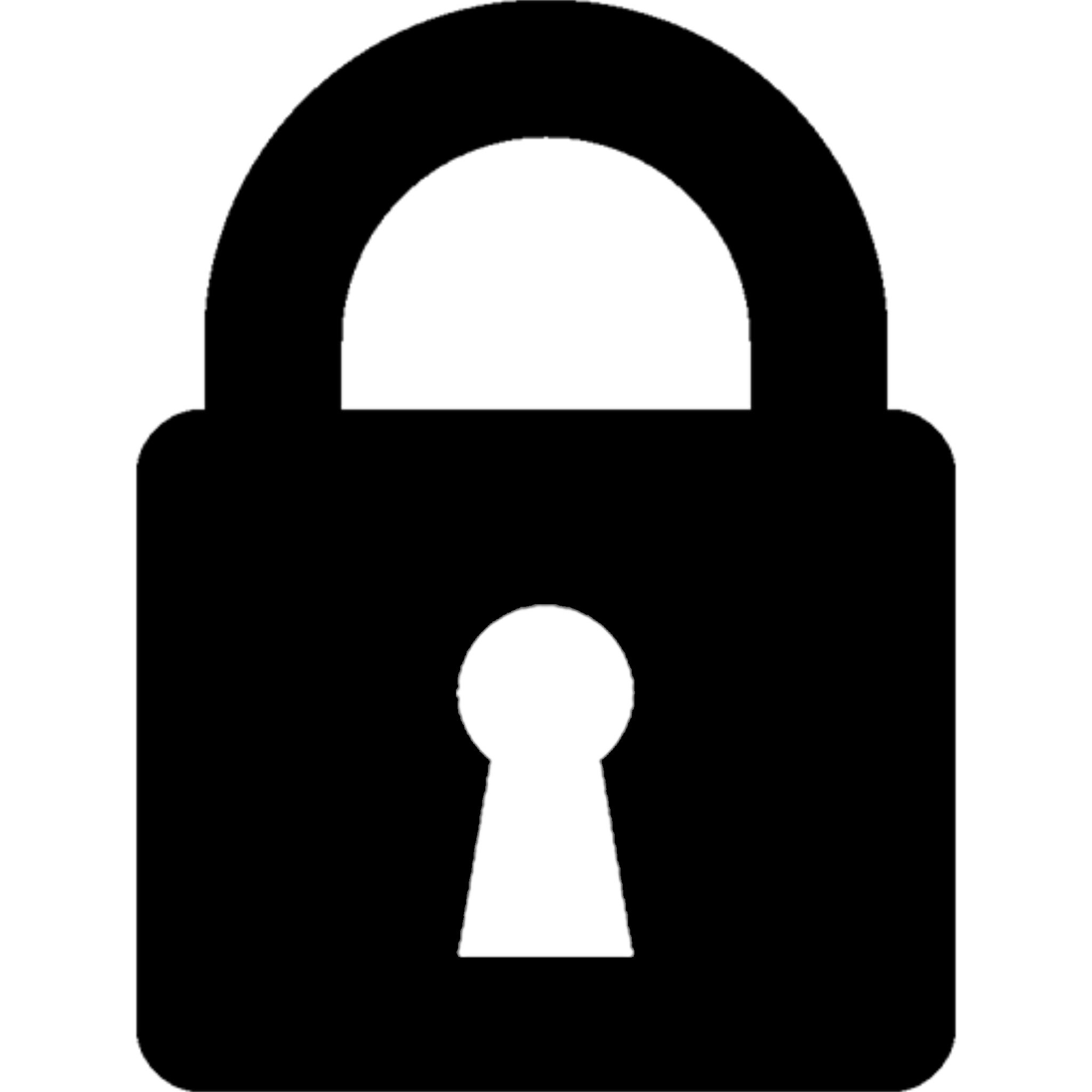}};
        \node[inner sep=0pt, opacity=.5]at(deco-rew2.center){\includegraphics[width=.5cm]{imgs/lock.png}};
        \node[inner sep=0pt, opacity=.5]at(hlpolicy2.north west){\includegraphics[width=.5cm]{imgs/lock.png}};

        \node[inner sep=0pt, draw=ffgreen_pv, fill=white, fill opacity=.5, text opacity=1, line width=1pt, rounded corners=2pt, right](actor1)at([xshift=5.5cm, yshift=2cm]bg3.center){\includegraphics[width=1.5cm]{imgs/neural_network.png}};
        \node[inner sep=0pt, right](ant1)at([xshift=.1cm]actor1.east){\includegraphics[width=1.5cm]{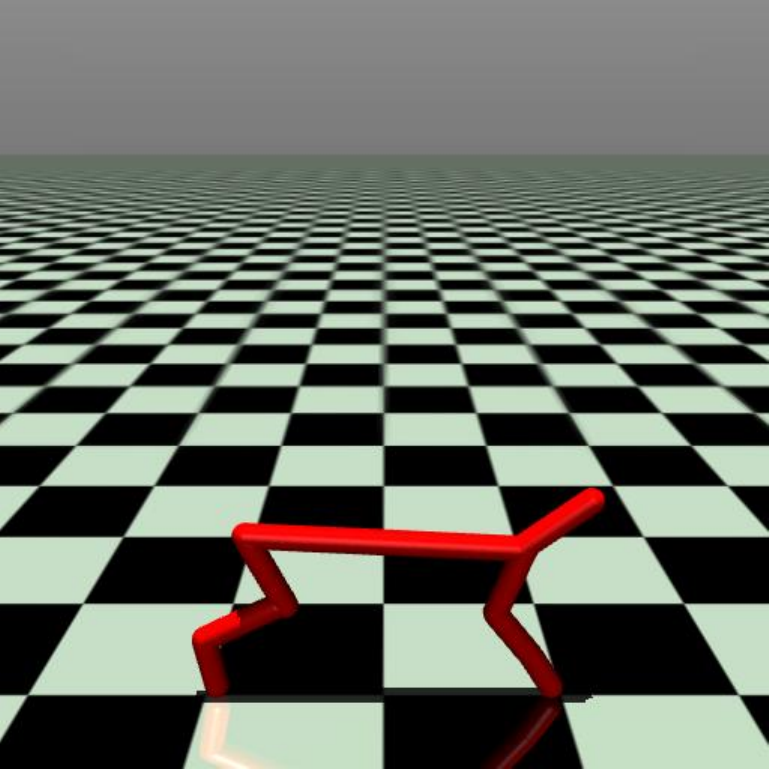}};
        \node[inner sep=0pt, draw=ffgreen_pv, fill=white, fill opacity=.5, text opacity=1, line width=1pt, rounded corners=2pt, right](actor2)at([xshift=5.5cm]bg3.center){\includegraphics[width=1.5cm]{imgs/neural_network.png}};
        \node[inner sep=0pt, right](cheetah)at([xshift=.1cm]actor2.east){\includegraphics[width=1.5cm]{imgs/ant.pdf}};
        \node[inner sep=0pt, draw=ffgreen_pv, fill=white, fill opacity=.5, text opacity=1, line width=1pt, rounded corners=2pt, right](actor3)at([xshift=5.5cm, yshift=-2cm]bg3.center){\includegraphics[width=1.5cm]{imgs/neural_network.png}};
        \node[inner sep=0pt, right](walker)at([xshift=.1cm]actor3.east){\includegraphics[width=1.5cm]{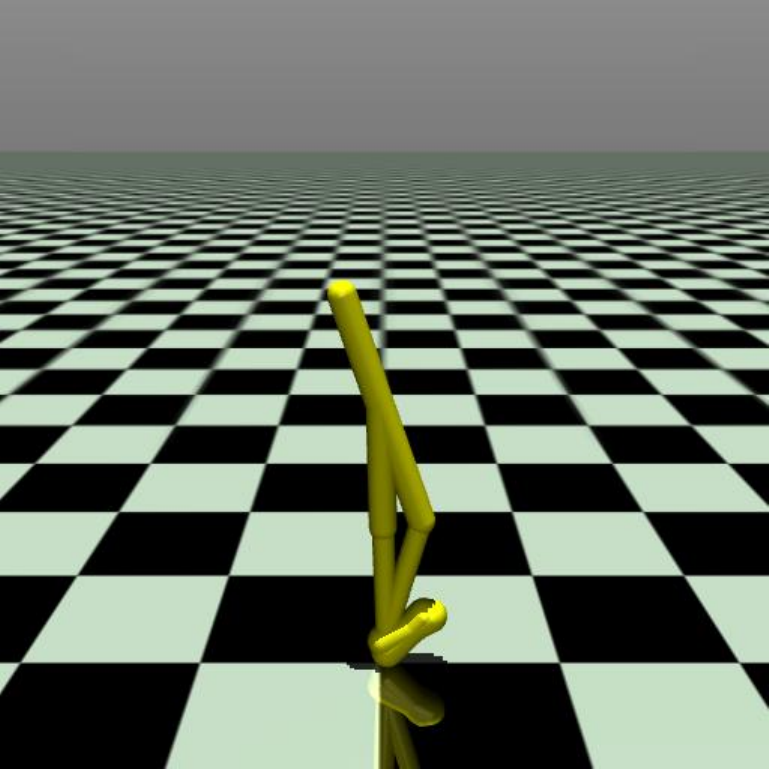}};
        \node[inner sep=0pt, opacity=.5]at(actor1.north west){\includegraphics[width=.5cm]{imgs/lock.png}};
        \node[inner sep=0pt, opacity=.5]at(actor2.north west){\includegraphics[width=.5cm]{imgs/lock.png}};
        \node[inner sep=0pt, opacity=.5]at(actor3.north west){\includegraphics[width=.5cm]{imgs/lock.png}};
        
        \draw[->, color=ffgreen_pv, line width=1pt, rounded corners=2pt](encoder2.north) -- (dpmm2.west);
        \draw[->, color=ffgreen_pv, line width=1pt, rounded corners=2pt]([xshift=-.5cm]deco-task2.west) -- (deco-task2.west);
        \draw[->, color=ffgreen_pv, line width=1pt, rounded corners=2pt](dpmm2.south) |- (hlpolicy2.west);
        \draw[->, color=ffgreen_pv, line width=1pt, rounded corners=2pt](map.west) -| ([xshift=-1cm, yshift=.5cm]encoder2.south) -- ([yshift=.5cm]encoder2.south);
        \draw[->, color=ffgreen_pv, line width=1pt, rounded corners=2pt](map.west) -| ([xshift=-1cm]encoder2.south) -- (encoder2.south);
        \draw[->, color=ffgreen_pv, line width=1pt, rounded corners=2pt](map.west) -| ([xshift=-1cm, yshift=-.5cm]encoder2.south) -- ([yshift=-.5cm]encoder2.south);
        \draw[->, color=ffgreen_pv, line width=1pt, rounded corners=2pt](map.west) -| (hlpolicy2.south);
        \draw[->, color=ffgreen_pv, line width=1pt, rounded corners=2pt](ant1.east) -| ([xshift=.5cm]walker.south east) |- (map.east);
        \draw[->, color=ffgreen_pv, line width=1pt, rounded corners=2pt](cheetah.east) -| ([xshift=.5cm]walker.south east) |- (map.east);
        \draw[->, color=ffgreen_pv, line width=1pt, rounded corners=2pt](walker.east) -| ([xshift=.5cm]walker.south east) |- (map.east);
        \draw[->, color=ffgreen_pv, line width=1pt, rounded corners=2pt](stride.east) --++ (.5cm, 0cm) circle[radius=2pt] --++ (.5cm, .5cm) circle[radius=2pt] |- (actor1.west);
        \draw[->, color=ffgreen_pv, line width=1pt, rounded corners=2pt]([xshift=1cm, yshift=-.5cm]stride.east) circle[radius=2pt] |- (actor3.west);
        \draw[->, color=ffgreen_pv, line width=1pt, rounded corners=2pt]([xshift=1.2cm]stride.east) circle[radius=2pt] -- (actor2.west);
        \draw[->, color=ffgreen_pv, line width=1pt, rounded corners=2pt](stride.east) --++ (.45cm, 0cm);
        \draw[->, color=ffgreen_pv, line width=1pt, rounded corners=2pt](hlpolicy2.east) -| ([xshift=-.25cm,yshift=.25cm]smai2.south);
        \draw[->, color=ffgreen_pv, line width=1pt, rounded corners=2pt](deco-task2.east) -| ([xshift=-.5cm, yshift=.5cm]smai2.south east);
        \draw[->, color=ffgreen_pv, line width=1pt, rounded corners=2pt](smai2.east) -- (stride.west);
        \draw[->, color=ffblue, dashed, line width=1pt, rounded corners=2pt](L-stride.south) -- (stride.north);

        \node[below]at([yshift=-.2cm]encoder2.east){Encoder};
        \node[below]at([yshift=-.1cm]deco-task2.south){Decoder};
        \node[above]at([yshift=.1cm]dpmm2.south){DPMM Prior};
        \node[above left](s)at([xshift=-.5cm]encoder2.south){$s$};
        \node[above left]at([xshift=-.4cm,yshift=.5cm]encoder2.south){$s'$};
        \node[above left]at([xshift=-.5cm,yshift=-.5cm]encoder2.south){$r$};
        \node[above left]at([xshift=-.25cm]hlpolicy2.west){$\bar{z}_t$};
        \node[below right]at(hlpolicy2.south){$s_t$};
        \node[above right]at(hlpolicy2.east){$a^{\mathrm{high}}_t$};
        \node[above right]at([xshift=.25cm]deco-task2.east){$\hat{\mathbf{y}}_t$};
        \node[below]at(smai2.south east){SMAI};
        \node[above]at([xshift=.5cm]stride.north east){$k_t$, $g_t$};
        \node[above right]at([xshift=.25cm]map.east){$s^{\mathrm{low}}_t, r^{\mathrm{low}}_t$};
        \node[below right]at(cheetah.south west){\large\(\bm{\cdots}\)};
        \node[below]at(ant1.north){\textcolor{white}{Agent 1}};
        \node[below]at(cheetah.north){\textcolor{white}{Agent 2}};
        \node[below]at(walker.north){\textcolor{white}{Agent $i$}};
        \node[below, text width=1.75cm, align=center]at(stride.south){\scriptsize few shot online \\ adaptation};
        \node[below](task)at([yshift=-1cm]encoder2.east){Meta task inference:};
        \node[fill=ffgreen_pv!80, below right, label={[label distance=0pt]0:\textsf{task 1}}](l1)at([xshift=.1cm,yshift=-.25cm]task.south west){};
        \node[fill=fflightgreen!80, below, label={[label distance=0pt]0:\textsf{task 2}}](l2)at([yshift=-.25cm]l1.south){};
        \node[fill=ffyellow!80, below, label={[label distance=0pt]0:\textsf{task 3}}](l3)at([yshift=-.25cm]l2.south){};
        \node[fill=fforange_pv!80, below, label={[label distance=0pt]0:\textsf{task n}}](l4)at([yshift=-.25cm]l3.south){};

        \node[below right]at(bg1.north west){\large\textbf{a}};
        \node[below right]at(bg2-3.north west){\large\textbf{b}};
        \node[below right]at(bg3.north west){\large\textbf{c}};
        
    \end{tikzpicture}
    }
    \vskip -.05in
    \caption{Overview of ReMAP. 
    \textbf{a}, The framework first learns reusable task-level meta-knowledge on a dynamics-simplified agent through DPMM-based task inference and high-level policy learning, while 
    \textbf{b}, independently warming up embodiment-specific low-level policies with SMAI-guided curriculum adaptive learning. 
    \textbf{c}, During deployment and meta-knowledge reutilization, the frozen meta-knowledge module generates semantic-magnitude subgoals, which are temporally aligned by a stride predictor and executed by different target agents with lightweight online adaptation.}
    \label{fig:framework}
\vskip -.15in
\end{figure*}


%% file: img_tex/simplified_agent_overview.tex
\begin{figure}[t]
    \centering
    \resizebox{\linewidth}{!}{
    \begin{tikzpicture}[font=\sffamily]
        \node[inner sep=0pt](a)at(0, 0){\includegraphics[width=5cm]{imgs/toy.pdf}};
        \node[inner sep=0pt, right](b)at([xshift=.25cm]a.east){\includegraphics[width=5cm]{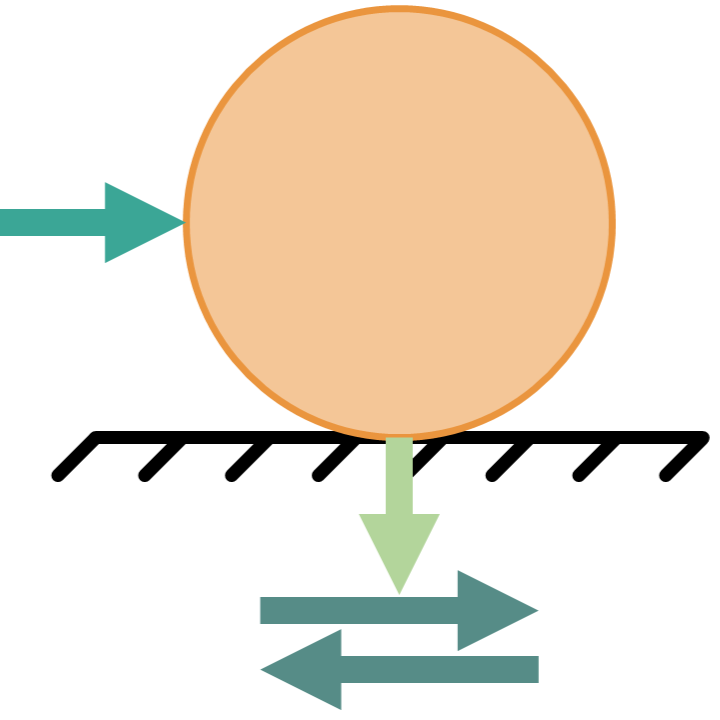}};
        \node[below right]at(a.north west){\large\textbf{a}};
        \node[below right]at(b.north west){\large\textbf{b}};
        \node[]at([xshift=1cm, yshift=.25cm]b.west){$F_{ext}$};
        \node[]at([xshift=1.25cm, yshift=1.25cm]b.south){$G=mg$};
        \node[]at([xshift=-1.5cm, yshift=.5cm]b.south){$f_{friction}$};
        \node[below, align=center, text width=1.5cm]at([xshift=.2cm, yshift=-1.25cm]b.north){Simplified Agent};
    \end{tikzpicture}
    }
    \vskip -.05in
    \caption{
    Simplified agent modeling.
    \textbf{a}, Illustration of the simplified agent instantiation in the Mujoco environment. 
    \textbf{b}, Abstracted mass--damper modeling with external, gravity, and friction forces corresponding to Eq.~(\ref{eq:toy_dynamic}).
    }
    \vskip -0.15in
    \label{fig:toy_agent}
\end{figure}

%% file: img_tex/env_setup.tex
\begin{figure}[ht!]
    \centering
    \resizebox{\linewidth}{!}{
    \begin{tikzpicture}[
        font=\sffamily,
        velarrow/.style={-{Latex[length=2.0mm]}, thick, red},
        goalarrow/.style={-{Latex[length=2.0mm]}, thick, blue},
        labelstyle/.style={font=\small\sffamily, align=center}
    ]

    \def\imgw{0.30\linewidth}

    \begin{scope}[shift={(0,0)}]
        \node[inner sep=0pt] (img1) at (0,0) {\includegraphics[width=\imgw]{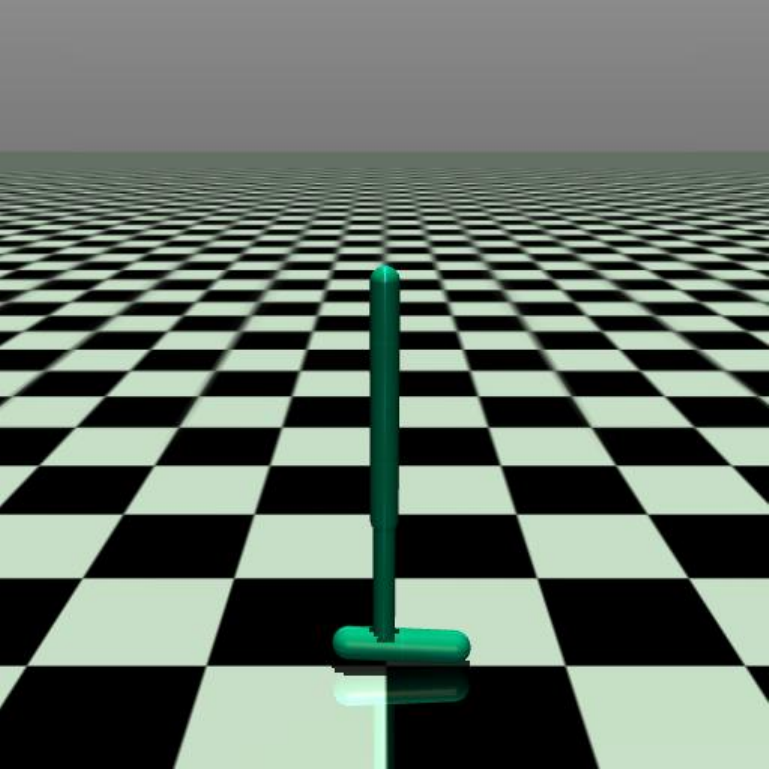}};

        \draw[->, line width=1pt, color=ffyellow] (-0.25,0.75) -- (0.55,0.75);
        \draw[->, line width=1pt, color=fforange_pv] (0.25,0.55) -- (-0.55,0.55);
        \node[inner sep=0pt, fill=white, fill opacity=.8, text opacity=1, left](gf)at([xshift=-.1cm]img1.east){\includegraphics[width=.5cm]{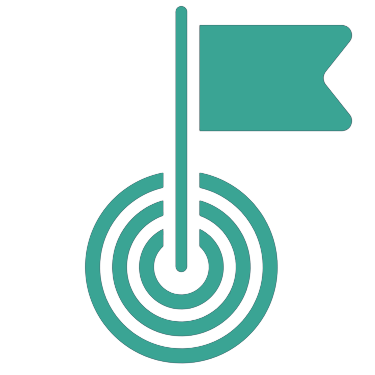}};
        \node[inner sep=0pt, fill=white, fill opacity=.8, text opacity=1, right](gb)at([xshift=.1cm]img1.west){\includegraphics[width=.5cm]{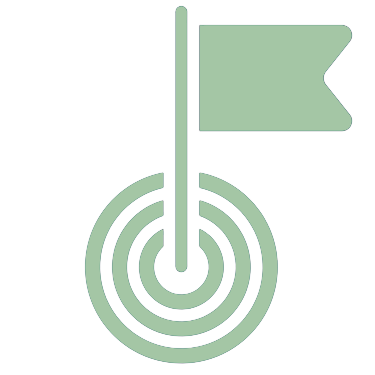}};

        \node[labelstyle, below] at (img1.south) {Hopper};
    \end{scope}

    \begin{scope}[shift={(2.75cm,0)}]
        \node[inner sep=0pt] (img2) at (0,0) {\includegraphics[width=\imgw]{imgs/walker.pdf}};

        \draw[->, line width=1pt, color=ffyellow] (-0.25,0.75) -- (0.55,0.75);
        \draw[->, line width=1pt, color=fforange_pv] (0.25,0.55) -- (-0.55,0.55);
        \node[inner sep=0pt, fill=white, fill opacity=.8, text opacity=1, left](gf)at([xshift=-.1cm]img2.east){\includegraphics[width=.5cm]{imgs/goal-flag-ffgreen.png}};
        \node[inner sep=0pt, fill=white, fill opacity=.8, text opacity=1, right](gb)at([xshift=.1cm]img2.west){\includegraphics[width=.5cm]{imgs/goal-flag-fflightgreen.png}};

        \node[labelstyle, below] at (img2.south) {Walker};
    \end{scope}

    \begin{scope}[shift={(5.5cm,0)}]
        \node[inner sep=0pt] (img3) at (0,0) {\includegraphics[width=\imgw]{imgs/cheetah.pdf}};

        \draw[->, line width=1pt, color=ffyellow] (-0.25,0.75) -- (0.55,0.75);
        \draw[->, line width=1pt, color=fforange_pv] (0.25,0.55) -- (-0.55,0.55);
        \node[inner sep=0pt, fill=white, fill opacity=.8, text opacity=1, left](gf)at([xshift=-.1cm]img3.east){\includegraphics[width=.5cm]{imgs/goal-flag-ffgreen.png}};
        \node[inner sep=0pt, fill=white, fill opacity=.8, text opacity=1, right](gb)at([xshift=.1cm]img3.west){\includegraphics[width=.5cm]{imgs/goal-flag-fflightgreen.png}};

        \node[labelstyle, below] at (img3.south) {Half-Cheetah};
    \end{scope}

    \begin{scope}[shift={(8.25cm,0)}]
        \node[inner sep=0pt] (img4) at (0,0) {\includegraphics[width=\imgw]{imgs/ant.pdf}};

        \draw[->, line width=1pt, color=ffyellow] (-0.25,0.75) -- (0.55,0.75);
        \draw[->, line width=1pt, color=fforange_pv] (0.25,0.55) -- (-0.55,0.55);
        \node[inner sep=0pt, fill=white, fill opacity=.8, text opacity=1, left](gf)at([xshift=-.1cm]img4.east){\includegraphics[width=.5cm]{imgs/goal-flag-ffgreen.png}};
        \node[inner sep=0pt, fill=white, fill opacity=.8, text opacity=1, right](gb)at([xshift=.1cm]img4.west){\includegraphics[width=.5cm]{imgs/goal-flag-fflightgreen.png}};

        \node[labelstyle, below] at (img4.south) {Ant};
    \end{scope}

    \end{tikzpicture}
    }
    \vskip -.05in
    \caption{
    Cross-agent benchmark for meta-knowledge reuse.
    We evaluate four agents with different embodiments: \textit{Hopper}, \textit{Walker}, \textit{Half-Cheetah}, and \textit{Ant}.
    Each agent is tested on four locomotion tasks, including forward/backward goal reaching (flags) and forward/backward velocity tracking (arrows).
    }
    \label{fig:cross_agent_tasks}
\vskip -.1in
\end{figure}

%% file: img_tex/toy_results.tex
\begin{figure*}[t!]
    \centering
    \resizebox{.95\textwidth}{!}{
    \begin{tikzpicture}[font=\sffamily]
        \pgfplotsset{
            toyaxis/.style={
                tick label style={font=\sffamily\scriptsize},
                label style={font=\sffamily\small},
                title style={font=\sffamily\small},
                grid=major
            }
        }

        \begin{axis}[
            toyaxis,
            name=rewardplot,
            at={(0,0)},
            anchor=north west,
            width=5.5cm,
            height=5.7cm,
            title={Simplified Agent},
            xlabel={Epochs},
            ylabel={Average Reward},
            xmin=0, xmax=1000,
            ymin=-250, ymax=-25,
            ytick={-250, -200, -150, -100, -50},
        ]
            \pgfplotstableread[col sep=comma]{data/toy-reward.csv}\datatable
            \addplot [ffgreen_pv!80, line width=1.2pt] table [x=epoch, y=mean_smooth] {\datatable};
            \addplot [name path=upper, draw=none, forget plot] table[x=epoch, y=upper_smooth] {\datatable};
            \addplot [name path=lower, draw=none, forget plot] table[x=epoch, y=lower_smooth] {\datatable};
            \addplot [fill=ffgreen_pv!50, forget plot, opacity=0.3] fill between[of=upper and lower];
        \end{axis}

        \begin{axis}[
            toyaxis,
            name=posplot,
            at={(5.25cm,0)},
            anchor=north west,
            width=5.5cm,
            height=3.5cm,
            title={Inference Trajectories},
            xmin=0, xmax=60,
            ymin=-11.5, ymax=11.5,
            ylabel={Position (m)},
            ylabel style={yshift=-.2cm},
            xticklabels=\empty,
        ]

            \draw[fflightgreen, dashed, line width=1.2pt] (axis cs:0,-10) -- (axis cs:60,-10);
            \draw[fflightgreen!80, dashed, line width=1.2pt] (axis cs:0,-7.55) -- (axis cs:60,-7.55);
            \draw[fflightgreen!60, dashed, line width=1.2pt] (axis cs:0,-5.1) -- (axis cs:60,-5.1);
            
            \draw[ffgreen_pv!60, dashed, line width=1.2pt] (axis cs:0,5.1) -- (axis cs:60,5.1);
            \draw[ffgreen_pv!80, dashed, line width=1.2pt] (axis cs:0,7.55) -- (axis cs:60,7.55);
            \draw[ffgreen_pv, dashed, line width=1.2pt] (axis cs:0,10.0) -- (axis cs:60,10.0);

            \addplot[fflightgreen, line width=1.2pt]table[col sep=comma, x=t, y=x_before] {data/high_level_policy/smooth/subgoals_ep0_smooth.csv};
            \addplot[fflightgreen!80, line width=1.2pt] table[col sep=comma, x=t, y=x_before] {data/high_level_policy/smooth/subgoals_ep1_smooth.csv};
            \addplot[fflightgreen!60, line width=1.2pt] table[col sep=comma, x=t, y=x_before] {data/high_level_policy/smooth/subgoals_ep2_smooth.csv};

            \addplot[ffgreen_pv!60, line width=1.2pt] table[col sep=comma, x=t, y=x_before] {data/high_level_policy/smooth/subgoals_ep3_smooth.csv};
            \addplot[ffgreen_pv!80, line width=1.2pt] table[col sep=comma, x=t, y=x_before] {data/high_level_policy/smooth/subgoals_ep4_smooth.csv};
            \addplot[ffgreen_pv, line width=1.2pt] table[col sep=comma, x=t, y=x_before] {data/high_level_policy/smooth/subgoals_ep5_smooth.csv};

        \end{axis}

        \begin{axis}[
            toyaxis,
            name=velplot,
            at={(posplot.south west)},
            anchor=north west,
            yshift=-0.25cm,
            width=5.5cm,
            height=3.5cm,
            xmin=0, xmax=60,
            ymin=-3, ymax=3,
            xlabel={Time Step},
            ylabel={Velocity (m/s)},
        ]

            \draw[fforange_pv, dashed, line width=1.2pt] (axis cs:0,-2.5) -- (axis cs:60,-2.5);
            \draw[fforange_pv!80, dashed, line width=1.2pt] (axis cs:0,-1.75) -- (axis cs:60,-1.75);
            \draw[fforange_pv!60, dashed, line width=1.2pt] (axis cs:0,-1.375) -- (axis cs:60,-1.375);
            \draw[ffyellow!60, dashed, line width=1.2pt] (axis cs:0,1.375) -- (axis cs:60,1.375);
            \draw[ffyellow!80, dashed, line width=1.2pt] (axis cs:0,1.75) -- (axis cs:60,1.75);
            \draw[ffyellow, dashed, line width=1.2pt] (axis cs:0,2.5) -- (axis cs:60,2.5);

            \addplot[fforange_pv, line width=1.2pt]
                table[col sep=comma, x=t, y=vx_before]
                {data/high_level_policy/smooth/subgoals_ep6_smooth.csv};

            \addplot[fforange_pv!80, line width=1.2pt]
                table[col sep=comma, x=t, y=vx_before]
                {data/high_level_policy/smooth/subgoals_ep7_smooth.csv};

            \addplot[fforange_pv!60, line width=1.2pt]
                table[col sep=comma, x=t, y=vx_before]
                {data/high_level_policy/smooth/subgoals_ep8_smooth.csv};

            \addplot[ffyellow!60, line width=1.2pt]
                table[col sep=comma, x=t, y=vx_before]
                {data/high_level_policy/smooth/subgoals_ep9_smooth.csv};

            \addplot[ffyellow!80, line width=1.2pt]
                table[col sep=comma, x=t, y=vx_before]
                {data/high_level_policy/smooth/subgoals_ep10_smooth.csv};

            \addplot[ffyellow, line width=1.2pt]
                table[col sep=comma, x=t, y=vx_before]
                {data/high_level_policy/smooth/subgoals_ep11_smooth.csv};

        \end{axis}

        \node[inner sep=0pt, anchor=north west] (latentimg) at (11.5cm,.75cm) {\includegraphics[width=5.5cm]{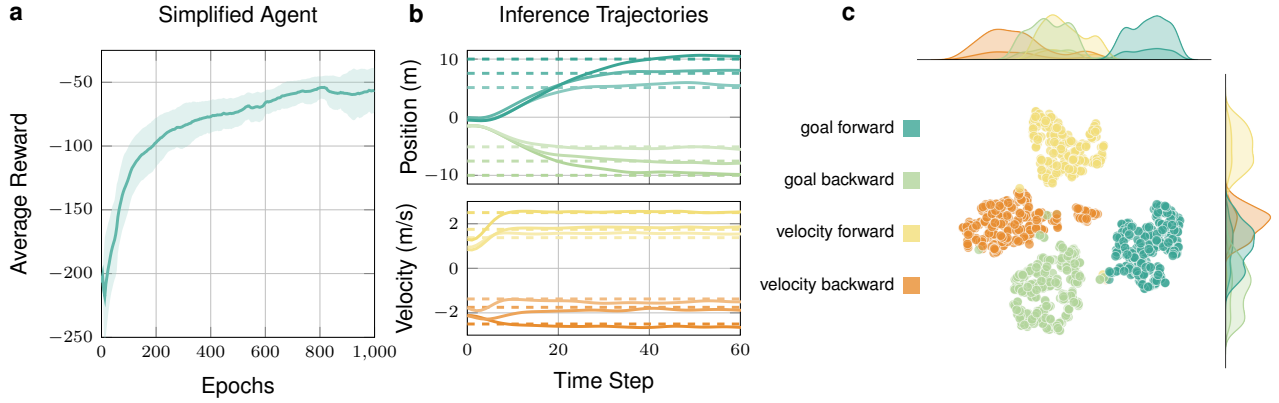}};
        \node[fill=ffgreen_pv!80, below left, label={[label distance=0pt]180:\scriptsize\textsf{goal forward}}](l1)at([xshift=0.25cm, yshift=-1.75cm]latentimg.north west){};
        \node[fill=fflightgreen!80, below, label={[label distance=0pt]180:\scriptsize\textsf{goal backward}}](l2)at([yshift=-.5cm]l1.south){};
        \node[fill=ffyellow!80, below, label={[label distance=0pt]180:\scriptsize\textsf{velocity forward}}](l3)at([yshift=-.5cm]l2.south){};
        \node[fill=fforange_pv!80, below, label={[label distance=0pt]180:\scriptsize\textsf{velocity backward}}](l4)at([yshift=-.5cm]l3.south){};
        
        \node[below right] (c) at ([xshift=-1cm]latentimg.north west) {\textbf{c}};
        \node[left] (b) at ([xshift=-5.75cm]c.west){\textbf{b}};
        \node[left] (a) at ([xshift=-11.5cm]c.west){\textbf{a}};

    \end{tikzpicture}
    }
    \vskip -.05in
    \caption{
    Evaluation of the simplified agent on four non-parametric tasks.
    \textbf{a}, Average reward during meta-knowledge acquisition over four task scenarios, including goal tracking forward/backward and velocity tracking forward/backward. The curve reports the mean reward with standard deviation over at least five trials ($\mu\pm\sigma$).
    \textbf{b}, Inference trajectories on twelve task variations, where dashed horizontal lines denote the ground-truth targets. The inference module and high-level policy can accurately track both position (above) and velocity (below) targets in both directions.
    \textbf{c}, Visualization of the learned latent space, where different task types form separable clusters, indicating that the inference module captures the underlying task structure.
    }
    \label{fig:toy_multi_results}
\vskip -.15in
\end{figure*}

%% file: img_tex/all_agents_reward_multi.tex
\begin{figure*}[t!]
    \centering
    \resizebox{.95\textwidth}{!}{
    \begin{tikzpicture}[font=\sffamily]

        \pgfplotsset{
            rewardaxis/.style={
                width=4.5cm,
                height=4cm,
                grid=major,
                xlabel={Epochs},
                ylabel={Average Reward},
                tick label style={font=\scriptsize\sffamily},
                label style={font=\small\sffamily},
                title style={font=\small\sffamily},
            }
        }

        \begin{axis}[
            rewardaxis,
            name=cheetah,
            at={(0,0)},
            anchor=north west,
            xmin=0, xmax=300,
            title={Half-Cheetah},
        ]
        \pgfplotstableread[col sep=comma]{data/reward_curve_original_data/cheetah_reward.csv}\cheetahdata

        \addplot [ffgreen_pv!80, line width=1pt]
            table [x=epoch, y=mean_smooth] {\cheetahdata};
        \addplot [name path=cheetahupper, draw=none, forget plot]
            table [x=epoch, y=upper_smooth] {\cheetahdata};
        \addplot [name path=cheetahlower, draw=none, forget plot]
            table [x=epoch, y=lower_smooth] {\cheetahdata};
        \addplot [fill=ffgreen_pv!50, forget plot, opacity=0.3]
            fill between[of=cheetahupper and cheetahlower];
        \end{axis}

        \begin{axis}[
            rewardaxis,
            name=ant,
            at={($(cheetah.north east)+(1.25cm,0cm)$)},
            anchor=north west,
            xmin=0, xmax=300,
            title={Ant},
        ]
        \pgfplotstableread[col sep=comma]{data/reward_curve_original_data/ant_reward.csv}\antdata

        \addplot [ffgreen_pv!80, line width=1pt]
            table [x=epoch, y=mean_smooth] {\antdata};
        \addplot [name path=antupper, draw=none, forget plot]
            table [x=epoch, y=upper_smooth] {\antdata};
        \addplot [name path=antlower, draw=none, forget plot]
            table [x=epoch, y=lower_smooth] {\antdata};
        \addplot [fill=ffgreen_pv!50, forget plot, opacity=0.3]
            fill between[of=antupper and antlower];
        \end{axis}

        \begin{axis}[
            rewardaxis,
            name=walker,
            at={($(ant.north east)+(1.25cm,0cm)$)},
            anchor=north west,
            xmin=0, xmax=3000,
            title={Walker},
        ]
        \pgfplotstableread[col sep=comma]{data/reward_curve_original_data/walker_reward.csv}\walkerdata

        \addplot [ffgreen_pv!80, line width=1pt]
            table [x=epoch, y=mean_smooth] {\walkerdata};
        \addplot [name path=walkerupper, draw=none, forget plot]
            table [x=epoch, y=upper_smooth] {\walkerdata};
        \addplot [name path=walkerlower, draw=none, forget plot]
            table [x=epoch, y=lower_smooth] {\walkerdata};
        \addplot [fill=ffgreen_pv!50, forget plot, opacity=0.3]
            fill between[of=walkerupper and walkerlower];
        \end{axis}

        \begin{axis}[
            rewardaxis,
            name=hopper,
            at={($(walker.north east)+(1.25cm,0cm)$)},
            anchor=north west,
            xmin=0, xmax=3000,
            title={Hopper},
        ]
        \pgfplotstableread[col sep=comma]{data/reward_curve_original_data/hopper_reward.csv}\hopperdata

        \addplot [ffgreen_pv!80, line width=1pt]
            table [x=epoch, y=mean_smooth] {\hopperdata};
        \addplot [name path=hopperupper, draw=none, forget plot]
            table [x=epoch, y=upper_smooth] {\hopperdata};
        \addplot [name path=hopperlower, draw=none, forget plot]
            table [x=epoch, y=lower_smooth] {\hopperdata};
        \addplot [fill=ffgreen_pv!50, forget plot, opacity=0.3]
            fill between[of=hopperupper and hopperlower];
        \end{axis}

    \end{tikzpicture}
    }
    \vskip -.05in
    \caption{
    Disentangled low-level policy adaptive learning reward curves of four agents with complex dynamics.
    The solid line represents the average reward, while the shaded region denotes the standard deviation across five trials $(\mu \pm \sigma)$.
    }
    \label{fig:four_agents_reward}
\vskip -.15in
\end{figure*}
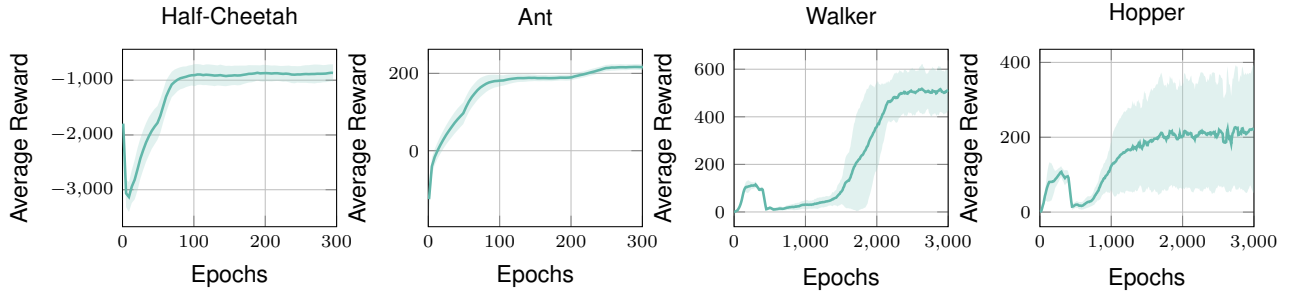

%% file: img_tex/single_task_inference_results.tex
\begin{figure}[t!]
    \centering
    \resizebox{\linewidth}{!}{
    \begin{tikzpicture}[font=\sffamily]

        \begin{axis}[
            name=goaltrack,
            at={(0,0)},
            anchor=north west,
            width=11cm,
            height=6cm,
            xlabel={Time Steps},
            ylabel={Position (m)},
            title={Goal Forward},
            grid=both,
            xmin=0,
            xmax=100,
        ]
            \addplot[ffgreen_pv, solid, line width=1.2pt]
            table[col sep=comma, x expr=\coordindex, y=current_goal, restrict x to domain=0:99]{data/single_task_inference_result/cheetah/cheetah_showcase_results/1.csv};
            \addplot[ffgreen_pv, dashed, line width=1.2pt]
            table[col sep=comma, x expr=\coordindex, y=target_goal, restrict x to domain=0:99]{data/single_task_inference_result/cheetah/cheetah_showcase_results/1.csv};
            
            \addplot[ffgreen_pv!80, solid, line width=1.2pt]
            table[col sep=comma, x expr=\coordindex, y=current_goal, restrict x to domain=0:99]{data/single_task_inference_result/cheetah/cheetah_showcase_results/6.csv};
            \addplot[ffgreen_pv!80, dashed, line width=1.2pt]
            table[col sep=comma, x expr=\coordindex, y=target_goal, restrict x to domain=0:99]{data/single_task_inference_result/cheetah/cheetah_showcase_results/6.csv};            
            
            \addplot[ffgreen_pv!60, solid, line width=1.2pt]
            table[col sep=comma, x expr=\coordindex, y=current_goal, restrict x to domain=0:99]{data/single_task_inference_result/cheetah/cheetah_showcase_results/2.csv};
            \addplot[ffgreen_pv!60, dashed, line width=1.2pt]
            table[col sep=comma, x expr=\coordindex, y=target_goal, restrict x to domain=0:99]{data/single_task_inference_result/cheetah/cheetah_showcase_results/2.csv};

            \addplot[fflightgreen, solid, line width=1.2pt]
            table[col sep=comma, x expr=\coordindex, y=current_goal, restrict x to domain=0:99]
            {data/single_task_inference_result/ant/ant_showcase_results/20.csv};

            \addplot[fflightgreen, dashed, line width=1.2pt]
            table[col sep=comma, x expr=\coordindex, y=target_goal, restrict x to domain=0:99]
            {data/single_task_inference_result/ant/ant_showcase_results/20.csv};
            
            \addplot[fflightgreen!80, solid, line width=1.2pt]
            table[col sep=comma, x expr=\coordindex, y=current_goal, restrict x to domain=0:99]
            {data/single_task_inference_result/ant/ant_showcase_results/4.csv};
            \addplot[fflightgreen!80, dashed, line width=1.2pt]
            table[col sep=comma, x expr=\coordindex, y=target_goal, restrict x to domain=0:99]
            {data/single_task_inference_result/ant/ant_showcase_results/4.csv};

            \addplot[fflightgreen!60, solid, line width=1.2pt]
            table[col sep=comma, x expr=\coordindex, y=current_goal, restrict x to domain=0:99]
            {data/single_task_inference_result/ant/ant_showcase_results/7.csv};
            \addplot[fflightgreen!60, dashed, line width=1.2pt]
            table[col sep=comma, x expr=\coordindex, y=target_goal, restrict x to domain=0:99]
            {data/single_task_inference_result/ant/ant_showcase_results/7.csv};

            \addplot[ffyellow, solid, line width=1.2pt]
            table[col sep=comma, x expr=\coordindex, y=current_goal, restrict x to domain=0:99]
            {data/single_task_inference_result/walker/walker_showcase_results/42.csv};
            \addplot[ffyellow, dashed, line width=1.2pt]
            table[col sep=comma, x expr=\coordindex, y=target_goal, restrict x to domain=0:99]
            {data/single_task_inference_result/walker/walker_showcase_results/42.csv};
            
            \addplot[ffyellow!80, solid, line width=1.2pt]
            table[col sep=comma, x expr=\coordindex, y=current_goal, restrict x to domain=0:99]
            {data/single_task_inference_result/walker/walker_showcase_results/4.csv};
            \addplot[ffyellow!80, dashed, line width=1.2pt]
            table[col sep=comma, x expr=\coordindex, y=target_goal, restrict x to domain=0:99]
            {data/single_task_inference_result/walker/walker_showcase_results/4.csv};

            \addplot[ffyellow!60, solid, line width=1.2pt]
            table[col sep=comma, x expr=\coordindex, y=current_goal, restrict x to domain=0:99]
            {data/single_task_inference_result/walker/walker_showcase_results/1.csv};
            \addplot[ffyellow!60, dashed, line width=1.2pt]
            table[col sep=comma, x expr=\coordindex, y=target_goal, restrict x to domain=0:99]
            {data/single_task_inference_result/walker/walker_showcase_results/1.csv};


            \addplot[fforange_pv, solid, line width=1.2pt]
            table[col sep=comma, x expr=\coordindex, y=current_goal, restrict x to domain=0:99]
            {data/single_task_inference_result/hopper/hopper_showcase_results/20.csv};
            \addplot[fforange_pv, dashed, line width=1.2pt]
            table[col sep=comma, x expr=\coordindex, y=target_goal, restrict x to domain=0:99]
            {data/single_task_inference_result/hopper/hopper_showcase_results/20.csv};

            \addplot[fforange_pv!80, solid, line width=1.2pt]
            table[col sep=comma, x expr=\coordindex, y=current_goal, restrict x to domain=0:99]
            {data/single_task_inference_result/hopper/hopper_showcase_results/33.csv};
            \addplot[fforange_pv!80, dashed, line width=1.2pt]
            table[col sep=comma, x expr=\coordindex, y=target_goal, restrict x to domain=0:99]
            {data/single_task_inference_result/hopper/hopper_showcase_results/33.csv};
            
            \addplot[fforange_pv!60, solid, line width=1.2pt]
            table[col sep=comma, x expr=\coordindex, y=current_goal, restrict x to domain=0:99]
            {data/single_task_inference_result/hopper/hopper_showcase_results/2.csv};
            \addplot[fforange_pv!60, dashed, line width=1.2pt]
            table[col sep=comma, x expr=\coordindex, y=target_goal, restrict x to domain=0:99]
            {data/single_task_inference_result/hopper/hopper_showcase_results/2.csv};

        \end{axis}

        \node[] at ([xshift=-0.75cm, yshift=0.25cm]goaltrack.north west) {\normalsize\textbf{a}};
        
        \begin{axis}[
            name=encodingplot,
            at={(goaltrack.south west)},
            yshift=-2cm,
            anchor=north west,
            width=11cm,
            height=6cm,
            xlabel={Target Goal},
            ylabel={Encoding},
            title={Encoding vs. Target Goal Across Agents},
            grid=both,
            legend style={
                font=\small\sffamily,
                draw=none,
                fill=none,
                at={(3.5cm,-1.5cm)},
                anchor=north,
                legend columns=4,
                /tikz/every even column/.append style={column sep=.5cm},
            }
        ]

            \foreach \i in {1,...,50}{
                \addplot[
                    ffgreen_pv,
                    opacity=0.18,
                    mark=*,
                    mark size=1.2pt,
                    only marks,
                    forget plot,
                    restrict expr to domain={\coordindex}{50:100000}
                ]
                table[
                    col sep=comma,
                    x=target_goal,
                    y=encoding,
                    each nth point=10
                ]
                {data/single_task_inference_result/cheetah/cheetah_single_task_inference_trajectories/\i.csv};
            }

            \addplot[
                ffgreen_pv!80,
                line width=1.2pt,
            ]
            table[
                col sep=comma,
                x=target_goal,
                y=encoding_regression
            ]
            {data/single_task_inference_result/cheetah_encoding_regression_line_downsampled.csv};
            \addlegendentry{Half-Cheetah}

            \foreach \i in {1,...,50}{
                \addplot[
                    fflightgreen,
                    opacity=0.18,
                    mark=*,
                    mark size=1.2pt,
                    only marks,
                    forget plot,
                    restrict expr to domain={\coordindex}{50:100000}
                ]
                table[
                    col sep=comma,
                    x=target_goal,
                    y=encoding,
                    each nth point=10
                ]
                {data/single_task_inference_result/ant/ant_single_task_inference_trajectories/\i.csv};
            }

            \addplot[
                fflightgreen!80,
                line width=1.2pt,
            ]
            table[
                col sep=comma,
                x=target_goal,
                y=encoding_regression
            ]
            {data/single_task_inference_result/ant_encoding_regression_line_downsampled.csv};
            \addlegendentry{Ant}

            \foreach \i in {1,...,50}{
                \addplot[
                    ffyellow,
                    opacity=0.18,
                    mark=*,
                    mark size=1.2pt,
                    only marks,
                    forget plot,
                    restrict expr to domain={\coordindex}{50:100000}
                ]
                table[
                    col sep=comma,
                    x=target_goal,
                    y=encoding,
                    each nth point=10
                ]
                {data/single_task_inference_result/walker/walker_single_task_inference_trajectories/\i.csv};
            }

            \addplot[
                ffyellow!80,
                line width=1.2pt,
            ]
            table[
                col sep=comma,
                x=target_goal,
                y=encoding_regression
            ]
            {data/single_task_inference_result/walker_encoding_regression_line_downsampled.csv};
            \addlegendentry{Walker}

            \foreach \i in {1,...,50}{
                \addplot[
                    fforange_pv,
                    opacity=0.18,
                    mark=*,
                    mark size=1.2pt,
                    only marks,
                    forget plot,
                    restrict expr to domain={\coordindex}{50:100000}
                ]
                table[
                    col sep=comma,
                    x=target_goal,
                    y=encoding,
                    each nth point=10
                ]
                {data/single_task_inference_result/hopper/hopper_single_task_inference_trajectories/\i.csv};
            }

            \addplot[
                fforange_pv!80,
                line width=1.2pt,
            ]
            table[
                col sep=comma,
                x=target_goal,
                y=encoding_regression
            ]
            {data/single_task_inference_result/hopper_encoding_regression_line_downsampled.csv};
            \addlegendentry{Hopper}

        \end{axis}

        \node[] at ([xshift=-0.75cm,yshift=0.25cm]encodingplot.north west) {\normalsize\textbf{b}};

    \end{tikzpicture}
    }
    \vskip -.05in
    \caption{
    Visualization of inference results trained on the goal-forward task.
    \textbf{a}, Goal-tracking trajectories across four agents within the first 100 time steps. 
    Solid curves denote the current goals, while dashed curves denote the corresponding target goals.
    \textbf{b}, Encoding-target relationship after convergence across four agents. 
    Scatter points denote inferred latent encodings obtained from trajectory samples from the 50th time step onward, while solid lines show the linear regression trends.
    }
    \label{fig:ant_single_task_inference_visualization}
\vskip -.15in
\end{figure}
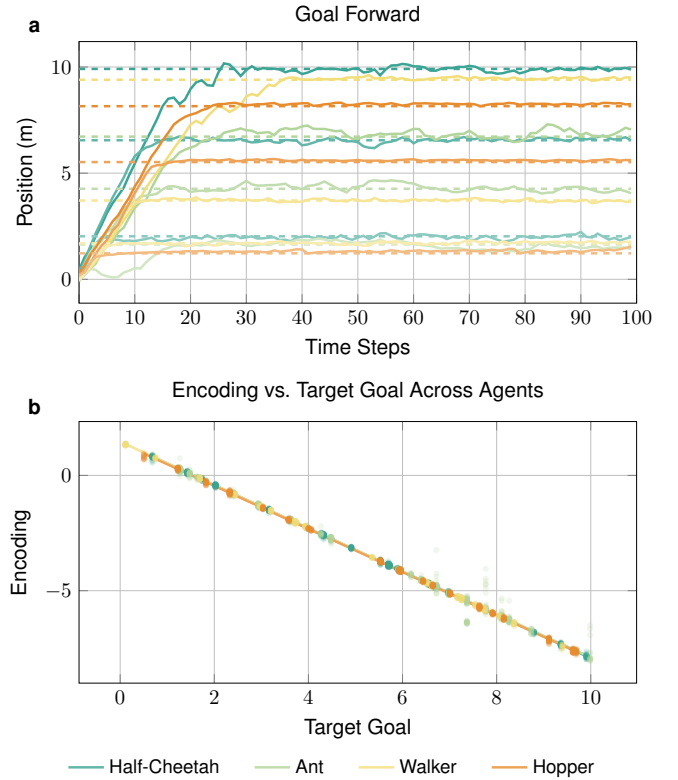

%% file: img_tex/comparison_4agents.tex
\begin{figure*}[t!]
    \centering
    \resizebox{.98\textwidth}{!}{
    \begin{tikzpicture}[font=\sffamily]
        \pgfplotsset{
            ablationaxis/.style={
                tick label style={font=\sffamily\scriptsize},
                label style={font=\sffamily\scriptsize},
                title style={font=\sffamily\small},
                grid=major,
                width=5.5cm,
                height=4cm,
                xmin=0,
                xmax=100
            }
        }

        \begin{axis}[
            ablationaxis,
            name=cheetahpos,
            at={(0,0)},
            anchor=north west,
            title={\textbf{Half-Cheetah (Ours)}},
            ymin=-13,
            ymax=13,
            ylabel={Position (m)},
            ylabel style={yshift=-.2cm},
            xticklabels=\empty,
        ]

            \addplot[fflightgreen, line width=1.2pt]
            table[col sep=comma, x expr=\coordindex, y=x_before]
            {data/multi_task_inference_result_dpmm/dpmm_cheetah/smooth/subgoals_ep0_smooth.csv};
            \addplot[fflightgreen, dashed, line width=1.2pt, forget plot]
            coordinates {(0,-9.02) (100,-9.02)};

            \addplot[fflightgreen!80, line width=1.2pt]
            table[col sep=comma, x expr=\coordindex, y=x_before]
            {data/multi_task_inference_result_dpmm/dpmm_cheetah/smooth/subgoals_ep1_smooth.csv};
            \addplot[fflightgreen!80, dashed, line width=1.2pt, forget plot]
            coordinates {(0,-5.10) (100,-5.10)};

            \addplot[fflightgreen!60, line width=1.2pt]
            table[col sep=comma, x expr=\coordindex, y=x_before]
            {data/multi_task_inference_result_dpmm/dpmm_cheetah/smooth/subgoals_ep2_smooth.csv};
            \addplot[fflightgreen!60, dashed, line width=1.2pt, forget plot]
            coordinates {(0,-3.14) (100,-3.14)};

            \addplot[ffgreen_pv!60, line width=1.2pt]
            table[col sep=comma, x expr=\coordindex, y=x_before]
            {data/multi_task_inference_result_dpmm/dpmm_cheetah/smooth/subgoals_ep3_smooth.csv};
            \addplot[ffgreen_pv!60, dashed, line width=1.2pt, forget plot]
            coordinates {(0,3.14) (100,3.14)};

            \addplot[ffgreen_pv!80, line width=1.2pt]
            table[col sep=comma, x expr=\coordindex, y=x_before]
            {data/multi_task_inference_result_dpmm/dpmm_cheetah/smooth/subgoals_ep4_smooth.csv};
            \addplot[ffgreen_pv!80, dashed, line width=1.2pt, forget plot]
            coordinates {(0,5.10) (100,5.10)};

            \addplot[ffgreen_pv, line width=1.2pt]
            table[col sep=comma, x expr=\coordindex, y=x_before]
            {data/multi_task_inference_result_dpmm/dpmm_cheetah/smooth/subgoals_ep5_smooth.csv};
            \addplot[ffgreen_pv, dashed, line width=1.2pt, forget plot]
            coordinates {(0,9.02) (100,9.02)};
        \end{axis}

        \begin{axis}[
            ablationaxis,
            name=cheetahvel,
            at={(cheetahpos.south west)},
            anchor=north west,
            yshift=-0.35cm,
            ymin=-3,
            ymax=3,
            xlabel={Time Step},
            ylabel={Velocity (m/s)},
        ]

            \addplot[fforange_pv, line width=1.2pt]
            table[col sep=comma, x expr=\coordindex, y=vx_before]
            {data/multi_task_inference_result_dpmm/dpmm_cheetah/smooth/subgoals_ep6_smooth.csv};
            \addplot[fforange_pv, dashed, line width=1.2pt, forget plot]
            coordinates {(0,-2.35) (100,-2.35)};

            \addplot[fforange_pv!80, line width=1.2pt]
            table[col sep=comma, x expr=\coordindex, y=vx_before]
            {data/multi_task_inference_result_dpmm/dpmm_cheetah/smooth/subgoals_ep7_smooth.csv};
            \addplot[fforange_pv!80, dashed, line width=1.2pt, forget plot]
            coordinates {(0,-1.75) (100,-1.75)};

            \addplot[fforange_pv!60, line width=1.2pt]
            table[col sep=comma, x expr=\coordindex, y=vx_before]
            {data/multi_task_inference_result_dpmm/dpmm_cheetah/smooth/subgoals_ep8_smooth.csv};
            \addplot[fforange_pv!60, dashed, line width=1.2pt, forget plot]
            coordinates {(0,-1.45) (100,-1.45)};

            \addplot[ffyellow!60, line width=1.2pt]
            table[col sep=comma, x expr=\coordindex, y=vx_before]
            {data/multi_task_inference_result_dpmm/dpmm_cheetah/smooth/subgoals_ep9_smooth.csv};
            \addplot[ffyellow!60, dashed, line width=1.2pt, forget plot]
            coordinates {(0,1.45) (100,1.45)};

            \addplot[ffyellow!80, line width=1.2pt]
            table[col sep=comma, x expr=\coordindex, y=vx_before]
            {data/multi_task_inference_result_dpmm/dpmm_cheetah/smooth/subgoals_ep10_smooth.csv};
            \addplot[ffyellow!80, dashed, line width=1.2pt, forget plot]
            coordinates {(0,1.75) (100,1.75)};

            \addplot[ffyellow, line width=1.2pt]
            table[col sep=comma, x expr=\coordindex, y=vx_before]
            {data/multi_task_inference_result_dpmm/dpmm_cheetah/smooth/subgoals_ep11_smooth.csv};
            \addplot[ffyellow, dashed, line width=1.2pt, forget plot]
            coordinates {(0,2.35) (100,2.35)};
        \end{axis}

        \begin{axis}[
            ablationaxis,
            name=antpos,
            at={(5.25cm,0)},
            anchor=north west,
            title={\textbf{Ant (Ours)}},
            ymin=-13,
            ymax=13,
            ylabel={Position (m)},
            ylabel style={yshift=-.2cm},
            xticklabels=\empty,
        ]

            \addplot[fflightgreen, line width=1.2pt]
            table[col sep=comma, x expr=\coordindex, y=x_before]
            {data/multi_task_inference_result_dpmm/dpmm_ant/smooth/subgoals_ep0_smooth.csv};
            \addplot[fflightgreen, dashed, line width=1.2pt, forget plot]
            coordinates {(0,-9.02) (100,-9.02)};

            \addplot[fflightgreen!80, line width=1.2pt]
            table[col sep=comma, x expr=\coordindex, y=x_before]
            {data/multi_task_inference_result_dpmm/dpmm_ant/smooth/subgoals_ep1_smooth.csv};
            \addplot[fflightgreen!80, dashed, line width=1.2pt, forget plot]
            coordinates {(0,-5.10) (100,-5.10)};

            \addplot[fflightgreen!60, line width=1.2pt]
            table[col sep=comma, x expr=\coordindex, y=x_before]
            {data/multi_task_inference_result_dpmm/dpmm_ant/smooth/subgoals_ep2_smooth.csv};
            \addplot[fflightgreen!60, dashed, line width=1.2pt, forget plot]
            coordinates {(0,-3.14) (100,-3.14)};

            \addplot[ffgreen_pv!60, line width=1.2pt]
            table[col sep=comma, x expr=\coordindex, y=x_before]
            {data/multi_task_inference_result_dpmm/dpmm_ant/smooth/subgoals_ep3_smooth.csv};
            \addplot[ffgreen_pv!60, dashed, line width=1.2pt, forget plot]
            coordinates {(0,3.14) (100,3.14)};

            \addplot[ffgreen_pv!80, line width=1.2pt]
            table[col sep=comma, x expr=\coordindex, y=x_before]
            {data/multi_task_inference_result_dpmm/dpmm_ant/smooth/subgoals_ep4_smooth.csv};
            \addplot[ffgreen_pv!80, dashed, line width=1.2pt, forget plot]
            coordinates {(0,5.10) (100,5.10)};

            \addplot[ffgreen_pv, line width=1.2pt]
            table[col sep=comma, x expr=\coordindex, y=x_before]
            {data/multi_task_inference_result_dpmm/dpmm_ant/smooth/subgoals_ep5_smooth.csv};
            \addplot[ffgreen_pv, dashed, line width=1.2pt, forget plot]
            coordinates {(0,9.02) (100,9.02)};
        \end{axis}

        \begin{axis}[
            ablationaxis,
            name=antvel,
            at={(antpos.south west)},
            anchor=north west,
            yshift=-0.35cm,
            ymin=-3,
            ymax=3,
            xlabel={Time Step},
            ylabel={Velocity (m/s)},
        ]

            \addplot[fforange_pv, line width=1.2pt]
            table[col sep=comma, x expr=\coordindex, y=vx_before]
            {data/multi_task_inference_result_dpmm/dpmm_ant/smooth/subgoals_ep6_smooth.csv};
            \addplot[fforange_pv, dashed, line width=1.2pt, forget plot]
            coordinates {(0,-2.35) (100,-2.35)};

            \addplot[fforange_pv!80, line width=1.2pt]
            table[col sep=comma, x expr=\coordindex, y=vx_before]
            {data/multi_task_inference_result_dpmm/dpmm_ant/smooth/subgoals_ep7_smooth.csv};
            \addplot[fforange_pv!80, dashed, line width=1.2pt, forget plot]
            coordinates {(0,-1.75) (100,-1.75)};

            \addplot[fforange_pv!60, line width=1.2pt]
            table[col sep=comma, x expr=\coordindex, y=vx_before]
            {data/multi_task_inference_result_dpmm/dpmm_ant/smooth/subgoals_ep8_smooth.csv};
            \addplot[fforange_pv!60, dashed, line width=1.2pt, forget plot]
            coordinates {(0,-1.45) (100,-1.45)};

            \addplot[ffyellow!60, line width=1.2pt]
            table[col sep=comma, x expr=\coordindex, y=vx_before]
            {data/multi_task_inference_result_dpmm/dpmm_ant/smooth/subgoals_ep9_smooth.csv};
            \addplot[ffyellow!60, dashed, line width=1.2pt, forget plot]
            coordinates {(0,1.45) (100,1.45)};

            \addplot[ffyellow!80, line width=1.2pt]
            table[col sep=comma, x expr=\coordindex, y=vx_before]
            {data/multi_task_inference_result_dpmm/dpmm_ant/smooth/subgoals_ep10_smooth.csv};
            \addplot[ffyellow!80, dashed, line width=1.2pt, forget plot]
            coordinates {(0,1.75) (100,1.75)};

            \addplot[ffyellow, line width=1.2pt]
            table[col sep=comma, x expr=\coordindex, y=vx_before]
            {data/multi_task_inference_result_dpmm/dpmm_ant/smooth/subgoals_ep11_smooth.csv};
            \addplot[ffyellow, dashed, line width=1.2pt, forget plot]
            coordinates {(0,2.35) (100,2.35)};
        \end{axis}

        \begin{axis}[
            ablationaxis,
            name=walkerpos,
            at={(10.5cm,0)},
            anchor=north west,
            title={\textbf{Walker (Ours)}},
            ymin=-13,
            ymax=13,
            ylabel={Position (m)},
            ylabel style={yshift=-.2cm},
            xticklabels=\empty,
        ]

            \addplot[fflightgreen, line width=1.2pt]
            table[col sep=comma, x expr=\coordindex, y=x_before]
            {data/multi_task_inference_result_dpmm/dpmm_walker/smooth/subgoals_ep0_smooth.csv};
            \addplot[fflightgreen, dashed, line width=1.2pt, forget plot]
            coordinates {(0,-9.02) (100,-9.02)};

            \addplot[fflightgreen!80, line width=1.2pt]
            table[col sep=comma, x expr=\coordindex, y=x_before]
            {data/multi_task_inference_result_dpmm/dpmm_walker/smooth/subgoals_ep1_smooth.csv};
            \addplot[fflightgreen!80, dashed, line width=1.2pt, forget plot]
            coordinates {(0,-5.10) (100,-5.10)};

            \addplot[fflightgreen!60, line width=1.2pt]
            table[col sep=comma, x expr=\coordindex, y=x_before]
            {data/multi_task_inference_result_dpmm/dpmm_walker/smooth/subgoals_ep2_smooth.csv};
            \addplot[fflightgreen!60, dashed, line width=1.2pt, forget plot]
            coordinates {(0,-3.14) (100,-3.14)};

            \addplot[ffgreen_pv!60, line width=1.2pt]
            table[col sep=comma, x expr=\coordindex, y=x_before]
            {data/multi_task_inference_result_dpmm/dpmm_walker/smooth/subgoals_ep3_smooth.csv};
            \addplot[ffgreen_pv!60, dashed, line width=1.2pt, forget plot]
            coordinates {(0,3.14) (100,3.14)};

            \addplot[ffgreen_pv!80, line width=1.2pt]
            table[col sep=comma, x expr=\coordindex, y=x_before]
            {data/multi_task_inference_result_dpmm/dpmm_walker/smooth/subgoals_ep4_smooth.csv};
            \addplot[ffgreen_pv!80, dashed, line width=1.2pt, forget plot]
            coordinates {(0,5.10) (100,5.10)};

            \addplot[ffgreen_pv, line width=1.2pt]
            table[col sep=comma, x expr=\coordindex, y=x_before]
            {data/multi_task_inference_result_dpmm/dpmm_walker/smooth/subgoals_ep5_smooth.csv};
            \addplot[ffgreen_pv, dashed, line width=1.2pt, forget plot]
            coordinates {(0,9.02) (100,9.02)};
        \end{axis}

        \begin{axis}[
            ablationaxis,
            name=walkervel,
            at={(walkerpos.south west)},
            anchor=north west,
            yshift=-0.35cm,
            ymin=-3,
            ymax=3,
            xlabel={Time Step},
            ylabel={Velocity (m/s)},
        ]

            \addplot[fforange_pv, line width=1.2pt]
            table[col sep=comma, x expr=\coordindex, y=vx_before]
            {data/multi_task_inference_result_dpmm/dpmm_walker/smooth/subgoals_ep6_smooth.csv};
            \addplot[fforange_pv, dashed, line width=1.2pt, forget plot]
            coordinates {(0,-2.35) (100,-2.35)};

            \addplot[fforange_pv!80, line width=1.2pt]
            table[col sep=comma, x expr=\coordindex, y=vx_before]
            {data/multi_task_inference_result_dpmm/dpmm_walker/smooth/subgoals_ep7_smooth.csv};
            \addplot[fforange_pv!80, dashed, line width=1.2pt, forget plot]
            coordinates {(0,-1.75) (100,-1.75)};

            \addplot[fforange_pv!60, line width=1.2pt]
            table[col sep=comma, x expr=\coordindex, y=vx_before]
            {data/multi_task_inference_result_dpmm/dpmm_walker/smooth/subgoals_ep8_smooth.csv};
            \addplot[fforange_pv!60, dashed, line width=1.2pt, forget plot]
            coordinates {(0,-1.45) (100,-1.45)};

            \addplot[ffyellow!60, line width=1.2pt]
            table[col sep=comma, x expr=\coordindex, y=vx_before]
            {data/multi_task_inference_result_dpmm/dpmm_walker/smooth/subgoals_ep9_smooth.csv};
            \addplot[ffyellow!60, dashed, line width=1.2pt, forget plot]
            coordinates {(0,1.45) (100,1.45)};

            \addplot[ffyellow!80, line width=1.2pt]
            table[col sep=comma, x expr=\coordindex, y=vx_before]
            {data/multi_task_inference_result_dpmm/dpmm_walker/smooth/subgoals_ep10_smooth.csv};
            \addplot[ffyellow!80, dashed, line width=1.2pt, forget plot]
            coordinates {(0,1.75) (100,1.75)};

            \addplot[ffyellow, line width=1.2pt]
            table[col sep=comma, x expr=\coordindex, y=vx_before]
            {data/multi_task_inference_result_dpmm/dpmm_walker/smooth/subgoals_ep11_smooth.csv};
            \addplot[ffyellow, dashed, line width=1.2pt, forget plot]
            coordinates {(0,2.35) (100,2.35)};
        \end{axis}

        \begin{axis}[
            ablationaxis,
            name=hopperpos,
            at={(15.75cm,0)},
            anchor=north west,
            title={\textbf{Hopper (Ours)}},
            ymin=-13,
            ymax=13,
            ylabel={Position (m)},
            ylabel style={yshift=-.2cm},
            xticklabels=\empty,
        ]

            \addplot[fflightgreen, line width=1.2pt]
            table[col sep=comma, x expr=\coordindex, y=x_before]
            {data/multi_task_inference_result_dpmm/dpmm_hopper/smooth/subgoals_ep0_smooth.csv};
            \addplot[fflightgreen, dashed, line width=1.2pt, forget plot]
            coordinates {(0,-9.02) (100,-9.02)};

            \addplot[fflightgreen!80, line width=1.2pt]
            table[col sep=comma, x expr=\coordindex, y=x_before]
            {data/multi_task_inference_result_dpmm/dpmm_hopper/smooth/subgoals_ep1_smooth.csv};
            \addplot[fflightgreen!80, dashed, line width=1.2pt, forget plot]
            coordinates {(0,-5.10) (100,-5.10)};

            \addplot[fflightgreen!60, line width=1.2pt]
            table[col sep=comma, x expr=\coordindex, y=x_before]
            {data/multi_task_inference_result_dpmm/dpmm_hopper/smooth/subgoals_ep2_smooth.csv};
            \addplot[fflightgreen!60, dashed, line width=1.2pt, forget plot]
            coordinates {(0,-3.14) (100,-3.14)};

            \addplot[ffgreen_pv!60, line width=1.2pt]
            table[col sep=comma, x expr=\coordindex, y=x_before]
            {data/multi_task_inference_result_dpmm/dpmm_hopper/smooth/subgoals_ep3_smooth.csv};
            \addplot[ffgreen_pv!60, dashed, line width=1.2pt, forget plot]
            coordinates {(0,3.14) (100,3.14)};

            \addplot[ffgreen_pv!80, line width=1.2pt]
            table[col sep=comma, x expr=\coordindex, y=x_before]
            {data/multi_task_inference_result_dpmm/dpmm_hopper/smooth/subgoals_ep4_smooth.csv};
            \addplot[ffgreen_pv!80, dashed, line width=1.2pt, forget plot]
            coordinates {(0,5.10) (100,5.10)};

            \addplot[ffgreen_pv, line width=1.2pt]
            table[col sep=comma, x expr=\coordindex, y=x_before]
            {data/multi_task_inference_result_dpmm/dpmm_hopper/smooth/subgoals_ep5_smooth.csv};
            \addplot[ffgreen_pv, dashed, line width=1.2pt, forget plot]
            coordinates {(0,9.02) (100,9.02)};
        \end{axis}

        \begin{axis}[
            ablationaxis,
            name=hoppervel,
            at={(hopperpos.south west)},
            anchor=north west,
            yshift=-0.35cm,
            ymin=-3,
            ymax=3,
            xlabel={Time Step},
            ylabel={Velocity (m/s)},
        ]

            \addplot[fforange_pv, line width=1.2pt]
            table[col sep=comma, x expr=\coordindex, y=vx_before]
            {data/multi_task_inference_result_dpmm/dpmm_hopper/smooth/subgoals_ep6_smooth.csv};
            \addplot[fforange_pv, dashed, line width=1.2pt, forget plot]
            coordinates {(0,-2.35) (100,-2.35)};

            \addplot[fforange_pv!80, line width=1.2pt]
            table[col sep=comma, x expr=\coordindex, y=vx_before]
            {data/multi_task_inference_result_dpmm/dpmm_hopper/smooth/subgoals_ep7_smooth.csv};
            \addplot[fforange_pv!80, dashed, line width=1.2pt, forget plot]
            coordinates {(0,-1.75) (100,-1.75)};

            \addplot[fforange_pv!60, line width=1.2pt]
            table[col sep=comma, x expr=\coordindex, y=vx_before]
            {data/multi_task_inference_result_dpmm/dpmm_hopper/smooth/subgoals_ep8_smooth.csv};
            \addplot[fforange_pv!60, dashed, line width=1.2pt, forget plot]
            coordinates {(0,-1.45) (100,-1.45)};

            \addplot[ffyellow!60, line width=1.2pt]
            table[col sep=comma, x expr=\coordindex, y=vx_before]
            {data/multi_task_inference_result_dpmm/dpmm_hopper/smooth/subgoals_ep9_smooth.csv};
            \addplot[ffyellow!60, dashed, line width=1.2pt, forget plot]
            coordinates {(0,1.45) (100,1.45)};

            \addplot[ffyellow!80, line width=1.2pt]
            table[col sep=comma, x expr=\coordindex, y=vx_before]
            {data/multi_task_inference_result_dpmm/dpmm_hopper/smooth/subgoals_ep10_smooth.csv};
            \addplot[ffyellow!80, dashed, line width=1.2pt, forget plot]
            coordinates {(0,1.75) (100,1.75)};

            \addplot[ffyellow, line width=1.2pt]
            table[col sep=comma, x expr=\coordindex, y=vx_before]
            {data/multi_task_inference_result_dpmm/dpmm_hopper/smooth/subgoals_ep11_smooth.csv};
            \addplot[ffyellow, dashed, line width=1.2pt, forget plot]
            coordinates {(0,2.35) (100,2.35)};
        \end{axis}

        \node[inner sep=-1pt, below] (cheetah-tsne)
        at ([yshift=-1cm]cheetahvel.south)
        {\includegraphics[width=5cm]{imgs/dpmm-cheetah-tsne.png}};

        \node[inner sep=-1pt, below] (ant-tsne)
        at ([yshift=-1cm]antvel.south)
        {\includegraphics[width=5cm]{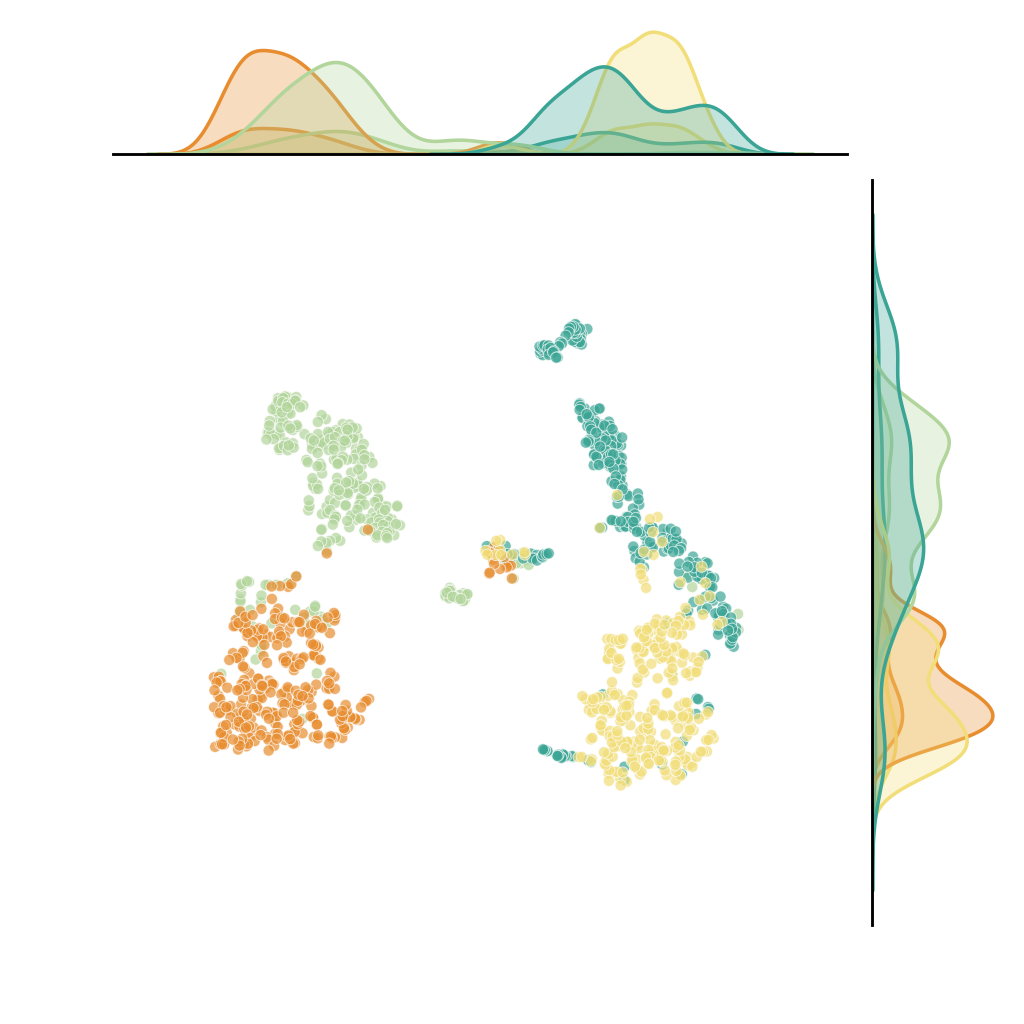}};

        \node[inner sep=-1pt, below] (walker-tsne)
        at ([yshift=-1cm]walkervel.south)
        {\includegraphics[width=5cm]{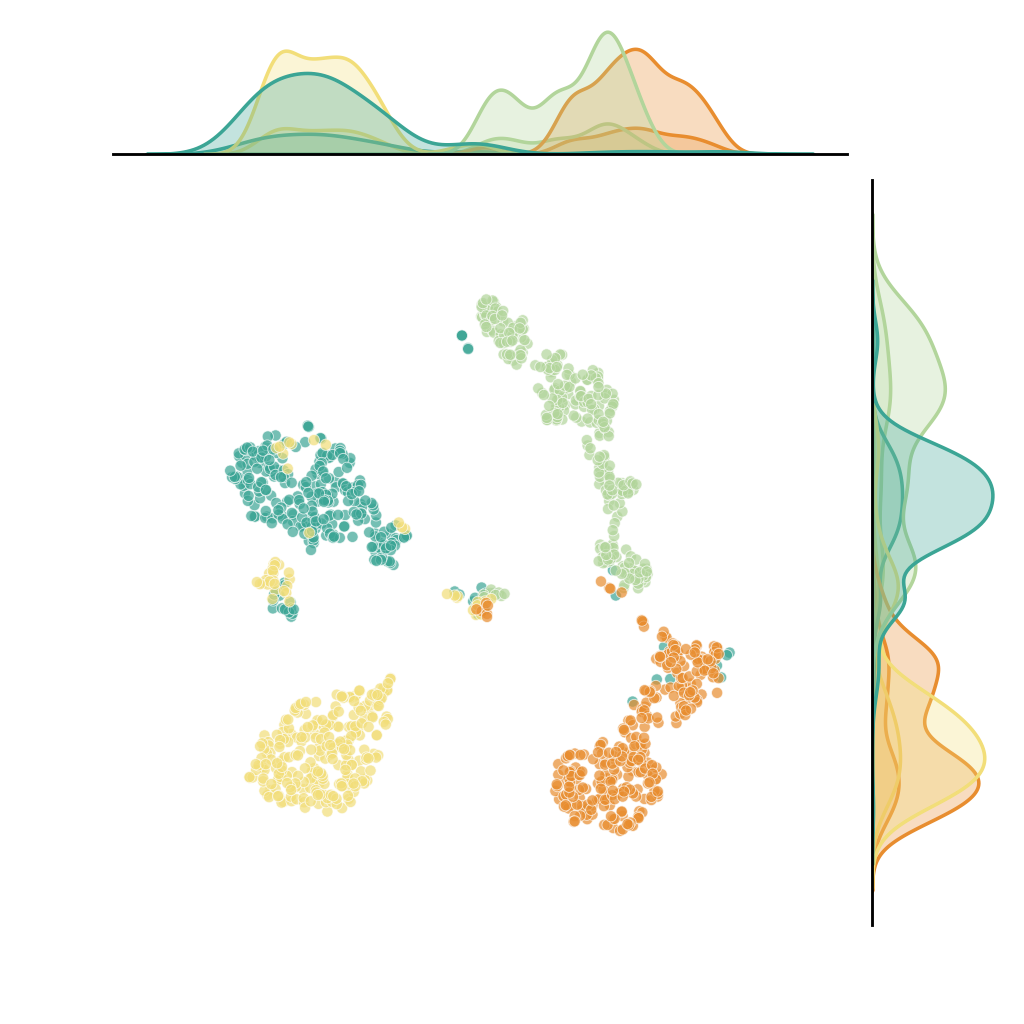}};

        \node[inner sep=-1pt, below] (hopper-tsne)
        at ([yshift=-1cm]hoppervel.south)
        {\includegraphics[width=5cm]{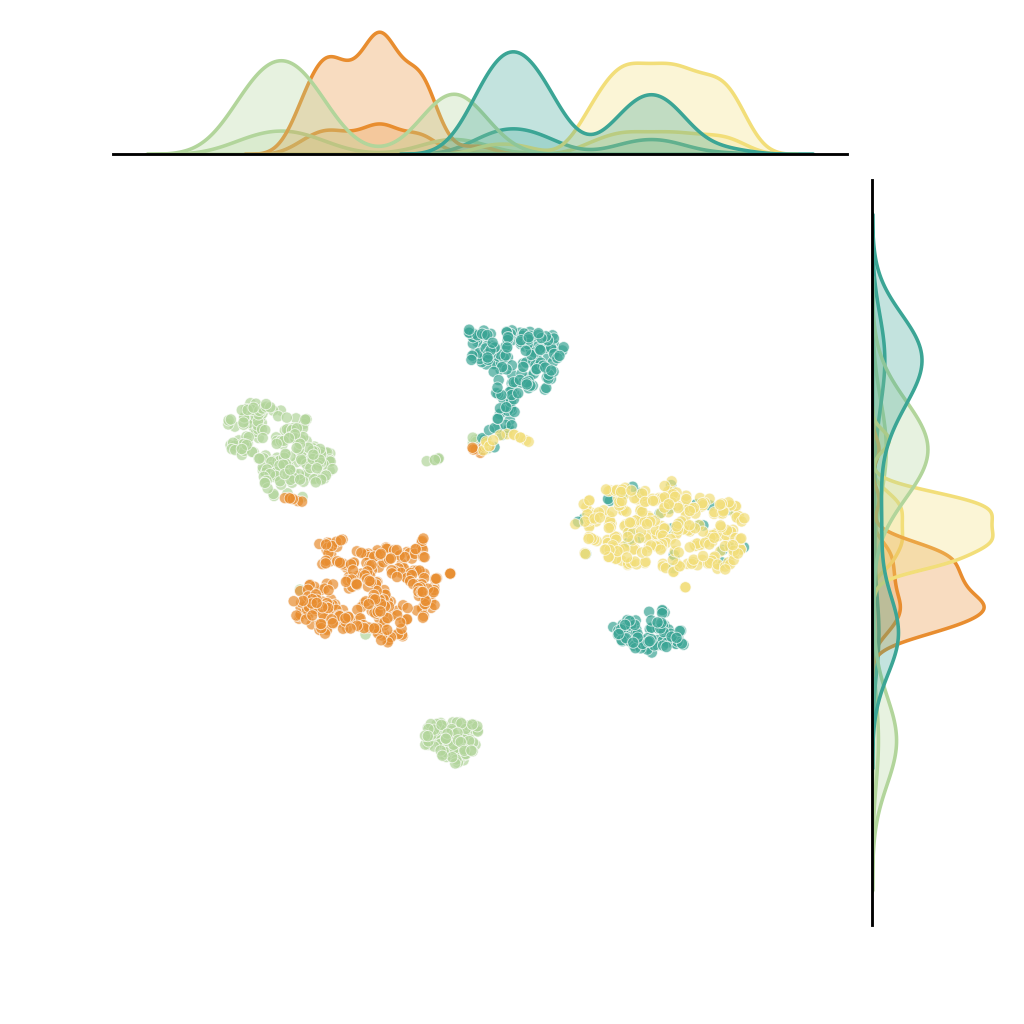}};

        \node[inner sep=-1pt, below right] (cheetah-cm)
        at ([yshift=-.75cm]cheetah-tsne.south west)
        {\includegraphics[width=4.5cm]{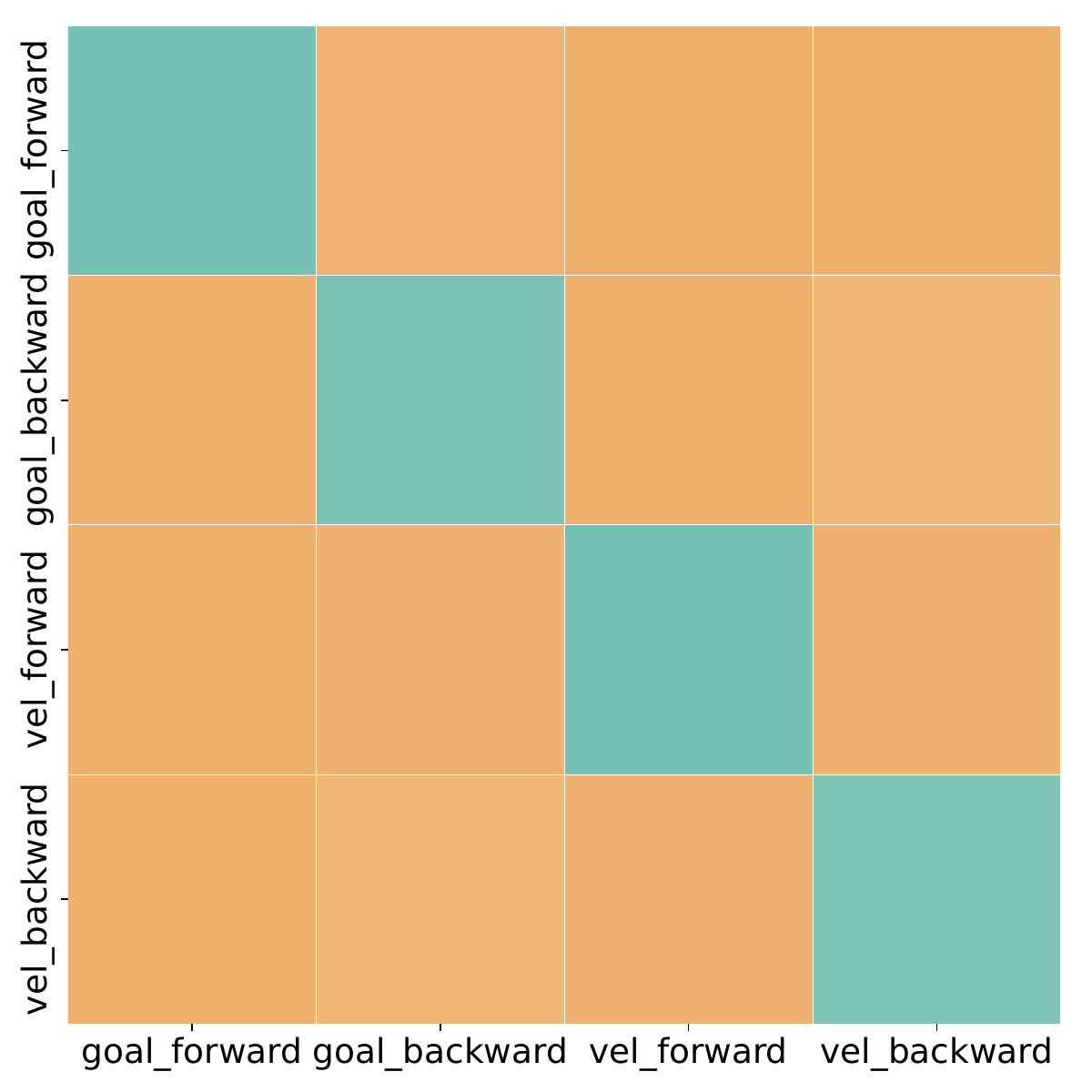}};

        \node[inner sep=-1pt, below right] (ant-cm)
        at ([yshift=-.75cm]ant-tsne.south west)
        {\includegraphics[width=4.5cm]{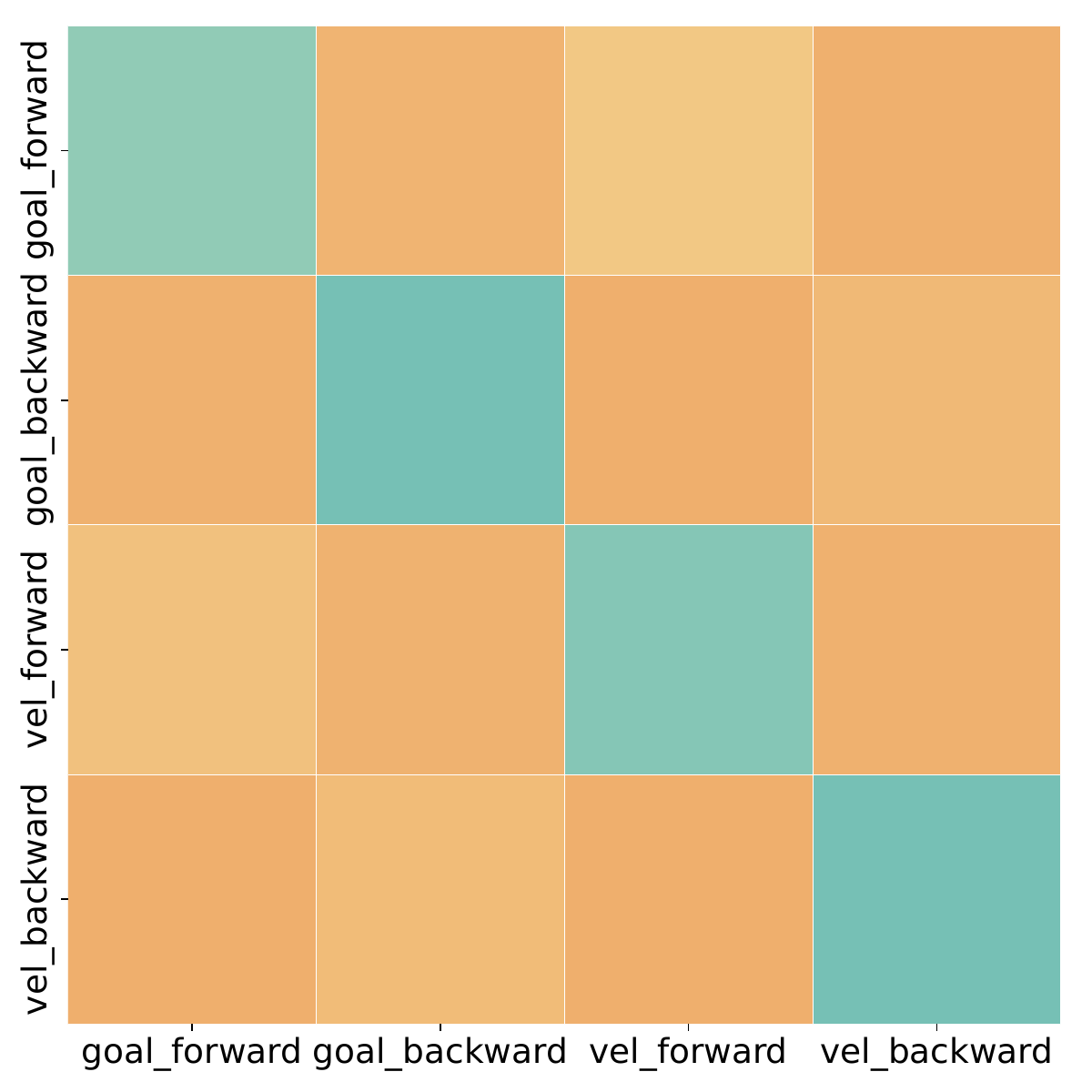}};

        \node[inner sep=-1pt, below right] (walker-cm)
        at ([yshift=-.75cm]walker-tsne.south west)
        {\includegraphics[width=4.5cm]{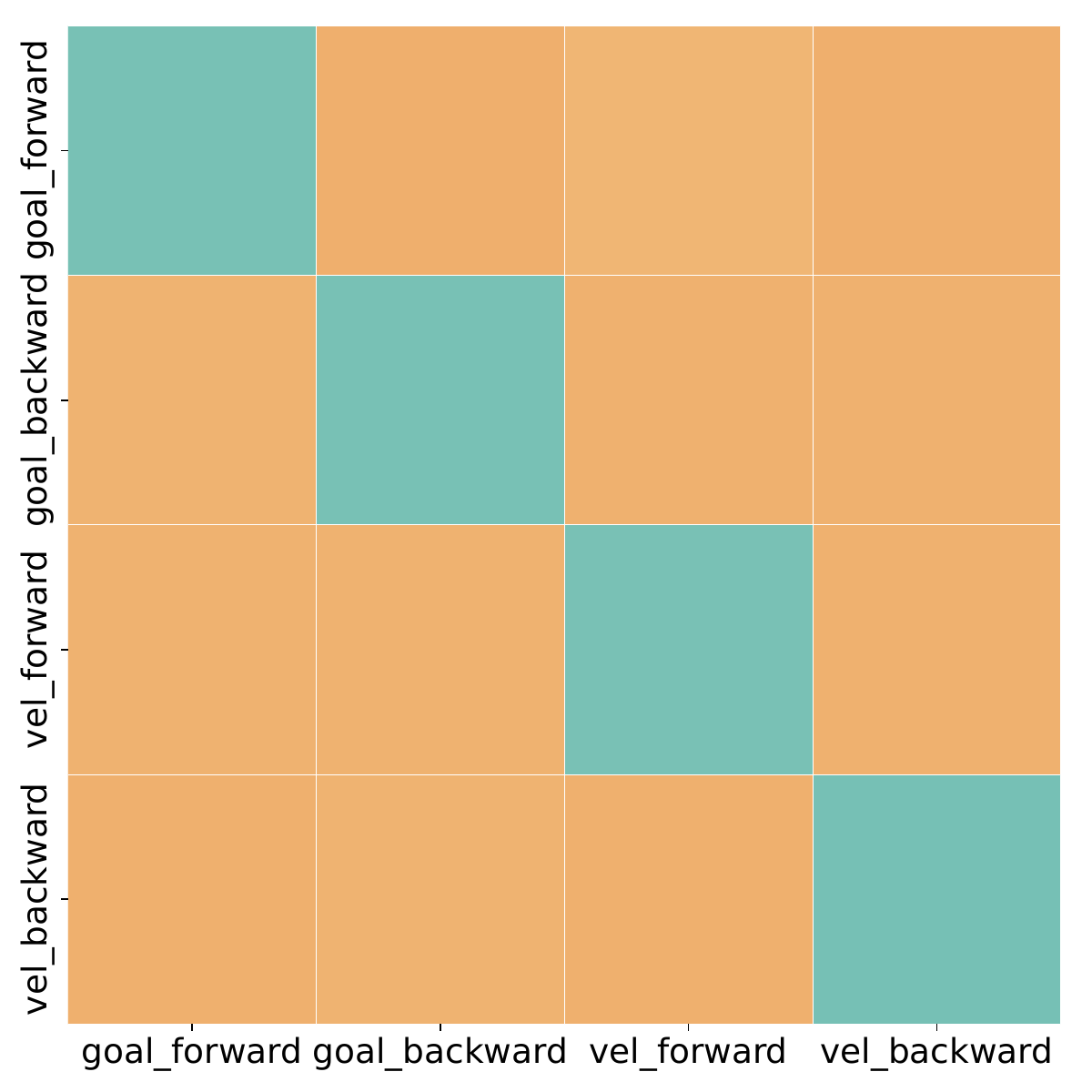}};

        \node[inner sep=-1pt, below right] (hopper-cm)
        at ([yshift=-.75cm]hopper-tsne.south west)
        {\includegraphics[width=4.5cm]{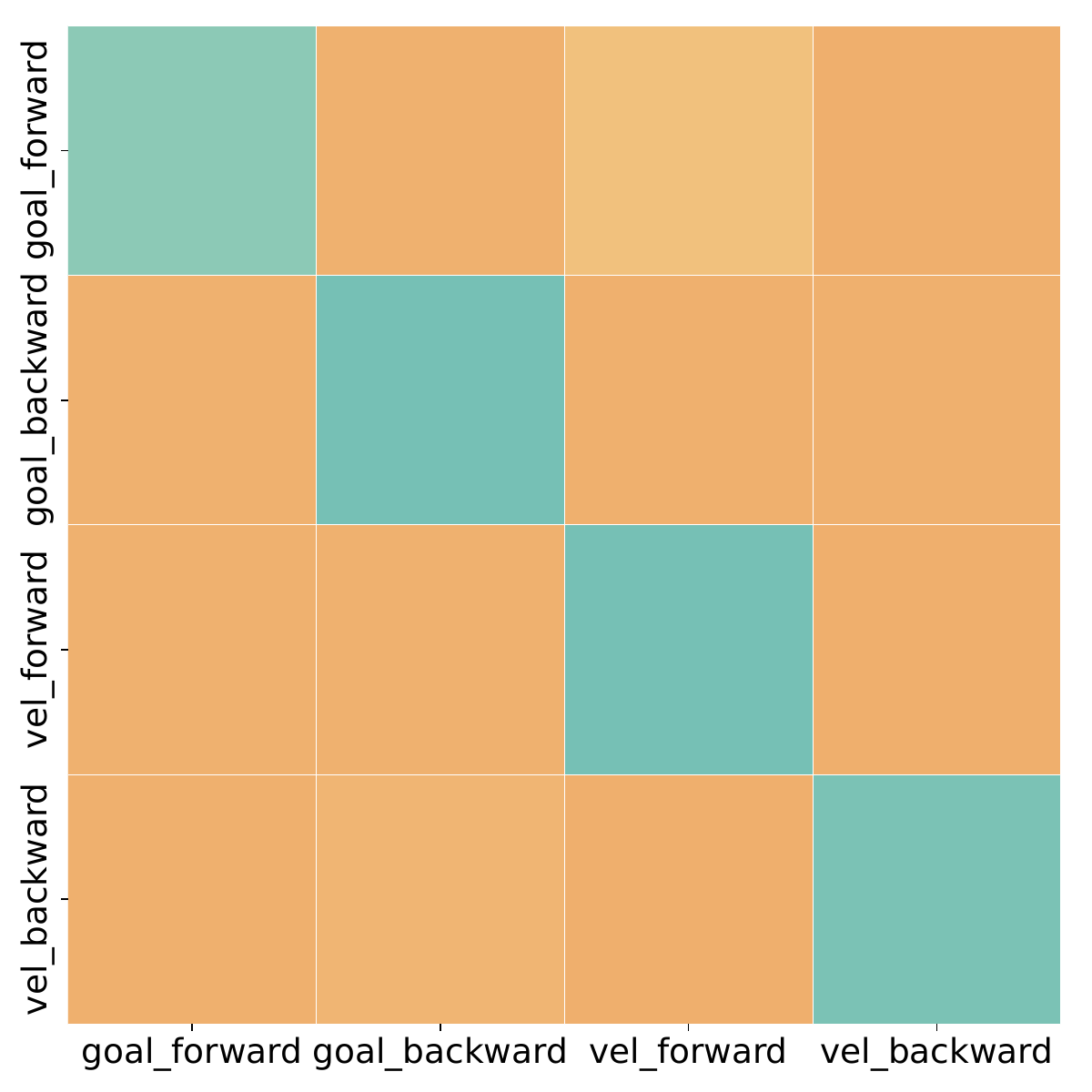}};

        \pgfdeclareverticalshading{accuracybarshade}{100bp}
        {
          color(0bp)=(fforange_pv!80);
          color(30bp)=(ffyellow!80);
          color(60bp)=(fflightgreen!80);
          color(100bp)=(ffgreen_pv!80)
        }
        
        \draw[
          draw=none,
          path picture={
            \fill[shading=accuracybarshade]
              (path picture bounding box.south west)
              rectangle
              (path picture bounding box.north east);
          }
        ]
        ([xshift=.05cm, yshift=-.1cm]cheetah-cm.north east)
        rectangle
        ([xshift=.35cm, yshift=.1cm]cheetah-cm.south east);
        \node[above] at ([xshift=.2cm, yshift=-.25cm]cheetah-cm.south east) {\scriptsize low};
        \node[below] at ([xshift=.2cm, yshift=.25cm]cheetah-cm.north east) {\scriptsize high};
        
        \draw[
          draw=none,
          path picture={
            \fill[shading=accuracybarshade]
              (path picture bounding box.south west)
              rectangle
              (path picture bounding box.north east);
          }
        ]
        ([xshift=.05cm, yshift=-.1cm]ant-cm.north east)
        rectangle
        ([xshift=.35cm, yshift=.1cm]ant-cm.south east);
        \node[above] at ([xshift=.2cm, yshift=-.25cm]ant-cm.south east) {\scriptsize low};
        \node[below] at ([xshift=.2cm, yshift=.25cm]ant-cm.north east) {\scriptsize high};
        
        \draw[
          draw=none,
          path picture={
            \fill[shading=accuracybarshade]
              (path picture bounding box.south west)
              rectangle
              (path picture bounding box.north east);
          }
        ]
        ([xshift=.05cm, yshift=-.1cm]walker-cm.north east)
        rectangle
        ([xshift=.35cm, yshift=.1cm]walker-cm.south east);
        \node[above] at ([xshift=.2cm, yshift=-.25cm]walker-cm.south east) {\scriptsize low};
        \node[below] at ([xshift=.2cm, yshift=.25cm]walker-cm.north east) {\scriptsize high};
        
        \draw[
          draw=none,
          path picture={
            \fill[shading=accuracybarshade]
              (path picture bounding box.south west)
              rectangle
              (path picture bounding box.north east);
          }
        ]
        ([xshift=.05cm, yshift=-.1cm]hopper-cm.north east)
        rectangle
        ([xshift=.35cm, yshift=.1cm]hopper-cm.south east);
        \node[above] at ([xshift=.2cm, yshift=-.25cm]hopper-cm.south east) {\scriptsize low};
        \node[below] at ([xshift=.2cm, yshift=.25cm]hopper-cm.north east) {\scriptsize high};

        \node[below] at (cheetah-cm.south) {\scriptsize Prediction};
        \node[above, rotate=90] at (cheetah-cm.west) {\scriptsize Ground Truth};

        \node[below] at (ant-cm.south) {\scriptsize Prediction};
        \node[above, rotate=90] at (ant-cm.west) {\scriptsize Ground Truth};

        \node[below] at (walker-cm.south) {\scriptsize Prediction};
        \node[above, rotate=90] at (walker-cm.west) {\scriptsize Ground Truth};

        \node[below] at (hopper-cm.south) {\scriptsize Prediction};
        \node[above, rotate=90] at (hopper-cm.west) {\scriptsize Ground Truth};

        \node[fill=ffgreen_pv!80, above left,
        label={[label distance=0pt]0:\textsf{goal forward}}]
        (l1) at ([xshift=-1cm]cheetah-tsne.south) {};

        \node[fill=fflightgreen!80, above left,
        label={[label distance=0pt]0:\textsf{goal backward}}]
        (l2) at ([xshift=-1cm]ant-tsne.south) {};

        \node[fill=ffyellow!80, above left,
        label={[label distance=0pt]0:\textsf{velocity forward}}]
        (l3) at ([xshift=-1cm]walker-tsne.south) {};

        \node[fill=fforange_pv!80, above left,
        label={[label distance=0pt]0:\textsf{velocity backward}}]
        (l4) at ([xshift=-1cm]hopper-tsne.south) {};

        \node[below, font=\bfseries\sffamily\small]at(cheetah-cm.north){Half-Cheetah (Ours)};
        \node[below, font=\bfseries\sffamily\small]at(ant-cm.north){Ant (Ours)};
        \node[below, font=\bfseries\sffamily\small]at(walker-cm.north){Walker (Ours)};
        \node[below, font=\bfseries\sffamily\small]at(hopper-cm.north){Hopper (Ours)};
        \node[above left] at ([xshift=-.5cm, yshift=.25cm]cheetahpos.north west) {\textbf{a}};
        \node[above left] at ([xshift=-.5cm, yshift=.25cm]antpos.north west) {\textbf{b}};
        \node[above left] at ([xshift=-.5cm, yshift=.25cm]walkerpos.north west) {\textbf{c}};
        \node[above left] at ([xshift=-.5cm, yshift=.25cm]hopperpos.north west) {\textbf{d}};

        \node[below right] at ([xshift=-.5cm, yshift=.25cm]cheetah-tsne.north west) {\textbf{e}};
        \node[below right] at ([xshift=-.5cm, yshift=.25cm]ant-tsne.north west) {\textbf{f}};
        \node[below right] at ([xshift=-.5cm, yshift=.25cm]walker-tsne.north west) {\textbf{g}};
        \node[below right] at ([xshift=-.5cm, yshift=.25cm]hopper-tsne.north west) {\textbf{h}};

        \node[above right] at ([xshift=-.5cm, yshift=.25cm]cheetah-cm.north west) {\textbf{i}};
        \node[above right] at ([xshift=-.5cm, yshift=.25cm]ant-cm.north west) {\textbf{j}};
        \node[above right] at ([xshift=-.5cm, yshift=.25cm]walker-cm.north west) {\textbf{k}};
        \node[above right] at ([xshift=-.5cm, yshift=.25cm]hopper-cm.north west) {\textbf{l}};

    \end{tikzpicture}
    }
    \vskip -.05in
    \caption{
    Cross-agent deployment of our ReMAP DPMM-based task inference module.
    \textbf{a}--\textbf{d}, Inference trajectories on Half-Cheetah, Ant, Walker, and Hopper, respectively. For each agent, the upper plot shows position tracking and the lower plot shows velocity tracking across twelve non-parametric tasks; dashed horizontal lines denote the corresponding ground-truth targets.
    \textbf{e}--\textbf{h}, t-SNE visualizations of the inferred latent representations after deploying the frozen task inference module to different target agents. Across all embodiments, different task types form clearly distinguishable clusters, indicating that the learned latent representation preserves task semantics across agents.
    \textbf{i}--\textbf{l}, Decoder prediction confusion matrices for Half-Cheetah, Ant, Walker, and Hopper. 
    The consistently high prediction accuracy shows that the inferred latent representations remain semantically informative after cross-agent deployment. 
    }
    \label{fig:dpmm_multi_agent_inference_all}
\vskip -.15in
\end{figure*}

%% file: img_tex/all_baselines_task_inferencen_results.tex
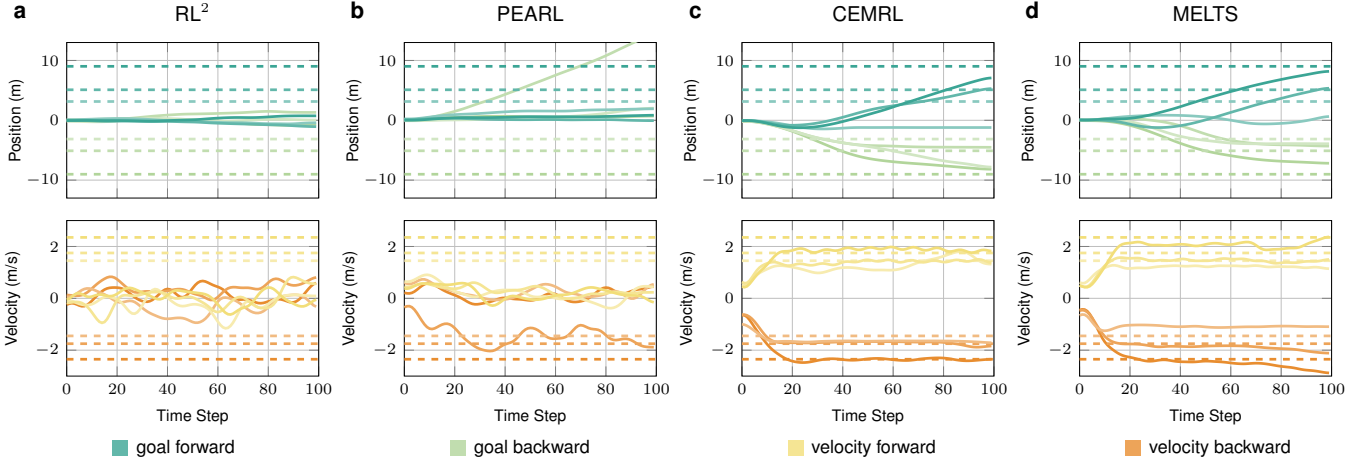
\begin{figure*}[t!]
    \centering
    \resizebox{\textwidth}{!}{
    \begin{tikzpicture}[font=\sffamily]
        \pgfplotsset{
            baselineaxis/.style={
                tick label style={font=\sffamily\scriptsize},
                label style={font=\sffamily\scriptsize},
                title style={font=\sffamily\small},
                grid=major,
                width=5.5cm,
                height=4cm,
                xmin=0,
                xmax=100
            }
        }

        \begin{axis}[
            baselineaxis,
            name=rl2pos,
            at={(0,0)},
            anchor=north west,
            title={RL$^2$},
            ymin=-13,
            ymax=13,
            ylabel={Position (m)},
            ylabel style={yshift=-.2cm},
            xticklabels=\empty,
        ]

            \addplot[fflightgreen, line width=1.2pt]
            table[col sep=comma, x expr=\coordindex, y=x_before]
            {data/baseline/rl2/logs/smooth/subgoals_ep0_smooth.csv};
            \addplot[fflightgreen, dashed, line width=1.2pt, forget plot]
            coordinates {(0,-9.02) (100,-9.02)};

            \addplot[fflightgreen!80, line width=1.2pt]
            table[col sep=comma, x expr=\coordindex, y=x_before]
            {data/baseline/rl2/logs/smooth/subgoals_ep1_smooth.csv};
            \addplot[fflightgreen!80, dashed, line width=1.2pt, forget plot]
            coordinates {(0,-5.10) (100,-5.10)};

            \addplot[fflightgreen!60, line width=1.2pt]
            table[col sep=comma, x expr=\coordindex, y=x_before]
            {data/baseline/rl2/logs/smooth/subgoals_ep2_smooth.csv};
            \addplot[fflightgreen!60, dashed, line width=1.2pt, forget plot]
            coordinates {(0,-3.14) (100,-3.14)};

            \addplot[ffgreen_pv!60, line width=1.2pt]
            table[col sep=comma, x expr=\coordindex, y=x_before]
            {data/baseline/rl2/logs/smooth/subgoals_ep3_smooth.csv};
            \addplot[ffgreen_pv!60, dashed, line width=1.2pt, forget plot]
            coordinates {(0,3.14) (100,3.14)};

            \addplot[ffgreen_pv!80, line width=1.2pt]
            table[col sep=comma, x expr=\coordindex, y=x_before]
            {data/baseline/rl2/logs/smooth/subgoals_ep4_smooth.csv};
            \addplot[ffgreen_pv!80, dashed, line width=1.2pt, forget plot]
            coordinates {(0,5.10) (100,5.10)};

            \addplot[ffgreen_pv, line width=1.2pt]
            table[col sep=comma, x expr=\coordindex, y=x_before]
            {data/baseline/rl2/logs/smooth/subgoals_ep5_smooth.csv};
            \addplot[ffgreen_pv, dashed, line width=1.2pt, forget plot]
            coordinates {(0,9.02) (100,9.02)};

        \end{axis}

        \begin{axis}[
            baselineaxis,
            name=rl2vel,
            at={(rl2pos.south west)},
            anchor=north west,
            yshift=-0.35cm,
            ymin=-3,
            ymax=3,
            xlabel={Time Step},
            ylabel={Velocity (m/s)},
        ]

            \addplot[fforange_pv, line width=1.2pt]
            table[col sep=comma, x expr=\coordindex, y=vx_before]
            {data/baseline/rl2/logs/smooth/subgoals_ep6_smooth.csv};
            \addplot[fforange_pv, dashed, line width=1.2pt, forget plot]
            coordinates {(0,-2.35) (100,-2.35)};

            \addplot[fforange_pv!80, line width=1.2pt]
            table[col sep=comma, x expr=\coordindex, y=vx_before]
            {data/baseline/rl2/logs/smooth/subgoals_ep7_smooth.csv};
            \addplot[fforange_pv!80, dashed, line width=1.2pt, forget plot]
            coordinates {(0,-1.75) (100,-1.75)};

            \addplot[fforange_pv!60, line width=1.2pt]
            table[col sep=comma, x expr=\coordindex, y=vx_before]
            {data/baseline/rl2/logs/smooth/subgoals_ep8_smooth.csv};
            \addplot[fforange_pv!60, dashed, line width=1.2pt, forget plot]
            coordinates {(0,-1.45) (100,-1.45)};

            \addplot[ffyellow!60, line width=1.2pt]
            table[col sep=comma, x expr=\coordindex, y=vx_before]
            {data/baseline/rl2/logs/smooth/subgoals_ep9_smooth.csv};
            \addplot[ffyellow!60, dashed, line width=1.2pt, forget plot]
            coordinates {(0,1.45) (100,1.45)};

            \addplot[ffyellow!80, line width=1.2pt]
            table[col sep=comma, x expr=\coordindex, y=vx_before]
            {data/baseline/rl2/logs/smooth/subgoals_ep10_smooth.csv};
            \addplot[ffyellow!80, dashed, line width=1.2pt, forget plot]
            coordinates {(0,1.75) (100,1.75)};

            \addplot[ffyellow, line width=1.2pt]
            table[col sep=comma, x expr=\coordindex, y=vx_before]
            {data/baseline/rl2/logs/smooth/subgoals_ep11_smooth.csv};
            \addplot[ffyellow, dashed, line width=1.2pt, forget plot]
            coordinates {(0,2.35) (100,2.35)};

        \end{axis}

        \begin{axis}[
            baselineaxis,
            name=pearlpos,
            at={(5.25cm,0)},
            anchor=north west,
            title={PEARL},
            ymin=-13,
            ymax=13,
            ylabel={Position (m)},
            ylabel style={yshift=-.2cm},
            xticklabels=\empty,
        ]

            \addplot[fflightgreen, line width=1.2pt]
            table[col sep=comma, x expr=\coordindex, y=x_before]
            {data/baseline/pearl/logs/smooth/subgoals_ep0_smooth.csv};
            \addplot[fflightgreen, dashed, line width=1.2pt, forget plot]
            coordinates {(0,-9.02) (100,-9.02)};

            \addplot[fflightgreen!80, line width=1.2pt]
            table[col sep=comma, x expr=\coordindex, y=x_before]
            {data/baseline/pearl/logs/smooth/subgoals_ep1_smooth.csv};
            \addplot[fflightgreen!80, dashed, line width=1.2pt, forget plot]
            coordinates {(0,-5.10) (100,-5.10)};

            \addplot[fflightgreen!60, line width=1.2pt]
            table[col sep=comma, x expr=\coordindex, y=x_before]
            {data/baseline/pearl/logs/smooth/subgoals_ep2_smooth.csv};
            \addplot[fflightgreen!60, dashed, line width=1.2pt, forget plot]
            coordinates {(0,-3.14) (100,-3.14)};

            \addplot[ffgreen_pv!60, line width=1.2pt]
            table[col sep=comma, x expr=\coordindex, y=x_before]
            {data/baseline/pearl/logs/smooth/subgoals_ep3_smooth.csv};
            \addplot[ffgreen_pv!60, dashed, line width=1.2pt, forget plot]
            coordinates {(0,3.14) (100,3.14)};

            \addplot[ffgreen_pv!80, line width=1.2pt]
            table[col sep=comma, x expr=\coordindex, y=x_before]
            {data/baseline/pearl/logs/smooth/subgoals_ep4_smooth.csv};
            \addplot[ffgreen_pv!80, dashed, line width=1.2pt, forget plot]
            coordinates {(0,5.10) (100,5.10)};

            \addplot[ffgreen_pv, line width=1.2pt]
            table[col sep=comma, x expr=\coordindex, y=x_before]
            {data/baseline/pearl/logs/smooth/subgoals_ep5_smooth.csv};
            \addplot[ffgreen_pv, dashed, line width=1.2pt, forget plot]
            coordinates {(0,9.02) (100,9.02)};

        \end{axis}

        \begin{axis}[
            baselineaxis,
            name=pearlvel,
            at={(pearlpos.south west)},
            anchor=north west,
            yshift=-0.35cm,
            ymin=-3,
            ymax=3,
            xlabel={Time Step},
            ylabel={Velocity (m/s)},
        ]

            \addplot[fforange_pv, line width=1.2pt]
            table[col sep=comma, x expr=\coordindex, y=vx_before]
            {data/baseline/pearl/logs/smooth/subgoals_ep6_smooth.csv};
            \addplot[fforange_pv, dashed, line width=1.2pt, forget plot]
            coordinates {(0,-2.35) (100,-2.35)};

            \addplot[fforange_pv!80, line width=1.2pt]
            table[col sep=comma, x expr=\coordindex, y=vx_before]
            {data/baseline/pearl/logs/smooth/subgoals_ep7_smooth.csv};
            \addplot[fforange_pv!80, dashed, line width=1.2pt, forget plot]
            coordinates {(0,-1.75) (100,-1.75)};

            \addplot[fforange_pv!60, line width=1.2pt]
            table[col sep=comma, x expr=\coordindex, y=vx_before]
            {data/baseline/pearl/logs/smooth/subgoals_ep8_smooth.csv};
            \addplot[fforange_pv!60, dashed, line width=1.2pt, forget plot]
            coordinates {(0,-1.45) (100,-1.45)};

            \addplot[ffyellow!60, line width=1.2pt]
            table[col sep=comma, x expr=\coordindex, y=vx_before]
            {data/baseline/pearl/logs/smooth/subgoals_ep9_smooth.csv};
            \addplot[ffyellow!60, dashed, line width=1.2pt, forget plot]
            coordinates {(0,1.45) (100,1.45)};

            \addplot[ffyellow!80, line width=1.2pt]
            table[col sep=comma, x expr=\coordindex, y=vx_before]
            {data/baseline/pearl/logs/smooth/subgoals_ep10_smooth.csv};
            \addplot[ffyellow!80, dashed, line width=1.2pt, forget plot]
            coordinates {(0,1.75) (100,1.75)};

            \addplot[ffyellow, line width=1.2pt]
            table[col sep=comma, x expr=\coordindex, y=vx_before]
            {data/baseline/pearl/logs/smooth/subgoals_ep11_smooth.csv};
            \addplot[ffyellow, dashed, line width=1.2pt, forget plot]
            coordinates {(0,2.35) (100,2.35)};

        \end{axis}

        \begin{axis}[
            baselineaxis,
            name=cemrlpos,
            at={(10.5cm,0)},
            anchor=north west,
            title={CEMRL},
            ymin=-13,
            ymax=13,
            ylabel={Position (m)},
            ylabel style={yshift=-.2cm},
            xticklabels=\empty,
        ]

            \addplot[fflightgreen, line width=1.2pt]
            table[col sep=comma, x expr=\coordindex, y=x_before]
            {data/baseline/cemrl/logs/smooth/subgoals_ep0_smooth.csv};
            \addplot[fflightgreen, dashed, line width=1.2pt, forget plot]
            coordinates {(0,-9.02) (100,-9.02)};

            \addplot[fflightgreen!80, line width=1.2pt]
            table[col sep=comma, x expr=\coordindex, y=x_before]
            {data/baseline/cemrl/logs/smooth/subgoals_ep1_smooth.csv};
            \addplot[fflightgreen!80, dashed, line width=1.2pt, forget plot]
            coordinates {(0,-5.10) (100,-5.10)};

            \addplot[fflightgreen!60, line width=1.2pt]
            table[col sep=comma, x expr=\coordindex, y=x_before]
            {data/baseline/cemrl/logs/smooth/subgoals_ep2_smooth.csv};
            \addplot[fflightgreen!60, dashed, line width=1.2pt, forget plot]
            coordinates {(0,-3.14) (100,-3.14)};

            \addplot[ffgreen_pv!60, line width=1.2pt]
            table[col sep=comma, x expr=\coordindex, y=x_before]
            {data/baseline/cemrl/logs/smooth/subgoals_ep3_smooth.csv};
            \addplot[ffgreen_pv!60, dashed, line width=1.2pt, forget plot]
            coordinates {(0,3.14) (100,3.14)};

            \addplot[ffgreen_pv!80, line width=1.2pt]
            table[col sep=comma, x expr=\coordindex, y=x_before]
            {data/baseline/cemrl/logs/smooth/subgoals_ep4_smooth.csv};
            \addplot[ffgreen_pv!80, dashed, line width=1.2pt, forget plot]
            coordinates {(0,5.10) (100,5.10)};

            \addplot[ffgreen_pv, line width=1.2pt]
            table[col sep=comma, x expr=\coordindex, y=x_before]
            {data/baseline/cemrl/logs/smooth/subgoals_ep5_smooth.csv};
            \addplot[ffgreen_pv, dashed, line width=1.2pt, forget plot]
            coordinates {(0,9.02) (100,9.02)};

        \end{axis}

        \begin{axis}[
            baselineaxis,
            name=cemrlvel,
            at={(cemrlpos.south west)},
            anchor=north west,
            yshift=-0.35cm,
            ymin=-3,
            ymax=3,
            xlabel={Time Step},
            ylabel={Velocity (m/s)},
        ]

            \addplot[fforange_pv, line width=1.2pt]
            table[col sep=comma, x expr=\coordindex, y=vx_before]
            {data/baseline/cemrl/logs/smooth/subgoals_ep6_smooth.csv};
            \addplot[fforange_pv, dashed, line width=1.2pt, forget plot]
            coordinates {(0,-2.35) (100,-2.35)};

            \addplot[fforange_pv!80, line width=1.2pt]
            table[col sep=comma, x expr=\coordindex, y=vx_before]
            {data/baseline/cemrl/logs/smooth/subgoals_ep7_smooth.csv};
            \addplot[fforange_pv!80, dashed, line width=1.2pt, forget plot]
            coordinates {(0,-1.75) (100,-1.75)};

            \addplot[fforange_pv!60, line width=1.2pt]
            table[col sep=comma, x expr=\coordindex, y=vx_before]
            {data/baseline/cemrl/logs/smooth/subgoals_ep8_smooth.csv};
            \addplot[fforange_pv!60, dashed, line width=1.2pt, forget plot]
            coordinates {(0,-1.45) (100,-1.45)};

            \addplot[ffyellow!60, line width=1.2pt]
            table[col sep=comma, x expr=\coordindex, y=vx_before]
            {data/baseline/cemrl/logs/smooth/subgoals_ep9_smooth.csv};
            \addplot[ffyellow!60, dashed, line width=1.2pt, forget plot]
            coordinates {(0,1.45) (100,1.45)};

            \addplot[ffyellow!80, line width=1.2pt]
            table[col sep=comma, x expr=\coordindex, y=vx_before]
            {data/baseline/cemrl/logs/smooth/subgoals_ep10_smooth.csv};
            \addplot[ffyellow!80, dashed, line width=1.2pt, forget plot]
            coordinates {(0,1.75) (100,1.75)};

            \addplot[ffyellow, line width=1.2pt]
            table[col sep=comma, x expr=\coordindex, y=vx_before]
            {data/baseline/cemrl/logs/smooth/subgoals_ep11_smooth.csv};
            \addplot[ffyellow, dashed, line width=1.2pt, forget plot]
            coordinates {(0,2.35) (100,2.35)};

        \end{axis}

        \begin{axis}[
            baselineaxis,
            name=meltspos,
            at={(15.75cm,0)},
            anchor=north west,
            title={MELTS},
            ymin=-13,
            ymax=13,
            ylabel={Position (m)},
            ylabel style={yshift=-.2cm},
            xticklabels=\empty,
        ]

            \addplot[fflightgreen, line width=1.2pt]
            table[col sep=comma, x expr=\coordindex, y=x_before]
            {data/baseline/melts/logs/smooth/subgoals_ep0_smooth.csv};
            \addplot[fflightgreen, dashed, line width=1.2pt, forget plot]
            coordinates {(0,-9.02) (100,-9.02)};

            \addplot[fflightgreen!80, line width=1.2pt]
            table[col sep=comma, x expr=\coordindex, y=x_before]
            {data/baseline/melts/logs/smooth/subgoals_ep1_smooth.csv};
            \addplot[fflightgreen!80, dashed, line width=1.2pt, forget plot]
            coordinates {(0,-5.10) (100,-5.10)};

            \addplot[fflightgreen!60, line width=1.2pt]
            table[col sep=comma, x expr=\coordindex, y=x_before]
            {data/baseline/melts/logs/smooth/subgoals_ep2_smooth.csv};
            \addplot[fflightgreen!60, dashed, line width=1.2pt, forget plot]
            coordinates {(0,-3.14) (100,-3.14)};

            \addplot[ffgreen_pv!60, line width=1.2pt]
            table[col sep=comma, x expr=\coordindex, y=x_before]
            {data/baseline/melts/logs/smooth/subgoals_ep3_smooth.csv};
            \addplot[ffgreen_pv!60, dashed, line width=1.2pt, forget plot]
            coordinates {(0,3.14) (100,3.14)};

            \addplot[ffgreen_pv!80, line width=1.2pt]
            table[col sep=comma, x expr=\coordindex, y=x_before]
            {data/baseline/melts/logs/smooth/subgoals_ep4_smooth.csv};
            \addplot[ffgreen_pv!80, dashed, line width=1.2pt, forget plot]
            coordinates {(0,5.10) (100,5.10)};

            \addplot[ffgreen_pv, line width=1.2pt]
            table[col sep=comma, x expr=\coordindex, y=x_before]
            {data/baseline/melts/logs/smooth/subgoals_ep5_smooth.csv};
            \addplot[ffgreen_pv, dashed, line width=1.2pt, forget plot]
            coordinates {(0,9.02) (100,9.02)};

        \end{axis}

        \begin{axis}[
            baselineaxis,
            name=meltsvel,
            at={(meltspos.south west)},
            anchor=north west,
            yshift=-0.35cm,
            ymin=-3,
            ymax=3,
            xlabel={Time Step},
            ylabel={Velocity (m/s)},
        ]

            \addplot[fforange_pv, line width=1.2pt]
            table[col sep=comma, x expr=\coordindex, y=vx_before]
            {data/baseline/melts/logs/smooth/subgoals_ep6_smooth.csv};
            \addplot[fforange_pv, dashed, line width=1.2pt, forget plot]
            coordinates {(0,-2.35) (100,-2.35)};

            \addplot[fforange_pv!80, line width=1.2pt]
            table[col sep=comma, x expr=\coordindex, y=vx_before]
            {data/baseline/melts/logs/smooth/subgoals_ep7_smooth.csv};
            \addplot[fforange_pv!80, dashed, line width=1.2pt, forget plot]
            coordinates {(0,-1.75) (100,-1.75)};

            \addplot[fforange_pv!60, line width=1.2pt]
            table[col sep=comma, x expr=\coordindex, y=vx_before]
            {data/baseline/melts/logs/smooth/subgoals_ep8_smooth.csv};
            \addplot[fforange_pv!60, dashed, line width=1.2pt, forget plot]
            coordinates {(0,-1.45) (100,-1.45)};

            \addplot[ffyellow!60, line width=1.2pt]
            table[col sep=comma, x expr=\coordindex, y=vx_before]
            {data/baseline/melts/logs/smooth/subgoals_ep9_smooth.csv};
            \addplot[ffyellow!60, dashed, line width=1.2pt, forget plot]
            coordinates {(0,1.45) (100,1.45)};

            \addplot[ffyellow!80, line width=1.2pt]
            table[col sep=comma, x expr=\coordindex, y=vx_before]
            {data/baseline/melts/logs/smooth/subgoals_ep10_smooth.csv};
            \addplot[ffyellow!80, dashed, line width=1.2pt, forget plot]
            coordinates {(0,1.75) (100,1.75)};

            \addplot[ffyellow, line width=1.2pt]
            table[col sep=comma, x expr=\coordindex, y=vx_before]
            {data/baseline/melts/logs/smooth/subgoals_ep11_smooth.csv};
            \addplot[ffyellow, dashed, line width=1.2pt, forget plot]
            coordinates {(0,2.35) (100,2.35)};

        \end{axis}

        \node[fill=ffgreen_pv!80, below left,
        label={[label distance=0pt]0:\textsf{goal forward}}]
        (l1) at ([xshift=-1cm, yshift=-1cm]rl2vel.south) {};
        \node[fill=fflightgreen!80, below left,
        label={[label distance=0pt]0:\textsf{goal backward}}]
        (l2) at ([xshift=-1cm, yshift=-1cm]pearlvel.south) {};
        \node[fill=ffyellow!80, below left,
        label={[label distance=0pt]0:\textsf{velocity forward}}]
        (l3) at ([xshift=-1cm, yshift=-1cm]cemrlvel.south) {};
        \node[fill=fforange_pv!80, below left,
        label={[label distance=0pt]0:\textsf{velocity backward}}]
        (l4) at ([xshift=-1cm, yshift=-1cm]meltsvel.south) {};

        \node[above left] at ([xshift=-.5cm, yshift=.25cm]rl2pos.north west) {\textbf{a}};
        \node[above left] at ([xshift=-.5cm, yshift=.25cm]pearlpos.north west) {\textbf{b}};
        \node[above left] at ([xshift=-.5cm, yshift=.25cm]cemrlpos.north west) {\textbf{c}};
        \node[above left] at ([xshift=-.5cm, yshift=.25cm]meltspos.north west) {\textbf{d}};

    \end{tikzpicture}
    }
    \vskip -.05in
    \caption{
    Baseline comparison of inference trajectories on four non-parametric tasks for Half-Cheetah embodiment.
    \textbf{a}, RL$^2$; \textbf{b}, PEARL; \textbf{c}, CEMRL; and \textbf{d}, MELTS.
    For each baseline, the upper plot shows position tracking and the lower plot shows velocity tracking.
    Dashed horizontal lines denote the corresponding ground-truth targets.
    }
    \label{fig:baseline_multi_inference_all}
\vskip -.1in
\end{figure*}

%% file: img_tex/tracking_mse_comparison.tex
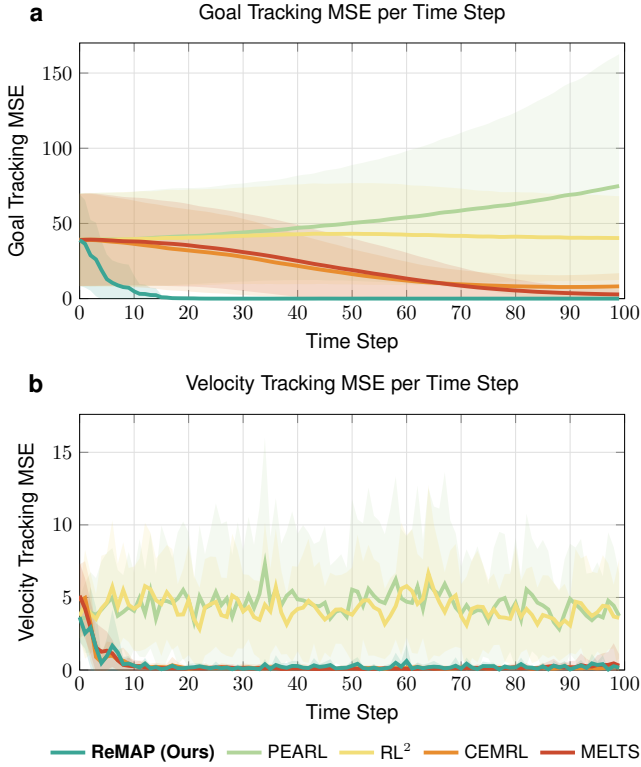
\begin{figure}[t!]
    \centering
    \resizebox{\linewidth}{!}{
    \begin{tikzpicture}[font=\sffamily]

    \pgfplotstableread[col sep=comma]
    {data/baseline/mse_baselines_comparison/tracking_mse_goal.csv}
    \goalmsedata

    \pgfplotstableread[col sep=comma]
    {data/baseline/mse_baselines_comparison/tracking_mse_velocity.csv}
    \velocitymsedata

    \pgfplotsset{
        mseaxis/.style={
            width=11cm,
            height=6cm,
            xlabel={Time Step},
            grid=major,
            grid style={gray!25},
            xmin=0,
            xmax=100,
            ymin=0,
            axis line style={black},
            every axis plot/.append style={mark=none},
            set layers=standard
        }
    }

    \begin{axis}[
        mseaxis,
        name=goalplot,
        at={(0,0)},
        anchor=north west,
        ylabel={Goal Tracking MSE},
        title={Goal Tracking MSE per Time Step},
        ymax=170,
        legend to name=mselegend,
        legend columns=5,
        legend style={
            font=\small\sffamily,
            draw=none,
            fill=none,
            /tikz/every even column/.append style={column sep=0.1cm}
        },
    ]

    \addplot[name path=goal_ours_upper, draw=none, forget plot]
    table[x=timestep, y expr=\thisrow{Ours_mean_mse}+\thisrow{Ours_std_mse}]{\goalmsedata};

    \addplot[name path=goal_ours_lower, draw=none, forget plot]
    table[x=timestep, y expr=\thisrow{Ours_mean_mse}-\thisrow{Ours_std_mse}]{\goalmsedata};

    \addplot[ffgreen_pv, fill opacity=0.16, forget plot]
    fill between[of=goal_ours_upper and goal_ours_lower];

    \addplot[ffgreen_pv, line width=1.9pt, on layer=axis foreground]
    table[x=timestep, y=Ours_mean_mse]{\goalmsedata};
    \addlegendentry{\textbf{ReMAP (Ours)}}

    \addplot[name path=goal_pearl_upper, draw=none, forget plot]
    table[x=timestep, y expr=\thisrow{PEARL_mean_mse}+\thisrow{PEARL_std_mse}]{\goalmsedata};

    \addplot[name path=goal_pearl_lower, draw=none, forget plot]
    table[x=timestep, y expr=\thisrow{PEARL_mean_mse}-\thisrow{PEARL_std_mse}]{\goalmsedata};

    \addplot[fflightgreen, fill opacity=0.14, forget plot]
    fill between[of=goal_pearl_upper and goal_pearl_lower];

    \addplot[fflightgreen, line width=1.9pt]
    table[x=timestep, y=PEARL_mean_mse]{\goalmsedata};
    \addlegendentry{PEARL}

    \addplot[name path=goal_rl2_upper, draw=none, forget plot]
    table[x=timestep, y expr=\thisrow{RL2_mean_mse}+\thisrow{RL2_std_mse}]{\goalmsedata};

    \addplot[name path=goal_rl2_lower, draw=none, forget plot]
    table[x=timestep, y expr=\thisrow{RL2_mean_mse}-\thisrow{RL2_std_mse}]{\goalmsedata};

    \addplot[ffyellow, fill opacity=0.15, forget plot]
    fill between[of=goal_rl2_upper and goal_rl2_lower];

    \addplot[ffyellow, line width=1.9pt]
    table[x=timestep, y=RL2_mean_mse]{\goalmsedata};
    \addlegendentry{RL$^2$}

    \addplot[name path=goal_cemrl_upper, draw=none, forget plot]
    table[x=timestep, y expr=\thisrow{CEMRL_mean_mse}+\thisrow{CEMRL_std_mse}]{\goalmsedata};

    \addplot[name path=goal_cemrl_lower, draw=none, forget plot]
    table[x=timestep, y expr=\thisrow{CEMRL_mean_mse}-\thisrow{CEMRL_std_mse}]{\goalmsedata};

    \addplot[fforange_pv, fill opacity=0.14, forget plot]
    fill between[of=goal_cemrl_upper and goal_cemrl_lower];

    \addplot[fforange_pv, line width=1.9pt]
    table[x=timestep, y=CEMRL_mean_mse]{\goalmsedata};
    \addlegendentry{CEMRL}

    \addplot[name path=goal_melts_upper, draw=none, forget plot]
    table[x=timestep, y expr=\thisrow{MELTS_mean_mse}+\thisrow{MELTS_std_mse}]{\goalmsedata};

    \addplot[name path=goal_melts_lower, draw=none, forget plot]
    table[x=timestep, y expr=\thisrow{MELTS_mean_mse}-\thisrow{MELTS_std_mse}]{\goalmsedata};

    \addplot[ffred_pv, fill opacity=0.14, forget plot]
    fill between[of=goal_melts_upper and goal_melts_lower];

    \addplot[ffred_pv, line width=1.9pt]
    table[x=timestep, y=MELTS_mean_mse]{\goalmsedata};
    \addlegendentry{MELTS}

    \end{axis}

    \begin{axis}[
        mseaxis,
        name=velplot,
        at={(goalplot.south west)},
        anchor=north west,
        yshift=-2cm,
        ylabel={Velocity Tracking MSE},
        title={Velocity Tracking MSE per Time Step},
    ]

    \addplot[name path=vel_ours_upper, draw=none, forget plot]
    table[x=timestep, y expr=\thisrow{Ours_mean_mse}+\thisrow{Ours_std_mse}]{\velocitymsedata};

    \addplot[name path=vel_ours_lower, draw=none, forget plot]
    table[x=timestep, y expr=\thisrow{Ours_mean_mse}-\thisrow{Ours_std_mse}]{\velocitymsedata};

    \addplot[ffgreen_pv, fill opacity=0.16, forget plot]
    fill between[of=vel_ours_upper and vel_ours_lower];

    \addplot[ffgreen_pv, line width=1.9pt, on layer=axis foreground]
    table[x=timestep, y=Ours_mean_mse]{\velocitymsedata};

    \addplot[name path=vel_pearl_upper, draw=none, forget plot]
    table[x=timestep, y expr=\thisrow{PEARL_mean_mse}+\thisrow{PEARL_std_mse}]{\velocitymsedata};

    \addplot[name path=vel_pearl_lower, draw=none, forget plot]
    table[x=timestep, y expr=\thisrow{PEARL_mean_mse}-\thisrow{PEARL_std_mse}]{\velocitymsedata};

    \addplot[fflightgreen, fill opacity=0.14, forget plot]
    fill between[of=vel_pearl_upper and vel_pearl_lower];

    \addplot[fflightgreen, line width=1.9pt]
    table[x=timestep, y=PEARL_mean_mse]{\velocitymsedata};

    \addplot[name path=vel_rl2_upper, draw=none, forget plot]
    table[x=timestep, y expr=\thisrow{RL2_mean_mse}+\thisrow{RL2_std_mse}]{\velocitymsedata};

    \addplot[name path=vel_rl2_lower, draw=none, forget plot]
    table[x=timestep, y expr=\thisrow{RL2_mean_mse}-\thisrow{RL2_std_mse}]{\velocitymsedata};

    \addplot[ffyellow, fill opacity=0.15, forget plot]
    fill between[of=vel_rl2_upper and vel_rl2_lower];

    \addplot[ffyellow, line width=1.9pt]
    table[x=timestep, y=RL2_mean_mse]{\velocitymsedata};

    \addplot[name path=vel_cemrl_upper, draw=none, forget plot]
    table[x=timestep, y expr=\thisrow{CEMRL_mean_mse}+\thisrow{CEMRL_std_mse}]{\velocitymsedata};

    \addplot[name path=vel_cemrl_lower, draw=none, forget plot]
    table[x=timestep, y expr=\thisrow{CEMRL_mean_mse}-\thisrow{CEMRL_std_mse}]{\velocitymsedata};

    \addplot[fforange_pv, fill opacity=0.14, forget plot]
    fill between[of=vel_cemrl_upper and vel_cemrl_lower];

    \addplot[fforange_pv, line width=1.9pt]
    table[x=timestep, y=CEMRL_mean_mse]{\velocitymsedata};

    \addplot[name path=vel_melts_upper, draw=none, forget plot]
    table[x=timestep, y expr=\thisrow{MELTS_mean_mse}+\thisrow{MELTS_std_mse}]{\velocitymsedata};

    \addplot[name path=vel_melts_lower, draw=none, forget plot]
    table[x=timestep, y expr=\thisrow{MELTS_mean_mse}-\thisrow{MELTS_std_mse}]{\velocitymsedata};

    \addplot[ffred_pv, fill opacity=0.14, forget plot]
    fill between[of=vel_melts_upper and vel_melts_lower];

    \addplot[ffred_pv, line width=1.9pt]
    table[x=timestep, y=MELTS_mean_mse]{\velocitymsedata};

    \end{axis}

    \node[anchor=north] at ([xshift=-.1cm, yshift=-3cm]velplot.south) {\ref{mselegend}};

    \node[above left]at([xshift=-.5cm, yshift=.25cm]goalplot.north west){\large\textbf{a}};
    \node[above left]at([xshift=-.5cm, yshift=-1.75cm]velplot.north west){\large\textbf{b}};

    \end{tikzpicture}
    }
    \vskip -.05in
    \caption{
    Tracking MSE comparison between our ReMAP and baseline methods (Half-Cheetah 4 tasks).
    \textbf{a}, Goal tracking forward and backward MSE over time steps.
    \textbf{b}, Velocity tracking forward and backward MSE over time steps.
    Solid curves denote the mean MSE over at least 50 trials for each task, and shaded regions indicate the standard deviation ($\mu\pm\sigma$).
    Lower MSE indicates more accurate tracking performance.
    }
    \label{fig:tracking_mse_comparison}
\vskip -.15in
\end{figure}

%% file: img_tex/cost_comparison.tex
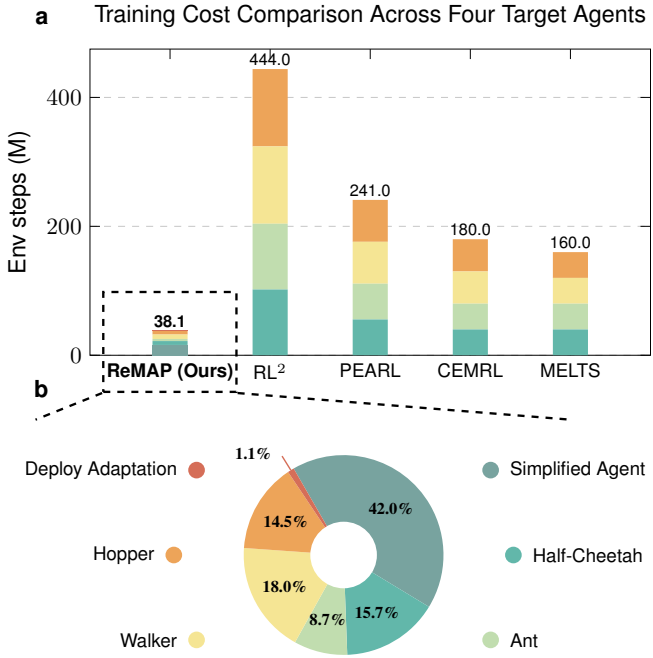
\begin{figure}[t!]
\centering
\resizebox{\linewidth}{!}{
\begin{tikzpicture}[font=\sffamily]

\tikzset{
    legbox/.style={
        draw=none,
        minimum width=0.22cm,
        minimum height=0.14cm,
        inner sep=0pt
    },
    legtext/.style={
        font=\scriptsize\sffamily,
        anchor=west,
        inner sep=1pt
    }
}

\begin{axis}[
    name=costplot,
    title={Training Cost Comparison Across Four Target Agents},
    width=10cm,
    height=6.18cm,
    ymin=0,
    ymax=475,
    ymajorgrids,
    grid style={dashed, gray!45},
    axis line style={black},
    tick style={black},
    xtick=data,
    ylabel={Env steps (M)},
    symbolic x coords={Ours, RL$^2$, PEARL, CEMRL, MELTS},
    xticklabels={\textbf{ReMAP (Ours)}, RL$^2$, PEARL, CEMRL, MELTS},
    xticklabel style={font=\footnotesize\sffamily},
    enlarge x limits=0.2,
    ybar stacked,
    bar width=15pt,
]

\addplot[fill=ffdarkgreen!80, draw=none] coordinates {(Ours,16.0)  (RL$^2$,0) (PEARL,0)  (CEMRL,0) (MELTS,0)}; 
\addplot[fill=ffgreen_pv!80, draw=none] coordinates {(Ours,5.99) (RL$^2$, 102) (PEARL, 55.5) (CEMRL, 40) (MELTS, 40)}; 
\addplot[fill=fflightgreen!80, draw=none] coordinates {(Ours, 3.30) (RL$^2$, 102) (PEARL, 55.5)  (CEMRL, 40) (MELTS, 40)}; 
\addplot[fill=ffyellow!80, draw=none] coordinates {(Ours,6.87) (RL$^2$, 120) (PEARL, 65)  (CEMRL, 50) (MELTS, 40)}; 
\addplot[fill=fforange_pv!80, draw=none] coordinates {(Ours, 5.52) (RL$^2$, 120) (PEARL, 65) (CEMRL, 50) (MELTS, 40)}; 
\addplot[fill=ffred_pv!80, draw=none] coordinates {(Ours,0.40) (RL$^2$,0) (PEARL,0) (CEMRL,0) (MELTS,0)}; 

\node[font=\bfseries\scriptsize\sffamily] at (axis cs:Ours,53) {38.1};
\node[font=\scriptsize\sffamily] at (axis cs:PEARL,256) {241.0};
\node[font=\scriptsize\sffamily] at (axis cs:RL$^2$,459) {444.0};
\node[font=\scriptsize\sffamily] at (axis cs:MELTS,175) {160.0};
\node[font=\scriptsize\sffamily] at (axis cs:CEMRL,195) {180.0};

\end{axis}

\node[rectangle, draw=black, line width=1pt, dashed, minimum width=2cm, minimum height=1.5cm](note)at(1.2cm,0.2cm){};
\draw[line width=1pt, dashed](note.south west) -- ++ (-1cm, -0.4cm);
\draw[line width=1pt, dashed](note.south east) -- ++ (5cm, -0.4cm);


\def\R{1.5cm}           
\def\Hole{1cm}          
\def\LabelR{1.0cm}      
\def\LegendR{2.55cm}    

\def\angA{120}
\def\angB{-31.26}       
\def\angC{-87.89}       
\def\angD{-119.09}      
\def\angE{-184.04}      
\def\angF{-236.22}      
\def\angG{-240.00}      

\begin{scope}[shift={(costplot.south)}, xshift=-.4cm, yshift=-3cm]

    \fill[ffdarkgreen!80]
        (0,0) -- (\angA:\R) arc (\angA:\angB:\R) -- cycle;

    \fill[ffgreen_pv!80]
        (0,0) -- (\angB:\R) arc (\angB:\angC:\R) -- cycle;

    \fill[fflightgreen!80]
        (0,0) -- (\angC:\R) arc (\angC:\angD:\R) -- cycle;

    \fill[ffyellow!80]
        (0,0) -- (\angD:\R) arc (\angD:\angE:\R) -- cycle;

    \fill[fforange_pv!80]
        (0,0) -- (\angE:\R) arc (\angE:\angF:\R) -- cycle;

    \fill[ffred_pv!80]
        (0,0) -- (\angF:\R) arc (\angF:\angG:\R) -- cycle;

    \node[circle, fill=white, minimum size=\Hole] at (0,0) {};


    \node[font=\scriptsize\bfseries] at (44.37:\LabelR)    {42.0\%};
    \node[font=\scriptsize\bfseries] at (-59.57:\LabelR)   {15.7\%};
    \node[font=\scriptsize\bfseries] at (-103.49:\LabelR)  {8.7\%};
    \node[font=\scriptsize\bfseries] at (-151.56:\LabelR)  {18.0\%};
    \node[font=\scriptsize\bfseries] at (149.87:\LabelR)   {14.5\%};

    \draw[ffred_pv!80, line width=0.6pt]
        (-238.11:\R) -- (-238.11:1.75cm);
    \node[font=\scriptsize\bfseries, anchor=east]
        at (-238.11:1.8cm) {1.1\%};


    \node[circle, fill=ffdarkgreen!80, inner sep=0pt, minimum size=.25cm,
        label={[label distance=1pt]0:\textsf{Simplified Agent}}]
        at (30:\LegendR) {};

    \node[circle, fill=ffgreen_pv!80, inner sep=0pt, minimum size=.25cm,
        label={[label distance=1pt]0:\textsf{Half-Cheetah}}]
        at (0:\LegendR) {};

    \node[circle, fill=fflightgreen!80, inner sep=0pt, minimum size=.25cm,
        label={[label distance=1pt]0:\textsf{Ant}}]
        at (-30:\LegendR) {};

    \node[circle, fill=ffred_pv!80, inner sep=0pt, minimum size=.25cm,
        label={[label distance=1pt]180:\textsf{Deploy Adaptation}}]
        at (150:\LegendR) {};

    \node[circle, fill=fforange_pv!80, inner sep=0pt, minimum size=.25cm,
        label={[label distance=1pt]180:\textsf{Hopper}}]
        at (180:\LegendR) {};

    \node[circle, fill=ffyellow!80, inner sep=0pt, minimum size=.25cm,
        label={[label distance=1pt]180:\textsf{Walker}}]
        at (210:\LegendR) {};

\end{scope}

\node[above left]at([xshift=-.5cm, yshift=.25cm]costplot.north west){\textbf{a}};
\node[below left]at([xshift=-.5cm, yshift=-.25cm]costplot.south west){\textbf{b}};

\end{tikzpicture}
}
\vskip -.05in
\caption{
Training cost comparison between ReMAP and baseline methods across four target agents.
\textbf{a}, Total environment steps, measured in millions, required for training convergence. ReMAP achieves substantially lower total cost by reusing high-level meta-knowledge and training only embodiment-specific low-level controllers for target agents.
\textbf{b}, Cost decomposition of ReMAP, showing the relative contribution of simplified-agent high-level training, target-agent low-level controller training, and online deployment adaptation.
}
\label{fig:cost_comparison_transfer}
\vskip -.15in
\end{figure}

%% file: main.bbl
\begin{thebibliography}{10}
\providecommand{\url}[1]{#1}
\csname url@samestyle\endcsname
\providecommand{\newblock}{\relax}
\providecommand{\bibinfo}[2]{#2}
\providecommand{\BIBentrySTDinterwordspacing}{\spaceskip=0pt\relax}
\providecommand{\BIBentryALTinterwordstretchfactor}{4}
\providecommand{\BIBentryALTinterwordspacing}{\spaceskip=\fontdimen2\font plus
\BIBentryALTinterwordstretchfactor\fontdimen3\font minus
  \fontdimen4\font\relax}
\providecommand{\BIBforeignlanguage}[2]{{%
\expandafter\ifx\csname l@#1\endcsname\relax
\typeout{** WARNING: IEEEtran.bst: No hyphenation pattern has been}%
\typeout{** loaded for the language `#1'. Using the pattern for}%
\typeout{** the default language instead.}%
\else
\language=\csname l@#1\endcsname
\fi
#2}}
\providecommand{\BIBdecl}{\relax}
\BIBdecl

\bibitem{duan2016rl}
Y.~Duan, J.~Schulman, X.~Chen, P.~L. Bartlett, I.~Sutskever, and P.~Abbeel,
  ``Rl$^{2}$: Fast reinforcement learning via slow reinforcement learning,''
  \emph{arXiv preprint arXiv:1611.02779}, 2016.

\bibitem{beck2023recurrent}
J.~Beck, R.~Vuorio, Z.~Xiong, and S.~Whiteson, ``Recurrent hypernetworks are
  surprisingly strong in meta-rl,'' \emph{Advances in Neural Information
  Processing Systems}, vol.~36, pp. 62\,121--62\,138, 2023.

\bibitem{finn2017model}
C.~Finn, P.~Abbeel, and S.~Levine, ``Model-agnostic meta-learning for fast
  adaptation of deep networks,'' in \emph{International conference on machine
  learning}.\hskip 1em plus 0.5em minus 0.4em\relax PMLR, 2017, pp. 1126--1135.

\bibitem{bing2022meta}
Z.~Bing, D.~Lerch, K.~Huang, and A.~Knoll, ``Meta-reinforcement learning in
  non-stationary and dynamic environments,'' \emph{IEEE Transactions on Pattern
  Analysis and Machine Intelligence}, vol.~45, no.~3, pp. 3476--3491, 2022.

\bibitem{rakelly2019efficient}
K.~Rakelly, A.~Zhou, C.~Finn, S.~Levine, and D.~Quillen, ``Efficient off-policy
  meta-reinforcement learning via probabilistic context variables,'' in
  \emph{International conference on machine learning}.\hskip 1em plus 0.5em
  minus 0.4em\relax PMLR, 2019, pp. 5331--5340.

\bibitem{yu2020meta}
T.~Yu, D.~Quillen, Z.~He, R.~Julian, K.~Hausman, C.~Finn, and S.~Levine,
  ``Meta-world: A benchmark and evaluation for multi-task and meta
  reinforcement learning,'' in \emph{Conference on robot learning}.\hskip 1em
  plus 0.5em minus 0.4em\relax PMLR, 2020, pp. 1094--1100.

\bibitem{10495171}
Z.~Bing, Y.~Yun, K.~Huang, and A.~Knoll, ``Context-based meta-reinforcement
  learning with bayesian nonparametric models,'' \emph{IEEE Transactions on
  Pattern Analysis and Machine Intelligence}, vol.~46, no.~10, pp. 6948--6965,
  2024.

\bibitem{hughes2013memoized}
M.~C. Hughes and E.~Sudderth, ``Memoized online variational inference for
  dirichlet process mixture models,'' \emph{Advances in neural information
  processing systems}, vol.~26, 2013.

\bibitem{nichol2018first}
A.~Nichol, J.~Achiam, and J.~Schulman, ``On first-order meta-learning
  algorithms,'' \emph{arXiv preprint arXiv:1803.02999}, 2018.

\bibitem{nguyen2021robust}
T.~Nguyen, T.~Luu, T.~Pham, S.~Rakhimkul, and C.~D. Yoo, ``Robust maml:
  Prioritization task buffer with adaptive learning process for model-agnostic
  meta-learning,'' in \emph{ICASSP 2021-2021 IEEE International Conference on
  Acoustics, Speech and Signal Processing (ICASSP)}.\hskip 1em plus 0.5em minus
  0.4em\relax IEEE, 2021, pp. 3460--3464.

\bibitem{mishra2018simple}
N.~Mishra, M.~Rohaninejad, X.~Chen, and P.~Abbeel, ``A simple neural attentive
  meta-learner,'' in \emph{International Conference on Learning
  Representations}, 2018.

\bibitem{shala2025efficient}
G.~Shala, A.~Biedenkapp, P.~Krack, F.~Walter, and J.~Grabocka, ``Efficient
  cross-episode meta-rl,'' in \emph{The Thirteenth International Conference on
  Learning Representations}, 2025.

\bibitem{zintgraf2020varibad}
L.~Zintgraf, K.~Shiarlis, M.~Igl, S.~Schulze, Y.~Gal, K.~Hofmann, and
  S.~Whiteson, ``Varibad: a very good method for bayes-adaptive deep rl via
  meta-learning,'' \emph{Proceedings of ICLR 2020}, 2020.

\bibitem{li2020focal}
L.~Li, R.~Yang, and D.~Luo, ``Focal: Efficient fully-offline meta-reinforcement
  learning via distance metric learning and behavior regularization,'' in
  \emph{International Conference on Learning Representations}, 2020.

\bibitem{wang2023meta}
M.~Wang, Z.~Bing, X.~Yao, S.~Wang, H.~Kai, H.~Su, C.~Yang, and A.~Knoll,
  ``Meta-reinforcement learning based on self-supervised task representation
  learning,'' in \emph{Proceedings of the AAAI Conference on Artificial
  Intelligence}, vol.~37, no.~8, 2023, pp. 10\,157--10\,165.

\bibitem{lu2024survey}
Y.~Lu, H.~Li, Y.~Li, Y.~Lin, and X.~Peng, ``A survey on deep clustering: from
  the prior perspective,'' \emph{Vicinagearth}, vol.~1, no.~1, p.~4, 2024.

\bibitem{jiang2016variational}
Z.~Jiang, Y.~Zheng, H.~Tan, B.~Tang, and H.~Zhou, ``Variational deep embedding:
  An unsupervised and generative approach to clustering,'' \emph{arXiv preprint
  arXiv:1611.05148}, 2016.

\bibitem{dilokthanakul2016deep}
N.~Dilokthanakul, P.~A. Mediano, M.~Garnelo, M.~C. Lee, H.~Salimbeni,
  K.~Arulkumaran, and M.~Shanahan, ``Deep unsupervised clustering with gaussian
  mixture variational autoencoders,'' \emph{arXiv preprint arXiv:1611.02648},
  2016.

\bibitem{reynolds2009gaussian}
D.~A. Reynolds \emph{et~al.}, ``Gaussian mixture models.'' \emph{Encyclopedia
  of biometrics}, vol. 741, no. 659-663, p.~3, 2009.

\bibitem{meng2025preserving}
Y.~Meng, Z.~Bing, X.~Yao, K.~Chen, K.~Huang, Y.~Gao, F.~Sun, and A.~Knoll,
  ``Preserving and combining knowledge in robotic lifelong reinforcement
  learning,'' \emph{Nature Machine Intelligence}, vol.~7, no.~2, pp. 256--269,
  2025.

\bibitem{nalisnick2017stick}
E.~Nalisnick and P.~Smyth, ``Stick-breaking variational autoencoders,'' in
  \emph{International Conference on Learning Representations}, 2017.

\bibitem{goyal2017nonparametric}
P.~Goyal, Z.~Hu, X.~Liang, C.~Wang, and E.~P. Xing, ``Nonparametric variational
  auto-encoders for hierarchical representation learning,'' in
  \emph{Proceedings of the IEEE International Conference on Computer Vision},
  2017, pp. 5094--5102.

\bibitem{ronen2022deepdpm}
M.~Ronen, S.~E. Finder, and O.~Freifeld, ``Deepdpm: Deep clustering with an
  unknown number of clusters,'' in \emph{Proceedings of the IEEE/CVF Conference
  on Computer Vision and Pattern Recognition}, 2022, pp. 9861--9870.

\bibitem{blei2006variational}
D.~M. Blei and M.~I. Jordan, ``Variational inference for dirichlet process
  mixtures,'' \emph{Bayesian Analysis}, vol.~1, no.~1, pp. 121--144, 2006.

\bibitem{hejna2020hierarchically}
D.~Hejna, L.~Pinto, and P.~Abbeel, ``Hierarchically decoupled imitation for
  morphological transfer,'' in \emph{International Conference on Machine
  Learning}.\hskip 1em plus 0.5em minus 0.4em\relax PMLR, 2020, pp. 4159--4171.

\bibitem{gupta2022metamorph}
A.~Gupta, L.~Fan, S.~Ganguli, and L.~Fei-Fei, ``Metamorph: Learning universal
  controllers with transformers,'' \emph{arXiv preprint arXiv:2203.11931},
  2022.

\bibitem{sutton1999between}
R.~S. Sutton, D.~Precup, and S.~Singh, ``Between mdps and semi-mdps: A
  framework for temporal abstraction in reinforcement learning,''
  \emph{Artificial intelligence}, vol. 112, no. 1-2, pp. 181--211, 1999.

\bibitem{kingma2013auto}
D.~P. Kingma and M.~Welling, ``Auto-encoding variational bayes,'' \emph{arXiv
  preprint arXiv:1312.6114}, 2013.

\bibitem{haarnoja2018soft}
T.~Haarnoja, A.~Zhou, P.~Abbeel, and S.~Levine, ``Soft actor-critic: Off-policy
  maximum entropy deep reinforcement learning with a stochastic actor,'' in
  \emph{International conference on machine learning}.\hskip 1em plus 0.5em
  minus 0.4em\relax Pmlr, 2018, pp. 1861--1870.

\bibitem{todorov2012mujoco}
E.~Todorov, T.~Erez, and Y.~Tassa, ``Mujoco: A physics engine for model-based
  control,'' in \emph{2012 IEEE/RSJ international conference on intelligent
  robots and systems}.\hskip 1em plus 0.5em minus 0.4em\relax IEEE, 2012, pp.
  5026--5033.

\end{thebibliography}
